\documentclass{article}

\usepackage{microtype}
\usepackage{graphicx}
\usepackage{booktabs} % 专业的表格线
\usepackage{hyperref}
\usepackage{amsmath}

\usepackage[preprint]{icml2026}

\usepackage{times}
\usepackage{latexsym}
\usepackage[T1]{fontenc}
\usepackage[utf8]{inputenc}
\usepackage{inconsolata}

\usepackage{amsmath}
\usepackage{amssymb}
\usepackage{amsfonts}
\usepackage{bm}          % 粗体向量
\usepackage{fp}          % 浮点运算
\usepackage{stmaryrd}    % 特殊数学符号
\usepackage{pifont}      % 特殊符号 (如 \ding)
\usepackage{textcomp}
\usepackage{gensymb}     % 角度符号等

\usepackage{multirow}
\usepackage{colortbl}    % 表格颜色
\usepackage{makecell}    % 表格换行
\usepackage{wrapfig}     % 文字绕排
\usepackage{caption}
\usepackage{subcaption}
\usepackage{float}
\usepackage{resizegather} % 自动缩放公式

\usepackage{tikz}
\usepackage{pgfplots}
\usepackage[most]{tcolorbox}

\usepackage{siunitx}     % 单位处理
\usepackage{enumitem}    % 列表定制
\usepackage{calc}        % 长度计算
\usepackage[normalem]{ulem} % 下划线
\usepackage{afterpage}
\usepackage{fontawesome5} % 图标
\usepackage{xspace}       % 智能空格

\usepackage{wrapfig}

\definecolor{indianred}{rgb}{0.8, 0.36, 0.36}
\definecolor{bleudefrance}{rgb}{0.19, 0.55, 0.91}
\definecolor{forestgreen}{rgb}{0.0, 0.5, 0.0}
\definecolor{ashgrey}{rgb}{0.7, 0.75, 0.71}
\definecolor{darkorange}{rgb}{1.0, 0.55, 0.0}
\definecolor{darkorchid}{rgb}{0.6, 0.2, 0.8}
\definecolor{backred}{rgb}{1.0, 0.6, 0.6}
\definecolor{wdcolor}{RGB}{128, 0, 255}

\definecolor{gray_header}{gray}{0.95} 
\definecolor{gray_row}{gray}{0.90}

\begin{document}

% ICML 标题块必须放在 twocolumn[...] 中
\twocolumn[
\icmltitle{UHR-BAT: Budget-Aware Token Compression Vision-Language model for Ultra-High-Resolution Remote Sensing}

\begin{icmlauthorlist}
\icmlauthor{Yunkai Dang\textsuperscript{*}}{sai}
\icmlauthor{Minxin Dai\textsuperscript{*}}{sai}
\icmlauthor{Yuekun Yang}{sai}
\icmlauthor{Zhangnan Li}{ese}
\icmlauthor{Wenbin Li\textsuperscript{$\dagger$}}{sai}
\icmlauthor{Feng Miao}{phy}
\icmlauthor{Yang Gao}{sai}
\end{icmlauthorlist}

\icmlaffiliation{sai}{School of Artificial Intelligence Science and Technology, Nanjing University}
\icmlaffiliation{ese}{School of Electronic Science and Engineering, Nanjing University}
\icmlaffiliation{phy}{School of Physics, Nanjing University}

\icmlcorrespondingauthor{Wenbin Li}{liwenbin.nju@gmail.com, yunkaidang1@gmail.com}

\icmlkeywords{Machine Learning, ICML, Vision Language Models}

\vskip 0.3in
]

\printAffiliationsAndNotice{\icmlEqualContribution \quad $\dagger$ Corresponding author.}

% =========================================================
% 6. 正文内容
% =========================================================

\begin{abstract}
Ultra-high-resolution (UHR) remote sensing imagery couples kilometer-scale context with query-critical evidence that may occupy only a few pixels.
Such vast spatial scale leads to a quadratic explosion of visual tokens and hinders the extraction of information from small objects.
Previous works utilize direct downsampling, dense tiling, or global top‑K pruning, which either compromise query-critical image details or incur unpredictable compute.
In this paper, we propose UHR-BAT, a query-guided and region-faithful token compression framework to efficiently select visual tokens under strict context budget.
Specifically, we leverage text-guided, multi-scale importance estimation for visual tokens, effectively tackling the challenge of achieving precise yet low-cost feature extraction. 
Furthermore, by introducing region-wise preserve and merge strategies, we mitigate visual token redundancy, further driving down the computational budget.
The experimental results show that UHR-BAT achieves state-of-the-art performance across various benchmarks.
Code will be available at \url{https://github.com/Yunkaidang/UHR}.
\end{abstract}

% This makes direct use} in multimodal large language models (MLLMs) infeasible under strict context budgets due to quadratic visual-token growth. 

\section{Introduction}
\label{sec:intro}

\begin{figure}[t]
    \centering
    \includegraphics[width=1\linewidth]{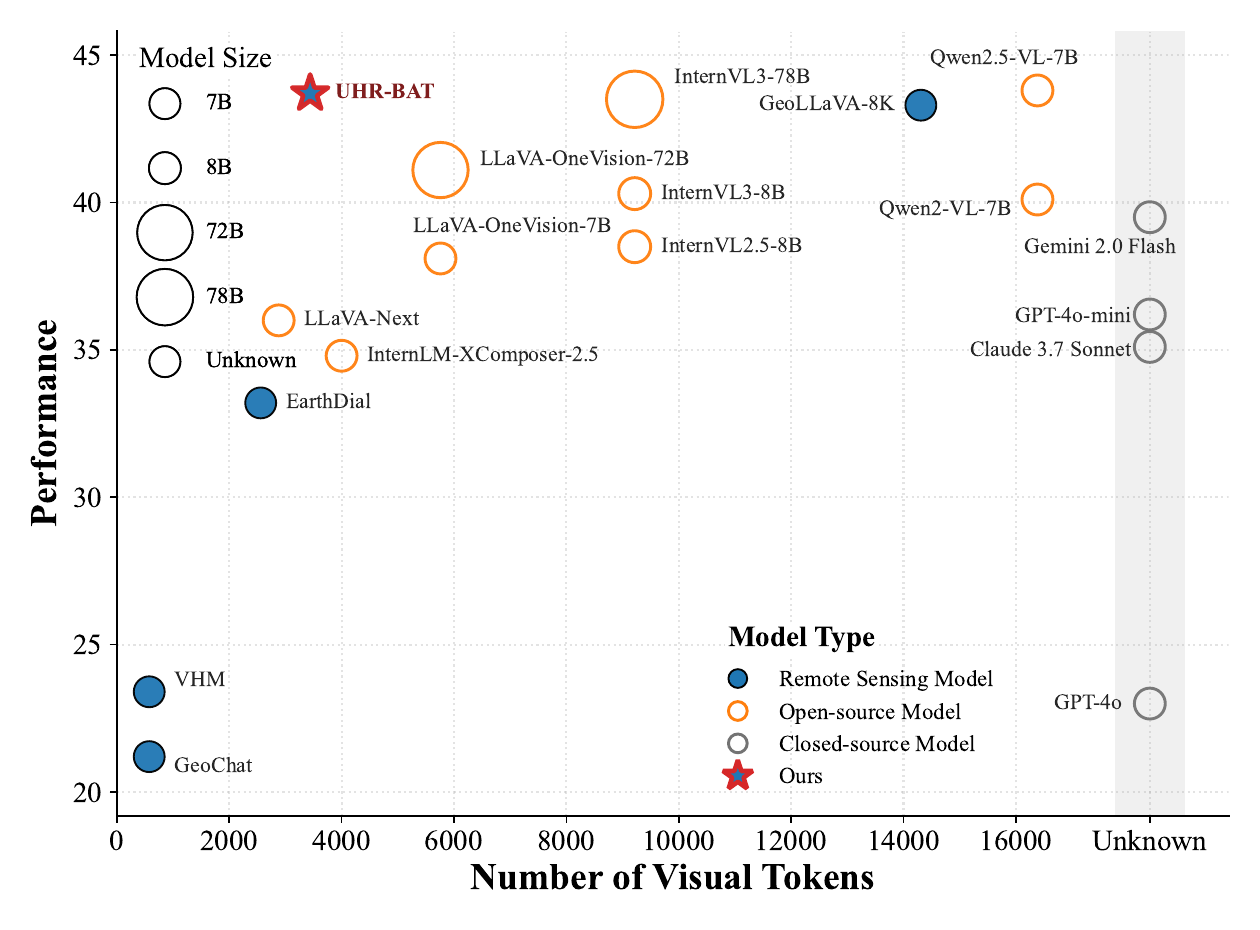}
    \caption{Accuracy--efficiency frontier for Ultra-High-Resolution Remote Sensing MLLMs on XLRS-Bench benchmark.
    We report overall weighted average accuracy versus the number of visual tokens.
    The results show that our method achieves superior performance using fewer tokens.
    }
    %which directly determines context usage, memory footprint, and latency in remote sensing MLLMs
    % Standard-resolution settings discard fine-grained geospatial evidence (e.g., small vehicles/vessels and sharp man-made boundaries), while naively increasing resolution leads to a token explosion that exceeds practical budgets.
    % \textbf{UHR-BAT} shifts the frontier by enforcing an explicit token budget through region-wise preserve-and-merge (retaining salient local evidence while softly aggregating redundant background) and by injecting additional scales via spatially aligned residual fusion without increasing the anchor sequence length.}
    \label{fig:motivation_tradeoff}
\end{figure}

Multimodal large language models (MLLMs) have rapidly advanced visual understanding by coupling robust image encoders with large language models~\citep{yin2024survey,liang2024survey,wu2023multimodal,kim2021vilt,yuan2021tokens,chen2022pali,dang2024explainable}. 
In remote sensing, these models have demonstrated strong performance on geospatial question answering and reasoning tasks~\citep{mall2023remote,cong2022satmae}. 
However, real-world satellite and aerial imagery is typically ultra-high-resolution (UHR)~\citep{liu2021swin,yang2021focal}. 
UHR visual question answering presents a critical challenge: images simultaneously encode kilometer-scale scene layouts, while the evidence required for reasoning is often extremely fine-grained~\citep{li2024vrsbench}. 
For instance, targets like small boats or vehicles may occupy only a few pixels within a multi-million-pixel image. 
Consequently, accurately localizing and reasoning about such small objects remains a critical challenge, yet it is essential for practical geospatial workflows.

In the domain of natural image processing, adaptive resolution and tiling-based strategies are widely adopted to extract details from high-resolution inputs~\citep{llava-uhd,zhang2024llava,InternLM-XComposer}.
However, these pipelines often fragment global context and incur prohibitive computational costs as the number of tiles scales~\citep{jaegle2021perceiver,jaegle2021perceiverio}.
%\textcolor{red}{~\citep{}}.
%加一个引用参考文献
In the context of remote sensing, general MLLMs typically resort to aggressive downsampling to address ultra-high-resolution (UHR) imagery. 
%discards，blurs，可以换一下词汇表述？
This compression sacrifices critical fine-grained details, thereby eliminating small-scale objects and obscuring semantic boundaries.
% general Multimodal Large Language Models (MLLMs) typically resort to aggressive downsampling or rigid slicing to accommodate ultra-high-resolution (UHR) imagery.
Recent specialized frameworks attempt to address these limitations but face trade-offs~\citep{wang2025geollava8k,liu2025zoomearth,zhou2025look,luo2025large}.
For instance, GeoLLaVA-8K~\cite{wang2025geollava8k} employs visual pruning to discard background tokens.
%这个看如何一句话介绍清楚
However, this text-agnostic approach frequently discards task-essential tokens while retaining substantial irrelevant information.
% yet it still retains a substantial number of tokens.
% Similarly, ZooMearth~\cite{liu2025zoomearth,zhou2025look} and Coarse-to-Fine Text-Guided Token Pruning~\cite{luo2025large} utilizes a training-free, coarse-to-fine zooming mechanism to recover answer-critical details.~\citep{rao2021dynamicvit,ryoo2021tokenlearner,liang2022not,bolya2022token}
Moreover, other works~\cite{liu2025zoomearth,zhou2025look,luo2025large} employ 
text-guided, coarse-to-fine zooming for token selection. 
%coarse to fine text guided Token Pruning
However, this iterative regional selection process inevitably increases inference latency. 
Furthermore, it preserves a high volume of visually redundant tokens, limiting computational efficiency.
% However, this iterative region selection precludes direct UHR processing and inevitably increases inference latency.
% Critically, existing methods overlook the challenge of processing UHR images within a \textbf{fixed token budget}.
% They lack effective mechanisms to maximize information density by pruning redundancy while preserving only the most query-relevant regions.
%这里可以一句话介绍清楚？
Collectively, existing methods fail to process UHR imagery within a fixed token budget, as they lack mechanisms to prune redundancy while isolating 
query-relevant features.

To address these challenges, we propose a \textit{Budget-Aware Token} Compression framework tailored for \textit{UHR} Remote-Sensing MLLMs (UHR-BAT).
UHR-BAT follows two principles.
Compression should be \emph{query-guided}, so that it retains evidence relevant to the question.
Compression should also be \emph{region-faithful}, so that it preserves coverage across semantically coherent regions and avoids suppressing sparse targets.
Specifically, we introduce a Top-down query-guided mechanism that leverages text-aligned global priors to localize regions of interest. 
This ensures that fine-grained details are extracted exclusively from query-relevant areas, effectively avoiding visual redundancy.
 % To differentiate these multi-scale features, we incorporate scale embeddings into the visual tokens to distinguish information across varying resolutions.
% As a result, our approach simultaneously captures global and local information in a single forward pass, forming a comprehensive yet efficient visual representation.
% Concretely, we construct multi-scale token streams to capture global context and fine details.
% It aligns token importance across scales by transferring a reference attention map via cross-scale interpolation.
% contribution2
% To capture more informative features across all scales, we then introduce region-wise preserve and merge to facilitate the efficient selection of task-relevant tokens.
To further enforce region faithfulness, we propose the Region-wise Preserve and Merge (RPM) strategy to minimize redundancy by selectively preserving tokens in query-relevant areas while compressing the background. 
This strategy filters out irrelevant noise while maintaining essential evidence.
As shown in Figure~\ref{fig:motivation_tradeoff}, our model achieves strong performance using substantially fewer visual tokens than other models.
Experiments on the XLRS-Bench and RSHR-Bench benchmarks demonstrate that 
our model significantly enhances performance under constrained context budgets. 
Furthermore, this approach effectively shifts the accuracy-efficiency frontier for ultra-high-resolution remote sensing.

% Experiments on UHR remote sensing VQA benchmarks XLRS-Bench and RSHR-Bench show that UHR-BAT improves performance under tighter context budgets.
% Besides, this effectively shifts the accuracy--efficiency frontier for UHR remote sensing MLLMs.
% It also reduces memory and latency by large margins, enabling practical UHR inference on modern hardware. 
% As shown in Figure~\ref{fig:motivation_tradeoff}, our lightweight architecture achieves superior results compared to high-parameter counterparts.
% Within each selected region, salient tokens are preserved to retain fine-grained evidence.
% The remaining tokens are softly aggregated into compact regional representatives instead of being discarded.
% This aggregation retains coarse regional context and ensures that every region contributes to the final representation.
% Crucially, UHR-BAT enforces an explicit token budget and outputs a fixed-length visual token sequence that fits within the context window.
% \textcolor{red}{Finally, we inject non-anchor scales into an anchor stream via \emph{spatially aligned residual fusion}, which adds multi-scale cues without increasing sequence length.}
% dataset , 

In summary, we highlight the primary novel contributions of this work as follows:
\begin{itemize}
    \item We propose UHR-BAT, a token compression framework designed for ultra high resolution remote-sensing MLLMs under strict context budgets.

    %这部分没讲设计位置编码，贡献有点弱
    \item  We propose a query-guided, multi-scale input mechanism to integrate text-derived global priors and capture both holistic context and fine-grained details.
    % \item \textcolor{red}{We propose a multi-scale input mechanism coupled with scale-aware positional embeddings to capture both global context and fine-grained details.}
    % We introduce a multi-scale input mechanism and positional embeddings to capture both global context and fine-grained details.
    %To differentiate between these scales, we incorporate multi-scale positional embedding.}
    % We develop cross-scale importance alignment and a spatially aligned residual fusion mechanism that injects additional scales without increasing the anchor sequence length.

    \item We propose region-wise preserve and merge strategies to preserve salient local evidence and aggregate redundant background into compact representatives.
    %Consequently, it retains more question-relevant information under strict context budgets.}
    % We propose region-wise preserve-and-merge, which preserves salient local evidence and softly aggregates redundant background into compact representatives while maintaining region coverage.

    \item Empirical results across standard benchmarks confirm that UHR-BAT establishes a new state-of-the-art for efficient UHR understanding, outperforming existing methods under strict token budgets.
    
\end{itemize}
\section{Related Works}

\paragraph{Multimodal Large Language Models.}
MLLMs integrate a vision encoder with an LLM through a projection or cross-attention interface, enabling unified multimodal understanding and generation~\citep{alayrac2022flamingo,ye2023mplugowl2,dang2025exploring}.
Early models such as Flamingo~\citep{alayrac2022flamingo} demonstrated few-shot multimodal prompting with tightly designed vision--language connectors.
A complementary line aligns frozen components via lightweight bridging modules: BLIP-2~\citep{li2023blip} introduces the Q-Former to couple a frozen vision encoder with an LLM, and InstructBLIP~\citep{dai2024instructblip} further improves instruction following with instruction-aware alignment.
Building on these designs, open-source instruction-tuned models such as LLaVA~\citep{li2024llava} and MiniGPT-4~\citep{zhu2023minigpt} show that curated instruction data is critical for robust multimodal dialogue.
Subsequent efforts such as ShareGPT4V~\citep{chen2023sharegpt4v} scale and refine instruction data and training recipes, strengthening generalization and response quality.
% Beyond global image understanding, grounding-oriented MLLMs such as Kosmos-2~\citep{peng2023kosmos} and Shikra~\citep{chen2023shikra} extend to region referring and grounded dialogue by incorporating localization details.
Recent models~\citep{abdin2024phi3,hurst2024gpt,anthropic2025claude37sonnet,gemini,li2024llava} further enhance fine-grained and text-centric reasoning with stronger visual backbones and improved training strategies, including Qwen2.5-VL~\citep{Qwen2.5-VL}, InternVL3~\citep{zhu2025internvl3}, and InternLM-XComposer~\citep{InternLM-XComposer}.
In parallel, lightweight variants such as MobileVLM~\citep{chu2023mobilevlm,chu2024mobilevlm}, TinyGPT-V~\citep{yuan2023tinygpt}, LLaVA-Phi~\citep{zhu2024llava}, and MiniCPM-V~\citep{yao2024minicpm} target efficient deployment under limited compute budgets.

\paragraph{Remote Sensing Foundation Models.} 
Remote-sensing vision-language models apply open-vocabulary reasoning to earth observation. 
% RemoteCLIP~\citep{liu2024remoteclip} is the first vision-language foundation model for remote sensing, which scales pre-training through a novel annotation unification pipeline to achieve state-of-the-art zero-shot performance.
RSGPT~\citep{lu2023rsicap} pioneers RS-specific conversational modeling for captioning and QA. 
GeoChat~\citep{geochat} enhances grounding by enabling region-specific prompts and outputting bounding boxes. 
EarthGPT~\citep{earthgpt2024} and EarthDial~\citep{soni2025earthdial} unify multiple interpretation tasks within a single interface. 
Similarly, SkyEyeGPT~\citep{Zhan2025SkyEyeGPT} extends instruction tuning for broad task coverage, while RS-LLaVA~\citep{Bazi2024RSLLaVA} adapts general VLM frameworks for RS captioning and VQA.
To address reliability, H2RSVLM~\citep{h2rsvlm2024} explicitly models honesty and rejects unanswerable queries. 
Other works incorporate structured knowledge; SkySenseGPT~\citep{skysensegpt2024} utilizes scene-graph supervision, and LHRS-Bot~\citep{muhtar2024lhrs} leverages VGI-enhanced signals for alignment.
MF-RSVLM~\citep{dang2025fuse} addresses the loss of fine-grained details by integrating multi-scale feature fusion and employing a recurrent mechanism to reinject visual signals during long-context generation.
% Most existing methods are restricted to low-resolution regimes. 
% They typically operate on images with dimensions below $1024 \times 1024$ pixels.
However, these models are limited to low-resolution datasets, typically processing images with resolution smaller than $1024 \times 1024$ pixels.

\paragraph{High-Resolution MLLMs in General Domains.}
High-resolution MLLMs are fundamentally constrained by token explosion and the resulting attention and memory overhead when preserving fine-grained structures~\citep{yang2026annotation}.
In general domains, LLaVA-UHD~\citep{llava-uhd} processes arbitrary aspect ratios via resolution-aware slicing and token-efficient encoding, and LLaVA-UHD v2~\citep{zhang2024llava} further improves multi-scale perception with feature pyramids and hierarchical window attention.
However, pipelines based on slicing or tiling often fragment global context and incur increased computation as the number of tiles grows.
To alleviate cross-slice discontinuity, HiRes-LLaVA~\citep{huang2025hires} introduces restoration and aggregation mechanisms, and InternLM-XComposer2-4KHD~\citep{dong2024internlm} supports dynamic resolutions up to 4K via flexible patch configurations for text-centric understanding.
Nevertheless, these methods are generally not query-adaptive and still spend substantial tokens and compute on redundant regions.
In remote sensing, GeoLLaVA-8K~\citep{wang2025geollava8k} scales to $8\text{K}\times 8\text{K}$ by exploiting background redundancy through background token pruning and anchored token selection, while ZoomEarth~\citep{liu2025zoomearth} and ZoomSearch~\citep{zhou2025look} adopt coarse-to-fine zooming to recover answer-critical details.
% Compared with these approaches, iterative zooming increases inference rounds/latency and pruning strategies can still under-allocate capacity to sparse, query-critical targets.
% Different from them, our method instead enforces a fixed visual-token budget, retaining both global context and localized evidence in a single fixed-length token sequence.
In contrast, our method enforces a strict visual-token budget. 
It preserves both global context and local evidence within a fixed-length sequence.
% via query-guided, region-wise preserve-and-merge with cross-scale importance alignment, 

% From the vision backbone perspective, CLIP-style encoders (e.g., EVA-CLIP) are widely adopted as general-purpose image representations~\citep{sun2023eva}, but their standard low-resolution settings can limit high-resolution perception. To better exploit high-resolution signals, several MLLMs adopt explicit grid-based partitioning and independent encoding of image regions, including Monkey~\citep{monkey}, SPHINX-X~\citep{gao2024sphinx}, liu2024llavanext~\citep{liu2024llavanext}, and InternLM-XComposer-2.5~\citep{zhang2024internlm}, which collectively improve recognition and reasoning in high-resolution, text-heavy scenarios by increasing effective spatial coverage while preserving compatibility with LLM-context constraints.

% =========================
% METHOD
% =========================
\section{Method}
\label{sec:method}

\subsection{Overview}
High-resolution remote sensing imagery demands simultaneously modeling (i) \emph{global context} such as large-scale land-use layout and long-range spatial dependencies, and (ii) \emph{fine-grained evidence} such as small objects and sharp boundaries.
Directly feeding high-resolution visual tokens into Multimodal Large Language Models (MLLMs) is infeasible: standard vision encoders produce prohibitively long token sequences that exceed the limited context length of modern MLLMs.
% Naïve remedies (e.g., aggressive downsampling, tiling, or global top-$K$ pruning) either destroy global coherence or systematically suppress localized cues.
Simple remedies (e.g., heavy downsampling, tiling, or global top-$K$ pruning) either compromise global coherence or suppress localized cues.
We therefore seek a \emph{query-guided} compression strategy that removes redundancy while remaining \emph{region-faithful} to small targets and structurally diverse scenes.

Our method comprises three components.
First, we compute query-aware visual token importance from the vision--language attention interface (\S\ref{sec:prelim}) and use it as a common scoring signal across views.
Second, we construct multi-scale token streams with \emph{scale-specific positional embedding} and \emph{cross-scale importance alignment} to make scores comparable across resolutions (\S\ref{sec:pos}).
Third, we introduce Region-wise Preserve-and-Merge, which preserves informative tokens within each region while merging redundant ones into compact representatives (\S\ref{sec:preserve-merge}).
% , and then fuse non-anchor scales via \emph{spatially aligned residual injection} without increasing sequence length (\S\ref{sec:preserve-merge}).

% (Optional) Method overview figure (replace the filename with yours)
\begin{figure*}[t]
    \centering
    \includegraphics[width=1\linewidth]{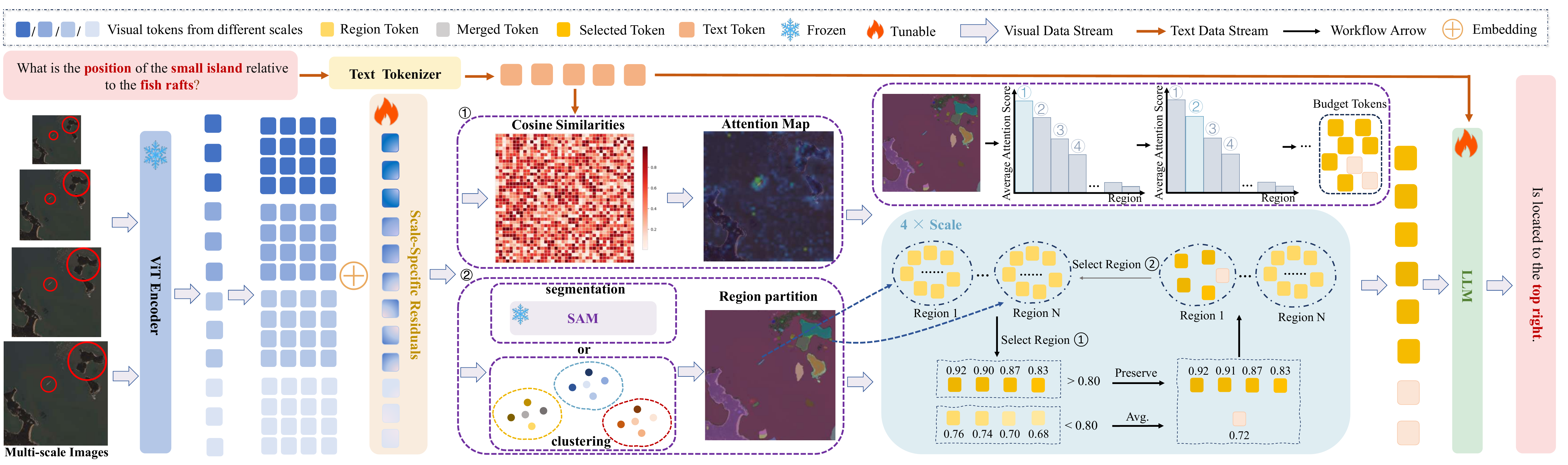}
    \caption{\textbf{Overview of our method.}
    We encode a high-resolution remote sensing image at multiple scales using a frozen ViT, applying Scale-Specific Positional Embeddings to distinguish visual tokens across scales, and then obtain an anchor-scale query-to-vision attention map from the MLLM interface. 
    A region partition (e.g., induced by SAM or feature+coordinate clustering) enables region-wise preserve-and-merge, while remaining tokens are merged via average pooling to retain coarse context. 
    A final top-$k$ step enforces the per-scale token budget. 
    Finally, the processed sequence is fed into the LLM to generate the final answer.}
    \label{fig:method_overview}
\end{figure*}

\subsection{Preliminaries}
\label{sec:prelim}

\textbf{MLLM Architecture.}
We consider a generic MLLM parameterized by $\theta$, consisting of a vision encoder $\mathcal{V}$, a text embedding layer $\mathcal{T}$, a vision--language projector $\phi$, and a stack of $L$ transformer layers.
Given a high-resolution image $I\in\mathbb{R}^{H\times W\times 3}$ and a textual query $\mathbf{x}$, the vision encoder outputs a grid of visual features $F=\mathcal{V}(I)\in\mathbb{R}^{N\times d_v}$, where $N=U\cdot V$ is the number of visual tokens on a $U\times V$ grid.
We map them into the model embedding space via $\phi:\mathbb{R}^{d_v}\!\to\mathbb{R}^{d}$, yielding visual tokens
\begin{equation}
E=\{\mathbf{e}_i\}_{i=1}^{N}=\phi(F),\qquad \mathbf{e}_i\in\mathbb{R}^{d}.
\label{eq:tokens_hr}
\end{equation}
Each token $i$ is also associated with a 2D coordinate $\mathbf{pos}_i\in[0,1]^2$ given by the center of its receptive field on the token grid.
In parallel, the query is embedded into text tokens $T=\mathcal{T}(\mathbf{x})\in\mathbb{R}^{M\times d}$.
% The transformer takes an ordered sequence with visual tokens followed by text tokens, denoted by $\mathrm{Concat}(E,T)$, and generates the response autoregressively.
The transformer processes a concatenated sequence of visual tokens $E$ and text tokens $T$, denoted as $\mathrm{Concat}(E, T)$. 
The model then generates the response in an autoregressive manner.

\textbf{Query-aware Token Importance.}
We quantify the relevance of each visual token $\mathbf{e}_i$ to the query using text-to-vision attention weights.
Let $\mathbf{A}\in\mathbb{R}^{M\times N}$ denote the cross-modal attention matrix extracted from the vision--language attention interface (we average attention over heads and use a fixed interface layer in our implementation).
We aggregate attention across text tokens to obtain a scalar importance score for each visual token:
% \citep{lai2024lisa}
\begin{equation}
a_i=\frac{1}{M}\sum_{j=1}^{M}\mathbf{A}_{j,i}.
\label{eq:attn_score}
\end{equation}

\textbf{Token Budget.}
Due to the limited context length, we impose a hard cap $B\ll N$ on the number of \emph{visual} tokens passed to the language model (text tokens and generated tokens are budgeted separately in practice).
To address this, we define a compression operator $C(\cdot)$ that maps dense visual tokens into a budget-feasible sequence:
\begin{equation}
\bar{E}=C(E;B,\{a_i\}_{i=1}^{N}),\qquad \text{s.t.}\quad |\bar{E}|\le B.
\label{eq:budget_hr}
\end{equation}
% In \S\ref{sec:preserve-merge}, we instantiate $C(\cdot)$ with Region-wise Preserve-and-Merge and an explicit budget enforcement step to guarantee the constraint.\
In \S\ref{sec:preserve-merge}, we instantiate $C(\cdot)$ via Region-wise Preserve-and-Merge. 
We further introduce an explicit budget enforcement step to ensure that the constraints are strictly satisfied.

% \textbf{Region partition.}
% Pure importance-based pruning can collapse into a few dominant areas and miss sparse targets.
% To impose structural diversity, we partition the token index set $\mathcal{I}=\{1,\dots,N\}$ into $R$ disjoint regions:
% \begin{equation}
% \mathcal{P}=\{\mathcal{S}_m\}_{m=1}^{R},\qquad 
% \bigcup_{m=1}^{R}\mathcal{S}_m=\mathcal{I},\quad \mathcal{S}_m\cap \mathcal{S}_n=\emptyset.
% \label{eq:partition_hr}
% \end{equation}
% The partition is defined on the token grid and can be instantiated by clustering in feature--coordinate space or by segmentation models (e.g., SAM) with token-level mapping (Appendix~\ref{sec:appendix_partition}).
% We apply the same idea independently at each scale.
\textbf{Region Partition.}
Pure importance-based pruning often retains tokens encoding overlapping information.
These redundant tokens consume a significant fraction of the budget.
Under a hard cap, this biases the selection toward dominant regions, causing sparse yet critical targets to be missed.
To encourage structural diversity and preserve fine-grained details, we partition the token index set $\mathcal{I}=\{1,\dots,N\}$ at the original resolution into $R$ disjoint regions:
% we partition the token index set $\mathcal{I}=\{1,\dots,N\}$ into $R$ disjoint regions:
\begin{equation}
\mathcal{P}=\{\mathcal{S}_m\}_{m=1}^{R},\qquad
\bigcup_{m=1}^{R}\mathcal{S}_m=\mathcal{I},\quad \mathcal{S}_m\cap \mathcal{S}_n=\emptyset.
\label{eq:partition_hr}
\end{equation}
Our goal is to group tokens that convey the same semantic information.
Therefore, we instantiate the partition using methods that detect such similarities.
Options include clustering in a joint feature-coordinate space~\citep{mcqueen1967some,hartigan1979algorithm,lloyd1982least}, or using segmentation models (e.g., SAM~\citep{kirillov2023segment}) followed by a token-level mapping (Appendix~\ref{sec:appendix_partition}).
We apply this independently at each scale.

\subsection{Scale-Specific Positional Embedding}
\label{sec:pos}
% To capture both global layout and fine details, we construct $S$ resized views $\{I^{(s)}\}_{s=1}^{S}$.
% We use $s=1$ as the \emph{anchor} (lowest-resolution) view to preserve global context, while higher $s$ provide progressively finer evidence.
% At each scale $s$, the vision encoder outputs tokens $E^{(s)}=\{\mathbf{e}_i^{(s)}\}_{i=1}^{N_s}$ on a $U_s\times V_s$ grid, where $N_s=U_sV_s$.
% Since tokens from different resolutions must remain spatially interpretable yet distinguishable, we augment each token to obtain updated features:
To capture both global context and fine details, we construct $S$ resized views $\{I^{(s)}\}_{s=1}^{S}$.
We designate the lowest-resolution view ($s=1$) as the \emph{anchor} to preserve global structure.
Higher scales provide progressively finer details.
For each scale $s$, the vision encoder outputs tokens $E^{(s)}=\{\mathbf{e}_i^{(s)}\}_{i=1}^{N_s}$.
These tokens form a $U_s\times V_s$ grid, where $N_s=U_s\cdot V_s$.
Tokens from different resolutions must remain spatially aligned and distinguishable.
Therefore, we augment each token to obtain updated features:
\begin{equation}
\mathbf{h}_i^{(s)}=\mathbf{e}_i^{(s)}+\mathbf{p}_i^{(s)}+\mathbf{q}^{(s)}.
\label{eq:scale-pos}
\end{equation}
Here $\mathbf{p}_i^{(s)}\in\mathbb{R}^{d}$ is obtained by bilinearly interpolating a base 2D positional embedding (defined on the pretraining grid) to the $U_s\times V_s$ grid, and $\mathbf{q}^{(s)}\in\mathbb{R}^{d}$ is a learned scale embedding (a lookup table over $s$).
Interpolation preserves geometric consistency across resized views, and $\mathbf{q}^{(s)}$ prevents ambiguity between tokens from different scales.

\textbf{Cross-scale Importance Alignment.}
Token selection should be comparable across scales under a unified notion of importance.
We take the anchor-scale importance as reference: we reshape the anchor token scores $\{a_i^{(1)}\}$ onto the $U_1\times V_1$ grid to form $\mathbf{A}^{(1)}\in\mathbb{R}^{U_1\times V_1}$.
For any other scale $s$, we map a target grid location $(u,v)$ to a continuous coordinate on the reference grid via a deterministic resize mapping $\tau_s(u,v)$, and define
\begin{equation}
\mathbf{A}^{(s)}_{u,v}=\Psi\!\left(\mathbf{A}^{(1)},\,\tau_s(u,v)\right),
\label{eq:attn-align}
\end{equation}
where $\Psi(\cdot)$ denotes bilinear interpolation (Appendix~\ref{sec:appendix_interp}).
Each token $i$ at scale $s$ covers a set of grid cells $\Omega_i^{(s)}$ (typically one cell for patch tokens), and we compute its aligned importance score by averaging:
\begin{equation}
a_i^{(s)}=\frac{1}{|\Omega_i^{(s)}|}\sum_{(u,v)\in\Omega_i^{(s)}}\mathbf{A}^{(s)}_{u,v}.
\label{eq:token-attn}
\end{equation}
Fixing importance to one reference map and transferring it across scales yields comparable scores $\{a_i^{(s)}\}$, which is crucial for per-scale budgeting.

% \textbf{Per-scale budgets.}
% Under a global limit $B$, we allocate per-scale budgets $\{B_s\}$ with $\sum_{s=1}^{S}B_s\le B$.
% In our default implementation, we set $B_s\propto N_s$ (and renormalize to meet the global cap), and enforce a minimum capacity $B_s\ge R_s$ to retain at least one token per region after compression.
% We then compress each scale into $\bar{E}^{(s)}=C(E^{(s)};B_s,\{a_i^{(s)}\})$ with $|\bar{E}^{(s)}|\le B_s$.
\textbf{Per-scale Budgets.}
% Under a global limit $B$, we allocate per-scale budgets $\{B_s\}$ such that $\sum_{s=1}^{S}B_s\le B$.
Given a global limit $B$, we allocate a budget $\{B_s\}$ to each scale $s$ such that $\sum_{s=1}^{S} B_s \le B$.
Higher resolutions correspond to larger $N_s$ and provide more detailed information.
Therefore, we assign larger $B_s$ values to these scales.
We also enforce a minimum capacity $B_s\ge R_s$. 
% Since the region partition $\mathcal{P}^{(s)}$ is computed using high-resolution features independent of the anchor attention, this constraint mathematically guarantees that every selected semantic region retains at least one merged representative token.
As the region partition $\mathcal{P}^{(s)}$ utilizes high-resolution features independent of the anchor attention, this constraint effectively compensates for potential deviations in attention scores, mathematically guaranteeing that all selected semantic regions retain at least one merged representative token.
% to ensure at least one token remains per region.
Finally, we compress each scale into $\bar{E}^{(s)}=C(E^{(s)};B_s,\{a_i^{(s)}\})$ with $|\bar{E}^{(s)}|\le B_s$.

% \subsection{Region-wise Preserve and Merge}
% \label{sec:preserve-merge}
% Small targets in high-resolution remote sensing imagery are intrinsically sparse.
% Global top-$K$ strategies therefore tend to concentrate the budget on visually dominant regions and marginalize localized evidence.
% We address this issue by enforcing a \emph{region-wise} retention rule: within each region we retain relatively important tokens and merge the remaining redundant tokens into a compact representative.
\subsection{Region-wise Preserve-and-Merge}
\label{sec:preserve-merge}
Small targets in high-resolution remote sensing imagery are intrinsically sparse.
Global top-$K$ strategies tend to concentrate the budget on visually dominant regions.
They often retain excessive tokens representing repetitive information.
Consequently, critical local evidence is marginalized, especially under low budget conditions.
We address this by enforcing a \emph{region-wise} retention rule.
Within each region, we preserve relatively important tokens.
The remaining redundant tokens are merged into a compact representative.

% For each scale $s$, given a region partition $\mathcal{P}^{(s)}=\{\mathcal{S}_m^{(s)}\}_{m=1}^{R_s}$, we compute the region-average importance $\mu_m^{(s)}$:
% \begin{equation}
% \mu_m^{(s)}=\frac{1}{|\mathcal{S}_m^{(s)}|}\sum_{i\in \mathcal{S}_m^{(s)}} a_i^{(s)}.
% \label{eq:region-mean-attn}
% \end{equation}
% We retain tokens whose importance meets this region-specific reference:
% \begin{equation}
% \mathcal{K}_m^{(s)}=\left\{\,i\in \mathcal{S}_m^{(s)} \mid a_i^{(s)}\ge \mu_m^{(s)}\,\right\},
% \label{eq:keep-set}
% \end{equation}
% and define the redundancy set $\mathcal{R}_m^{(s)}=\mathcal{S}_m^{(s)}\setminus \mathcal{K}_m^{(s)}$.
% By construction, $\mathcal{K}_m^{(s)}$ is non-empty for any region with real-valued scores (Appendix~\ref{sec:appendix_properties}), ensuring that every region contributes at least one informative token.

We operate on each scale $s$ independently.
Consider the region partition $\mathcal{P}^{(s)}=\{\mathcal{S}_m^{(s)}\}_{m=1}^{R_s}$.
For each region $m$, we calculate its average importance $\mu_m^{(s)}$:
\begin{equation}
\mu_m^{(s)}=\frac{1}{|\mathcal{S}_m^{(s)}|}\sum_{i\in \mathcal{S}_m^{(s)}} a_i^{(s)}.
\label{eq:region-mean-attn}
\end{equation}
Subsequently, we retain tokens with importance no lower than this region-specific reference:
\begin{equation}
\mathcal{K}_m^{(s)}=\left\{\,i\in \mathcal{S}_m^{(s)} \mid a_i^{(s)}\ge \mu_m^{(s)}\,\right\},
\label{eq:keep-set}
\end{equation}
and define the redundancy set as $\mathcal{R}_m^{(s)}=\mathcal{S}_m^{(s)}\setminus \mathcal{K}_m^{(s)}$.
By construction, $\mathcal{K}_m^{(s)}$ is non-empty for any region with real-valued scores (Appendix~\ref{sec:appendix_properties}).
This guarantees that every region contributes at least one informative token.

Rather than discarding $\mathcal{R}_m^{(s)}$, we merge it into a single representative token via mean pooling:
\begin{equation}
\tilde{\mathbf{e}}_m^{(s)}=\frac{1}{|\mathcal{R}_m^{(s)}|}\sum_{i\in \mathcal{R}_m^{(s)}} \mathbf{h}_i^{(s)},\qquad \text{if } |\mathcal{R}_m^{(s)}|>0.
\label{eq:merge-token}
\end{equation}
\textbf{Priority-based Serialization and Budgeting.}
To maximize information density, we enforce a serialization order that prioritizes regions with higher average importance. 
First, we sort the regions in descending order of their average scores $\mu_m^{(s)}$. 
Let $\pi(k)$ denote the index of the region with the $k$-th highest average importance.

Within each region, we arrange the retained tokens by descending importance $a_i^{(s)}$ to form a sorted local sequence:
\begin{equation}
\mathbf{S}_{\pi(k)}^{(s)} = \{ \mathbf{h}_i^{(s)} : i \in \mathcal{K}_{\pi(k)}^{(s)} \}_{\downarrow} ,
\label{eq:local_sort}
\end{equation}
where $\{\cdot\}_{\downarrow}$ indicates sorting in descending order based on internal token importance. 
We then construct the global candidate sequence by concatenating these ordered regional groups alongside their merged tokens:
\begin{equation}
E_{\mathrm{cand}}^{(s)} = \mathrm{Concat}_{k=1}^{R_s} \left( \mathbf{S}_{\pi(k)}^{(s)}, \, \{ \tilde{\mathbf{e}}_{\pi(k)}^{(s)} : |\mathcal{R}_{\pi(k)}^{(s)}| > 0 \} \right),
\label{eq:compressed}
\end{equation}
where $\tilde{\mathbf{e}}_{\pi(k)}^{(s)}$ is appended to the end of each regional group. 
Finally, we enforce the budget $B_s$ by selecting the top $B_s$ tokens from $E_{\mathrm{cand}}^{(s)}$. 
This ensures that the most critical regions and tokens are prioritized within the hard capacity limit.

\subsection{Overall}
\label{sec:algorithm}

The overall pipeline of our region-aware token pruning approach is summarized in 
Appendix~\ref{sec:algorithm_process}. 
Given the input visual tokens $\mathbf{X} \in \mathbb{R}^{N \times C}$, we first partition them into distinct semantic regions $\mathcal{P}=\{\mathcal{S}_1, \dots, \mathcal{S}_R\}$ using either the K-means-based or SAM-induced method described in Appendix~\ref{sec:appendix_partition}. 
Within each region $\mathcal{S}_r$, we evaluate token importance and retain only the most informative tokens set $\mathcal{K}_r$, effectively pruning the redundant background tokens.
The final output is the concatenation of retained tokens from selected regions, preserving the spatial structure and semantic integrity with reduced computational cost.
% The schematic overview of the pipeline is depicted in Appendix~\ref{alg:method}.

\section{Experiment}

\textbf{Datasets and Models.}
We evaluate on three complementary high-resolution remote-sensing benchmarks: MME-RealWorld-RS~\citep{mme}, RSHR-Bench~\citep{dang2025benchmark}, and XLRS-Bench~\citep{wang2025xlrs}.
MME-RealWorld-RS measures multiple-choice VQA skills on real-world overhead imagery, including position understanding, color recognition, and counting, and we report accuracy for each skill as well as the averaged score (Table~\ref{tab:mmerealworld}).
RSHR-Bench is a super-high-resolution benchmark built on native 4K+ remote-sensing imagery.
In this work, we follow its multiple-choice VQA protocol and report category-wise accuracies over Perception and Reasoning tasks together with their averages.
% LRS-VQA evaluates wide-area VQA on large scenes via standard answer-matching accuracy.
XLRS-Bench serves as our primary standardized suite for extremely large remote-sensing images, where we follow the official aggregation to report overall performance alongside representative subtasks.
We consider three model families, including open-source general-purpose MLLMs, closed-source MLLMs, and remote-sensing-specific VLMs.
All models are evaluated in a zero-shot setting with a unified prompt template.
For completeness, the task definitions, evaluation protocols, and table column abbreviations are provided in Appendix~\ref{Detailed Experiment Settings}.

\begin{table*}[htbp]
\footnotesize
\caption{
\textbf{Results on XLRS-Bench across perception and reasoning sub-tasks.}
`w.Avg.' indicates the sample-count-weighted average. 
The best results are \textbf{bolded}, and the second-best results are \underline{underlined}. 
Full sub-task names are provided in the appendix.
}
\label{tab:vqa}
\centering
\resizebox{\textwidth}{!}{
\begin{tabular}{l|c|cccccccc|ccccc|c}
\toprule
\multicolumn{1}{c}{\textbf{Method}} &
\multicolumn{1}{c}{\textbf{Size}} &
\multicolumn{8}{c}{\textbf{Perception}} &
\multicolumn{5}{c}{\textbf{Reasoning}} &
\multicolumn{1}{c}{\textbf{Overall}} \\
\midrule

\textbf{Sub-tasks (L-3 Capability)} &
&
\textbf{OC} & \textbf{RC} & \textbf{OLUC} & \textbf{RLUC} & \textbf{OCC} & \textbf{OCL} & \textbf{OMS} & \textbf{OSR} &
\textbf{AD} & \textbf{ECR} & \textbf{RP} & \textbf{RCCD} & \textbf{CCR} &
\textbf{w.Avg.} \\
\midrule

number of samples &
&
60 & 100 & 100 & 200 & 800 & 800 & 60 & 500 &
100 & 100 & 100 & 60 & 100 &
3080 \\
\midrule

\multicolumn{16}{l}{\textit{Open-source MLLMs}} \\
InternLM-XComposer-2.5~\cite{zhang2024internlm} &
7B &
21.7 & 42.0 & 7.0 & 68.0 & 31.8 & 27.8 & 6.7 & 26.0 &
72.0 & 81.0 & 41.0 & 36.7 & 47.0 &
34.8 \\
LLaVA-Next~\cite{liu2024llavanext} &
7B &
26.7 & 40.0 & 5.0 & 67.0 & 28.8 & 32.8 & \textbf{66.7} & 30.0 &
69.0 & 78.0 & 27.0 & 35.0 & 36.0 &
36.0 \\
LLaVA-OneVision-7B~\cite{li2024llava} &
7B &
25.0 & 38.0 & 2.0 & 69.5 & 35.9 & 35.3 & 65.0 & 25.2 &
\underline{76.0} & \textbf{83.0} & 24.0 & 43.3 & 36.0 &
38.1 \\
InternVL2.5-8B~\cite{chen2024expanding} &
8B &
38.3 & 37.0 & 10.0 & \textbf{77.0} & 33.4 & 35.5 & 65.0 & 21.6 &
73.0 & \textbf{83.0} & 34.0 & \textbf{50.0} & 43.0 &
38.5 \\
InternVL3-8B~\cite{chen2024internvl} &
8B &
\underline{40.0} & 39.0 & 10.0 & 71.5 & \textbf{44.5} & 30.8 & 65.0 & 25.2 &
\textbf{77.0} & \underline{82.0} & 36.0 & 21.7 & 50.0 &
40.3 \\
Qwen2-VL-7B~\cite{Qwen2-VL} &
7B &
26.7 & 40.0 & 11.0 & 73.0 & 35.9 & 34.6 & 61.7 & 31.8 &
70.0 & 81.0 & 35.0 & 46.7 & 48.0 &
40.1 \\
Qwen2.5-VL-7B~\cite{Qwen2.5-VL} &
7B &
33.3 & 40.0 & 31.0 & \textbf{77.0} & 40.6 & \textbf{40.5} & \textbf{66.7} & \underline{36.2} &
68.0 & 72.0 & 27.0 & 38.3 & 45.0 &
\underline{43.8} \\
\midrule
LLaVA-OneVision-72B~\cite{li2024llava} &
72B &
33.3 & 38.0 & 15.0 & 72.5 & 36.3 & 36.3 & \textbf{66.7} & 35.6 &
74.0 & \textbf{83.0} & 28.0 & 36.7 & 43.0 &
41.1 \\
% Qwen2.5-VL-72B~\cite{Qwen2.5-VL} &
% 72B &
% 33.3 & 47.0 & 39.0 & 80.0 & 45.3 & 42.1 & 65.0 & 34.0 &
% 71.0 & 74.0 & 37.0 & 43.3 & 42.0 &
% 46.2 \\
InternVL3-78B~\cite{zhu2025internvl3} &
78B &
23.3 & \textbf{49.0} & 33.0 & \underline{74.0} & 42.5 & \underline{37.4} & \textbf{66.7} & 30.0 &
\underline{76.0} & 81.0 & 40.0 & 45.0 & 42.0 &
43.5 \\
\midrule

\multicolumn{16}{l}{\textit{Closed-source MLLMs}} \\
GPT-4o~\cite{hurst2024gpt} &
-- &
25.0 & 32.0 & 15.0 & 66.0 & 9.5 & 11.3 & 11.7 & 24.6 &
73.0 & 73.0 & 35.0 & 20.0 & 25.0 &
23.0 \\
GPT-4o-mini~\cite{hurst2024gpt} &
-- &
23.3 & 25.0 & 19.0 & 59.5 & 40.9 & 31.0 & 65.0 & 23.6 &
71.0 & 71.0 & 29.0 & 6.7 & 30.0 &
36.2 \\
Claude 3.7 Sonnet~\cite{anthropic2025claude37sonnet} &
-- &
27.6 & 22.7 & 17.4 & 68.4 & 30.5 & 29.9 & 63.6 & 27.6 &
64.8 & 78.4 & 34.5 & 27.8 & 32.6 &
35.1 \\
Gemini 2.0 Flash~\cite{gemini} &
-- &
\textbf{41.7} & \underline{45.0} & 38.0 & 73.5 & 34.6 & 27.6 & 61.7 & 32.0 &
73.0 & \underline{82.0} & 43.0 & 30.0 & \underline{51.0} &
39.5 \\
\midrule

\multicolumn{16}{l}{\textit{Remote Sensing MLLMs}} \\
GeoChat~\cite{geochat} &
7B &
16.7 & 29.0 & 2.0 & 23.0 & 21.1 & 16.8 & 35.0 & 24.2 &
33.0 & 43.0 & 10.0 & 24.0 & 21.0 &
21.2 \\
GeoLLaVA-8K~\cite{wang2025geollava8k} &
7B &
26.7 & 38.0 & \underline{49.0} & 69.0 & 41.6 & 31.6 & 65.0 & 35.0 &
67.0 & 78.0 & \textbf{66.0} & \textbf{50.0} & \textbf{52.0} &
43.3 \\
EarthDial~\cite{soni2025earthdial} &
7B &
18.3 & 42.0 & 1.0 & 36.0 & 31.3 & 31.0 & 65.0 & 24.8 &
62.0 & 71.0 & 43.0 & \underline{48.3} & 50.0 &
33.8 \\
VHM~\cite{pang2025vhm} &
7B &
16.7 & 30.0 & 2.0 & 26.0 & 21.4 & 16.8 & 35.0 & 25.6 &
42.0 & 53.0 & 46.0 & 28.3 & 21.0 &
23.6 \\
\midrule
% NOTE: UHR-BAT row is excluded from underline highlighting.
\rowcolor{gray!10} \textbf{UHR-BAT (Ours)} &
7B &
21.7 & 33.0 & \textbf{50.0} & 55.5 & \underline{43.5} & 33.8 & \underline{65.0} & \textbf{44.8} &
62.0 & 71.0 & \underline{54.0} & 46.7 & \underline{51.0} &
\textbf{44.0} \\
\bottomrule
\end{tabular}
}
\end{table*}

\textbf{Main Results on XLRS-Bench.}
To verify the effectiveness of our method, we evaluate our method on XLRS-Bench~\citep{wang2025xlrs} across the perception and reasoning dimensions. 
As shown in Table~\ref{tab:vqa}, we follow the official protocol with a uniform prompt for all sub-tasks and report the overall accuracy weighted by the benchmark’s sub-task sample counts (w.Avg.). 
Our model achieves a state-of-the-art performance of 44.0 w.Avg. across open-source and closed-source models.
Relative to prior remote-sensing MLLMs, our method improves the overall weighted accuracy by +22.8\% over GeoChat~\citep{geochat}, +0.7\% over GeoLLaVA-8K~\citep{wang2025geollava8k}, +10.2\% over EarthDial~\citep{soni2025earthdial}, and +20.4\% over VHM~\citep{pang2025vhm}. 
We also outperform strong closed-source models, yielding +21.0\% and +8.9\% gains over GPT-4o~\citep{hurst2024gpt} and Claude 3.7 Sonnet~\citep{anthropic2025claude37sonnet}, respectively. 
Moreover, we surpass recent large open-source vision-language models, with improvements of +2.9\% over LLaVA-OneVision-72B~\citep{li2024llava}. 
Overall, these gains across both perception and reasoning tasks indicate that our method enhances multimodal capability for high-resolution remote sensing scenarios.

 % +0.5\% over InternVL3-78B~\citep{zhu2025internvl3} and

\begin{table*}[ht]
\footnotesize
\vspace{-0.2cm}
\caption{
\textbf{RSHR-Bench results across Perception and Reasoning tasks.}
We report accuracy (\%) for each subtask.
`P.Avg.' and `R.Avg.' denote the mean accuracy over subtasks within Perception and Reasoning, respectively.
`All Avg.' denotes the mean accuracy over all Perception and Reasoning subtasks.
Subtask abbreviations are defined in the appendix.
}
\label{tab:visual_reasoning_perception_table2}
\centering
\resizebox{\textwidth}{!}{
\begin{tabular}{l|c|ccccccccc|ccccc|ccc}
\toprule
\multirow{2}{*}{\textbf{Model}} &
\multirow{2}{*}{\textbf{Size}} &
\multicolumn{9}{c}{\textbf{Perception}} &
\multicolumn{5}{c}{\textbf{Reasoning}} &
\multicolumn{3}{c}{\textbf{Average}} \\
\cmidrule(lr){3-11}\cmidrule(lr){12-16}\cmidrule(lr){17-19}
& & \textbf{COL} & \textbf{SHP} & \textbf{DET} & \textbf{OC} & \textbf{REL} & \textbf{OGD} & \textbf{RG} & \textbf{OCN} & \textbf{RCN}
& \textbf{AD} & \textbf{FP} & \textbf{MRJC} & \textbf{MRJCS} & \textbf{OSJ}
& \textbf{P.Avg.} & \textbf{R.Avg.} & \textbf{Avg.} \\
\midrule

\multicolumn{19}{l}{\textit{Open-source VLMs}} \\
InternVL2.5-8B~\cite{chen2024expanding} & 8B & 25.5 & 22.0 & 26.0 & 26.0 & 22.5 & 24.5 & 30.0 & 22.5 & 20.0 & 26.0 & 20.0 & 22.5 & 34.0 & 20.0 & 24.3 & 24.5 & 24.4 \\
InternVL3.5-8B~\cite{wang2025internvl3} & 8B & 21.5 & 28.0 & 18.0 & 21.5 & 29.0 & 28.5 & 30.0 & 26.5 & 25.0 & 20.0 & 16.0 & 29.0 & 34.0 & 26.0 & 25.3 & 25.0 & 25.2 \\
MiniCPM2\_6~\cite{yao2024minicpm} & 7B & 21.5 & 28.0 & 30.0 & 24.0 & 19.5 & 29.5 & 34.3 & 22.0 & 29.0 & 26.0 & 30.0 & 35.0 & 32.0 & 30.0 & 27.4 & 30.6 & 28.5 \\
Phi-3.5-Vision~\cite{abdin2024phi3} & 7B & 25.0 & 24.0 & 25.0 & 25.0 & 23.5 & 25.0 & 22.9 & 25.0 & 25.0 & 24.0 & 22.0 & 23.5 & 30.0 & 22.0 & 24.5 & 24.3 & 24.4 \\
Qwen2.5-VL-7B~\cite{Qwen2.5-VL} & 7B & 29.5 & 25.0 & 22.0 & 28.0 & 25.0 & 24.5 & 24.3 & 26.5 & 22.0 & 26.0 & 28.0 & 25.0 & 10.0 & 20.0 & 25.2 & 21.8 & 24.0 \\
DeepSeek-VL~\citep{lu2024deepseek} & 7B & 22.5 & 22.0 & 21.0 & 25.0 & 20.5 & 26.0 & 28.6 & 20.5 & 22.0 & 22.0 & 28.0 & 50.0 & 32.0 & 20.0 & 23.1 & 30.4 & 25.7 \\
VILA-HD~\cite{shi2025scaling} & 7B & 40.0 & 22.0 & 22.0 & 37.0 & 35.5 & 26.0 & 21.4 & 24.5 & 24.0 & 58.0 & 30.0 & 55.0 & 32.0 & 58.0 & 28.0 & 46.6 & 34.6 \\
\midrule
\multicolumn{19}{l}{\textit{Closed-source VLMs}} \\
GPT-5~\cite{openai2025gpt5thinking} & -- & 29.0 & 10.0 & 23.0 & 23.0 & 37.0 & 24.5 & 31.4 & 20.0 & 23.0 & 74.0 & 58.0 & 35.0 & 34.0 & 66.0 & 24.5 & 53.4 & 34.8 \\
GPT-4o~\cite{hurst2024gpt} & -- & 49.5 & 23.0 & 15.0 & 35.5 & 30.5 & 28.0 & 27.1 & 22.5 & 41.0 & 68.0 & 56.0 & 30.5 & 32.0 & 64.0 & 30.2 & 50.1 & 37.3 \\
GPT-4o-mini~\cite{hurst2024gpt} & -- & 41.5 & 16.0 & 29.0 & 31.5 & 31.5 & 32.0 & 28.6 & 19.5 & 32.0 & 54.0 & 54.0 & 31.5 & 48.0 & 54.0 & 29.1 & 48.3 & 36.0 \\
% Gemini-2.5-pro~\cite{comanici2025gemini} & -- & 55.0 & 18.0 & 31.0 & 40.0 & 41.5 & 32.5 & 45.7 & 25.0 & 25.0 & 66.0 & 32.0 & 41.5 & 38.0 & 50.0 & 34.9 & 45.5 & 38.7 \\
\midrule

\multicolumn{19}{l}{\textit{Remote Sensing VLMs}} \\
EarthDial~\cite{soni2025earthdial} & 7B & 41.0 & 22.0 & 21.0 & 30.0 & 32.5 & 30.5 & 27.1 & 18.0 & 31.0 & 42.0 & 30.0 & 29.5 & 32.0 & 52.0 & 28.1 & 37.1 & 31.3 \\
GeoChat~\cite{geochat} & 7B & 32.5 & 22.0 & 24.0 & 29.5 & 40.0 & 25.0 & 22.9 & 22.5 & 29.0 & 30.0 & 24.0 & 25.5 & 30.0 & 32.0 & 25.9 & 28.3 & 26.8 \\
GeoLLaVA-8K~\cite{wang2025geollava8k} & 7B & 25.0 & 24.0 & 25.0 & 25.0 & 25.0 & 25.0 & 21.4 & 25.0 & 25.0 & 24.0 & 0.0 & 0.0 & 34.0 & 22.0 & 24.5 & 16.0 & 21.5 \\
VHM~\cite{pang2025vhm} & 7B & 25.5 & 25.0 & 26.0 & 26.5 & 55.0 & 25.0 & 22.9 & 25.0 & 25.0 & 26.0 & 24.0 & 26.5 & 34.0 & 28.0 & 25.7 & 27.7 & 26.4 \\
\midrule

\rowcolor{gray!10} \textbf{UHR-BAT (Ours)} & 7B & 46.5 & 32.0 & 38.0 & 12.5 & 38.5 & 24.0 & 17.1 & 17.0 & 37.0 & 44.0 & 34.0 & 65.0 & 32.0 & 50.0 & 29.2 & 45.0 & 34.8 \\
\bottomrule
\end{tabular}
}
\end{table*}

\textbf{Main Results on RSHR-Bench.}
To verify higher-resolution remote sensing understanding, we further evaluate our method on the RSHR-Bench dataset~\citep{dang2025benchmark}, which measures performance along three facets: \textit{Perception}, \textit{Reasoning}, and \textit{Multi-turn} interaction.
As shown in Table~\ref{tab:visual_reasoning_perception_table2}, our method achieves the best overall averages among open-source and remote-sensing baselines, with 29.2 on Perception and 45.0 on Reasoning. 
On Perception, our average improves over EarthDial~\citep{soni2025earthdial}, GeoChat~\citep{geochat}, GeoLLaVA-8K~\citep{wang2025geollava8k}, and VHM~\citep{pang2025vhm} by +1.1\%, +3.3\%, +4.7\%, and +3.5\%, respectively, and also exceeds representative 4K image input open-source baselines by +1.2\% over VILA-HD~\citep{lin2023vila}. 
On Reasoning, our average yields clear gains over remote-sensing VLMs, improving by +7.9\% over EarthDial, +16.7\% over GeoChat, +29.0\% over GeoLLaVA-8K and +17.3\% over VHM, while remaining comparable to the strongest general VLMs. 
% Beyond aggregate scores, we obtain the best MRJC performance (65.0), improving by +10.0\% over the next strongest baseline, indicating stronger region-aware reasoning. 
Beyond aggregate scores, our model achieves state-of-the-art results across several key metrics. Specifically, we obtain the best performance on MRJC (65.0), SHP (32.0), and DET (38.0), outperforming the next strongest baselines by margins of +10.0\%, +4.0\%, and +8.0\%, respectively. These consistent improvements underscore our model's superior region-aware detecting and reasoning capabilities.
% For multi-turn evaluation, our method achieves the best scores on multi-turn anomaly diagnosis (MAD = 90.0) and multi-turn future prediction (MTFP = 82.0), improving over the strongest baseline by +14.9\% and +11.9\%, respectively. 
Overall, these results demonstrate improved robustness for high-resolution remote sensing perceiving and understanding, particularly under iterative, region-conditioned inference.

\textbf{Main Results on MMERealworld Benchmark.}
We further evaluate our method on widely-used benchmarks in this work, and report results on the MMERealworld-RS dataset. As shown in Table~\ref{tab:mmerealworld}, we evaluate our method by the mean accuracy over three perception-oriented sub-tasks (Position, Color, and Count). 
% Overall, we observe that scaling the vision backbone and increasing input resolution improve position localization and color discrimination consistently.
% % more consistently than counting, which remains difficult in remote-sensing imagery due to dense small instances and large scale variation. 
% Our method achieves the best overall average score (Avg.\ = 33.30), with strong performance on position (44.00) and color (42.00), while count (14.00) remains the primary bottleneck. 
Overall, we observe that scaling the vision backbone and increasing input resolution consistently enhance spatial localization and color discrimination. 
Our method achieves the highest average score (Avg. = 33.33), demonstrating robust performance in position (44.00) and color (42.00) tasks.
% , while counting (14.00) remains the primary bottleneck.
In terms of Avg., our method improves over remote-sensing VLMs by +12.01\% relative to GeoChat, +9.15\% relative to VHM, and +4.92\% relative to GeoLLaVA-8K, and it also exceeds the closed-source GPT-4o baseline by +5.91\%. 
The gains are particularly pronounced on the perceptual dimensions: compared with the strongest remote-sensing baseline, we improve position by +8.76\% and color by +14.08\%. 
% These results suggest that robust numerosity estimation in remote-sensing imagery still requires better small-instance representations and more reliable aggregation under uncertainty.

\begin{table}[htbp]
\footnotesize
\caption{
\textbf{Results on the MMERealworld-RS dataset, with methods ranked by Avg.}
We report accuracy (\%) on Position, Color, and Count, and their mean (Avg.) for all methods.
`Params' denotes the number of model parameters. 
}
\label{tab:mmerealworld}
\centering
\resizebox{0.48\textwidth}{!}{
\begin{tabular}{l|c|c|c|c|c}
\toprule
\textbf{Method} & \textbf{Params} & \textbf{Position} & \textbf{Color} & \textbf{Count} & \textbf{Avg.} \\ \midrule
GPT-4o~\cite{hurst2024gpt} & -- & 33.52 & 29.83 & 18.90 & 27.42 \\
LLaVA-v1.5-7B~\cite{liu2023visual} & 7B & 21.48 & 22.95 & 16.31 & 20.28 \\
LLaVA-v1.6-7B~\cite{liu2024llavanext} & 7B & 26.49 & 24.06 & 20.47 & 23.70 \\
LLaVA-ov-7B~\cite{li2024llava} & 7B & 26.81 & 26.14 & 27.57 & 26.83 \\
Qwen2.5-VL-7B~\cite{Qwen2.5-VL} & 7B & 22.12 & 15.54 & 14.93 & 17.55 \\
LLaVA-HR~\cite{luo2024feast} & 7B & 35.56 & 44.30 & 7.91 & 29.26 \\
GeoChat~\cite{geochat} & 7B & 25.06 & 23.11 & 15.66 & 21.32 \\
VHM~\cite{pang2025vhm} & 7B & 35.24 & 20.32 & 16.80 & 24.18 \\
GeoLLaVA-8K~\cite{wang2025geollava8k} & 7B & 34.90 & 27.92 & 22.27 & 28.41 \\
% ImageRAG & Dynamic & 63.33 & 60.48 & 32.46 & 52.09 \\
% ZoomEye & Dynamic & 43.52 & 60.88 & 30.10 & 44.94 \\
% RAP & Dynamic & 57.62 & 64.53 & 40.25 & 54.20 \\
% ZoomSearch & Dynamic & \textbf{67.62} & \textbf{66.14} & \textbf{39.15} & \textbf{57.64} \\ 
\midrule
\rowcolor{gray!10} \textbf{UHR-BAT (Ours)} & 7B & 44.00 & 42.00 & 14.00 & 33.33 \\ \midrule
\end{tabular}
}
\end{table}

\begin{figure}[t]
    \centering
    \includegraphics[width=0.95\linewidth]{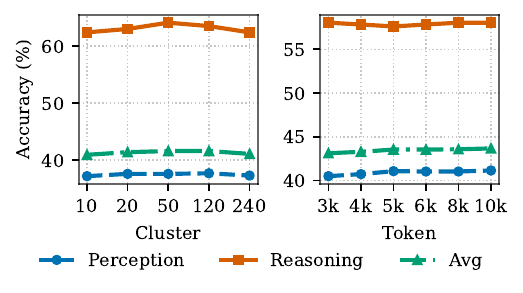}
    \vspace{-0.5cm}
    \caption{Ablation on clustering centers and retained tokens. \textbf{Left:} accuracy versus the number of cluster centers $k$. \textbf{Right:} accuracy versus the number of retained visual tokens.}
    \label{fig:cluster_token_ablation}
\end{figure}

\textbf{Ablation on Cluster Centers and Retained Tokens.}
We ablate two key hyper-parameters in our pipeline: (i) the number of cluster centers $k$ used in our clustering-based method (Figure~\ref{fig:cluster_token_ablation}, left), and (ii) the number of visual tokens retained (Figure~\ref{fig:cluster_token_ablation}, right).
We report accuracy on Perception, Reasoning, and the overall Avg score on XLRS-Bench~\citep{wang2025xlrs}.
% As shown in the left plot, varying $k$ leads to largely stable performance, suggesting that our approach is not sensitive to the precise choice of the number of cluster centers.
As shown in the left plot, the performance displays a gentle rise-and-fall pattern.
In particular, Reasoning exhibits a mild peak at a moderate $k$, while perception remains nearly unchanged across the tested range.
% As illustrated in the right plot, while increasing the token budget generally leads to performance gains, the overall improvements remain marginal. 
% Specifically, the Average score fluctuates within a narrow margin of 0.6\%, with the most noticeable improvement in Perception capped at 0.65\%. 
% Meanwhile, Reasoning performance stays comparatively stable, suggesting that reasoning-oriented tasks are less sensitive to the total token count. 
As illustrated in the right plot, increasing the token budget consistently yields performance gains, validating the effectiveness of our token selection strategy. 
However, the performance curves for both Perception and Reasoning rapidly saturate. 
This stability suggests that our model is highly efficient at extracting the most semantic-rich visual tokens, achieving near-optimal performance even under constrained budgets.
Overall, these results demonstrate that our method is capable of capturing the vast majority of essential visual features.
%even under a strict token budget, achieving a highly efficient trade-off between accuracy and computation.
% In the right plot, increasing the token budget generally improves performance, with the most pronounced gains on perception as more visual evidence is preserved.
% Meanwhile, Reasoning stays comparatively stable, indicating that reasoning-oriented questions are less sensitive to the token budget than perception-oriented ones.
% Overall, Avg. increases monotonically with the retained tokens, revealing a favorable accuracy--compute trade-off.
% Based on these results, we choose a moderate $k$ and a token budget that balances effectiveness and efficiency for all experiments.

\textbf{Computational Efficiency and Inference Latency.}
Table~\ref{tab:lat_flops_token_budget} provides a granular analysis of the computational overhead under varying token budgets. Given the massive input size typical of high-resolution remote sensing imagery (totaling 131,328 tokens in this baseline), processing the full sequence is often computationally prohibitive. Our dynamic token compression mechanism demonstrates significant efficiency gains: by increasing the compression ratio from $10.94\times$ (12k tokens) to $32.83\times$ (4k tokens), we observe a linear reduction in theoretical computational cost (TFLOPs drop from 374.24 to 267.68). Crucially, this reduction translates into tangible improvements in inference throughput, with generation latency decreasing by approximately \textbf{27.5\%} (from 5.99s to 4.34s). 
These results confirm that our method can effectively decouple input resolution from computational cost, enabling the deployment of our model on resource-constrained platforms.

\begin{table}[t]
\centering
\footnotesize
\caption{Efficiency analysis: Latency and FLOPs per sample under different token budgets (Total input tokens = 131,328).}
\label{tab:lat_flops_token_budget}
\resizebox{0.95\linewidth}{!}{%
\begin{tabular}{ccccc}
\toprule
\textbf{Compression ($\times$)} & \textbf{Kept Tokens} & \textbf{TFLOPs} & \textbf{Avg. Total (s)} & \textbf{Avg. Gen. (s)} \\
\midrule
32.83 & 4k  & 267.68 & 20.93 & 4.34 \\
26.27 & 5k  & 274.91 & 21.58 & 4.56 \\
21.89 & 6k  & 289.10 & 21.83 & 4.74 \\
16.42 & 8k  & 317.48 & 21.99 & 5.11 \\
13.13 & 10k & 345.75 & 25.50 & 5.65 \\
11.94 & 11k & 359.94 & 25.79 & 5.86 \\
10.94 & 12k & 374.24 & 22.79 & 5.99 \\
\bottomrule
\end{tabular}%
}
\end{table}

% \begin{table}[t]
% \centering
% \footnotesize
% \caption{Performance analysis on XLRS-Bench under varying resolutions}
% \label{tab:percep_reason_mean}
% \resizebox{0.95\linewidth}{!}{%
% \begin{tabular}{lcccc}
% \toprule
% \textbf{Image Resolution} & \textbf{Kept Tokens} & \textbf{Perception Mean} & \textbf{Reasoning Mean} & \textbf{w.Avg} \\
% \midrule
% 8k & 3k     & 42.43 & 56.93 & 43.00 \\
% 8k & 4k     & 42.50 & 56.64 & 43.12 \\
% 4k & 4k     & 42.56 & 56.93 & 43.28\\
% \bottomrule
% \end{tabular}%
% }
% \end{table}
% Unlabeled & 43.36 & 56.33 \\
% 4k/5k     & 42.88 & 56.73 \\
% 4k/6k     & 43.07 & 56.93 \\
% 4k/8k     & 41.13 & 57.13 \\
% 4k/10k    & 45.32 & 57.13 \\

\begin{figure}[t] % 使用figure环境
\includegraphics[width=\linewidth]{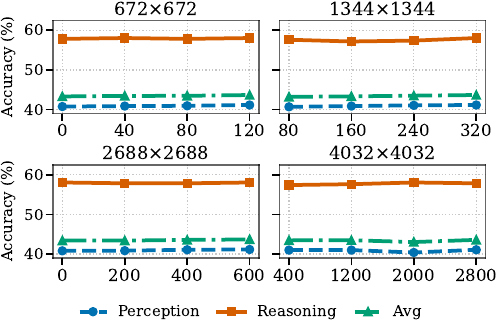}
\caption{Multi-scale token selection accuracy on XLRS-Bench. We evaluate four input resolutions (672$\times$672, 1344$\times$1344, 2688$\times$2688, 4032$\times$4032) and vary the number of retained visual tokens at each scale (x-axis).}
\end{figure}
\begin{figure}[t]
    \centering
    % \begin{subfigure}[t]{0.98\linewidth}
    %     \centering
    %     \includegraphics[width=\linewidth]{Figures/SAM_visual-compar.pdf}
    %     % \caption{Segmentation-guided pruning.}
    %     % \label{fig:sam_compare_sub}
    % \end{subfigure}

    % \vspace{-0.8em}

    \begin{subfigure}[t]{0.98\linewidth}
        \centering
        \includegraphics[width=\linewidth]{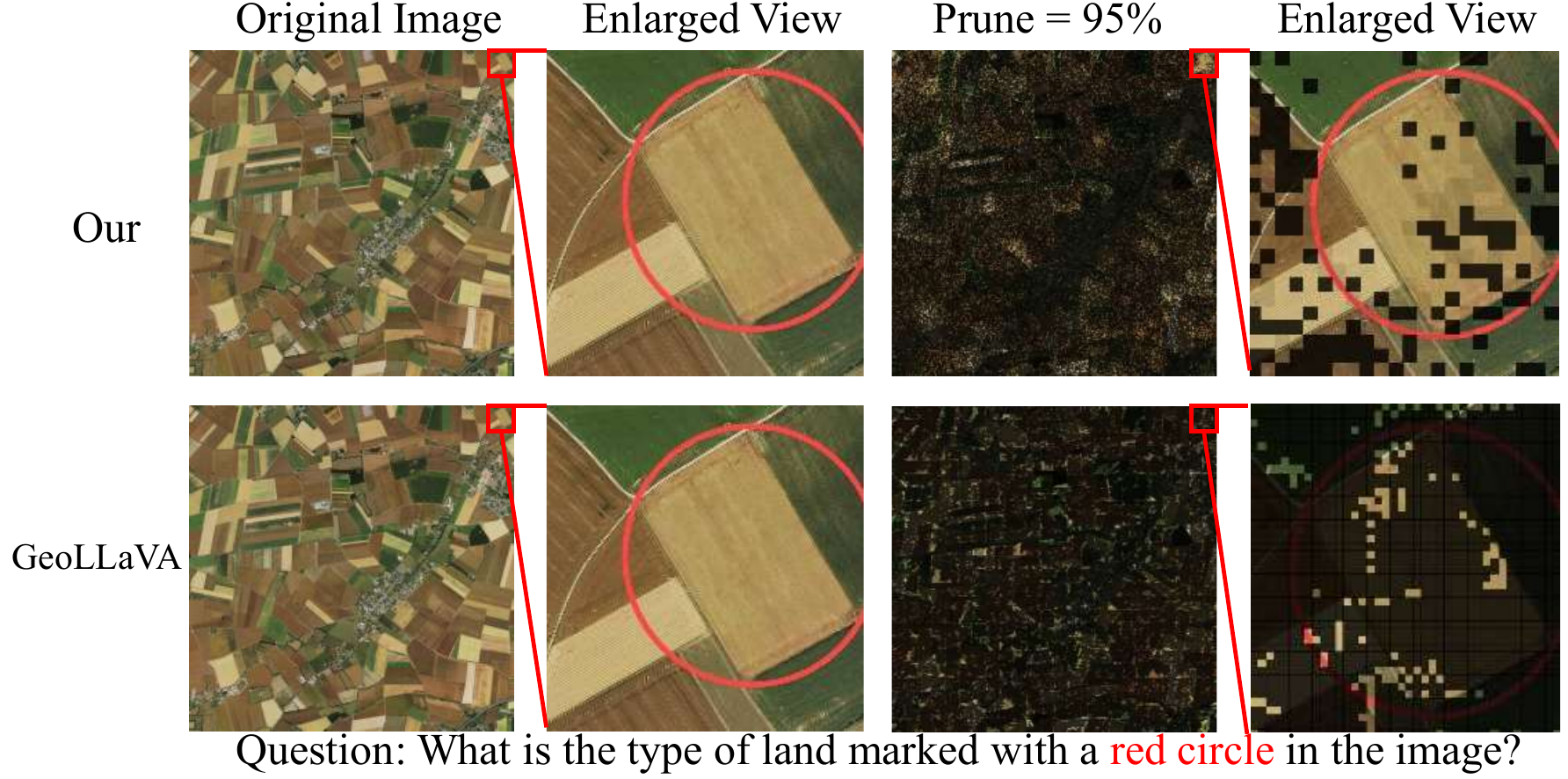}
        % \caption{Clustering-guided pruning.}
        % \label{fig:cluster_visualation_sub}
    \end{subfigure}

    \caption{Qualitative comparisons of our method and GeoLLaVA-8K under 95\% pruning ratio.}
    % \textbf{(Bottom)} Clustering-guided pruning: our clustering-based grouping preserves target objects/areas after pruning, whereas the baseline is more likely to prune away critical evidence, causing missing or fragmented targets.}
    \label{fig:pruning_qual_compare}
\end{figure}

\begin{table}[t]
\centering
\footnotesize
\caption{Performance analysis on XLRS-Bench under varying resolutions and token budgets.}
\label{tab:percep_reason_mean}
\resizebox{0.95\linewidth}{!}{%
\begin{tabular}{ccccc}
\toprule
\textbf{Kept Tokens} & \textbf{Image Resolution} & \textbf{Perception Mean} & \textbf{Reasoning Mean} & \textbf{w.Avg.} \\
\midrule
3k & 8k & 40.38 & 57.83 & 43.00 \\
4k & 8k & 40.61 & 57.52 & 43.12 \\
3k & 4k & 40.50 & 58.04 & 43.12 \\
4k & 4k & 40.73 & 57.83 & 43.28 \\
\bottomrule
\end{tabular}%
}
\end{table}

% \textbf{Impact of Token Filtering Threshold ($K$) across Scales.}
% Table~\ref{tab:k_sweep_by_resolution} presents a quantitative ablation of the attention-only token filtering parameter $K$ across varying input resolutions, ranging from $672\times672$ to $4032\times4032$. 
% We observe a distinct positive correlation between the token retention budget $K$ and downstream performance. Specifically, for resolutions up to $2688\times2688$, increasing $K$ results in a monotonic improvement in both Perception and Reasoning tasks, culminating in a consistent peak Average accuracy of $\approx 43.67\%$. 
% This scale invariance suggests that the model effectively leverages additional token capacity to preserve high-frequency details essential for remote sensing interpretation. 
% However, at the extreme resolution of $4032\times4032$, we observe performance saturation and minor volatility (e.g., at $K=2000$), indicating that while higher token budgets are generally beneficial, the marginal utility diminishes beyond a critical threshold. 
% These results empirically validate the necessity of dynamically scaling $K$ proportional to input resolution to maintain an optimal trade-off between computational efficiency and representational fidelity.
\textbf{Scaling Behavior of Token Selection at Multiple Resolutions.}
Table~\ref{tab:k_sweep_by_resolution} reports the performance on XLRS-Bench across four resolutions, ranging from $672 \times 672$ to $4032 \times 4032$. 
We observe that increasing the token budget $K$ at any specific scale consistently leads to an improvement in accuracy. 
This can be attributed to the synergistic and complementary nature of multi-scale tokens, where low-resolution tokens capture broader visual context, whereas high-resolution tokens provide essential fine-grained details. 
Notably, these performance gains are relatively modest and tend to plateau as $K$ increases. 
This suggests that our method is highly effective at identifying the most informative visual tokens even under stringent budget constraints, regardless of the input resolution.

\begin{table}[t]
\centering
\small
\caption{Performance on XLRS-Bench across multiple resolutions ($672 \times 672$, $1344 \times 1344$, $2688 \times 2688$ and $4032 \times 4032$) under varying scale-specific token budgets $K$.}
\label{tab:k_sweep_by_resolution}

\resizebox{0.45\textwidth}{!}{%
\begin{tabular}{@{}llcccc@{}}
\toprule
\textbf{Resolution} & \textbf{Other scales} & \textbf{$K$} & \textbf{Perception} & \textbf{Reasoning} & \textbf{w.Avg.} \\
\midrule
672$\times$672   & 320,600,2400 & 0   & 40.76 & 57.83 & 43.31 \\
672$\times$672   & 320,600,2400 & 40  & 40.88 & 58.04 & 43.44 \\
672$\times$672   & 320,600,2400 & 80  & 40.95 & 57.83 & 43.47 \\
672$\times$672   & 320,600,2400 & 120 & 41.15 & 58.04 & 43.67 \\
\midrule
1344$\times$1344 & 120,600,2400 & 80  & 40.69 & 57.61 & 43.21 \\
1344$\times$1344 & 120,600,2400 & 160 & 40.88 & 57.17 & 43.33 \\
1344$\times$1344 & 120,600,2400 & 240 & 41.07 & 57.39 & 43.51 \\
1344$\times$1344 & 120,600,2400 & 320 & 41.15 & 58.04 & 43.67 \\
\midrule
2688$\times$2688 & 120,320,2400 & 0   & 40.80 & 58.04 & 43.38 \\
2688$\times$2688 & 120,320,2400 & 200 & 40.84 & 57.83 & 43.38 \\
2688$\times$2688 & 120,320,2400 & 400 & 41.07 & 57.83 & 43.57 \\
2688$\times$2688 & 120,320,2400 & 600 & 41.15 & 58.04 & 43.67 \\
\midrule
4032$\times$4032 & 120,320,600  & 400  & 40.99 & 57.39 & 43.44 \\
4032$\times$4032 & 120,320,600  & 1200 & 40.99 & 57.61 & 43.47 \\
4032$\times$4032 & 120,320,600  & 2000 & 40.38 & 58.04 & 43.02 \\
4032$\times$4032 & 120,320,600  & 2800 & 41.07 & 57.83 & 43.57 \\
\bottomrule
\end{tabular}%
}
\end{table}

\textbf{Qualitative Visualization.}
% As shown in Figure~\ref{fig:pruning_qual_compare}, we provide a qualitative comparison between our method and GeoLLAVA on remote-sensing imagery under a 95\% pruning ratio. 
% Despite the aggressive pruning, our approach effectively preserves semantically meaningful regions critical for downstream tasks. 
% In contrast, GeoLLAVA discards essential visual information, leading to severe degradation in the region of interest (highlighted by the red circle). 
% Our method consistently retains this key area, ensuring the retention of crucial visual features, while GeoLLaVA's pruning results in missing or fragmented content. 
% This comparison underscores the strength of our region-aware pruning strategy, which maintains critical visual evidence even under extreme pruning conditions.
Figure~\ref{fig:pruning_qual_compare} compares our method with GeoLLAVA on remote-sensing imagery at a 95\% pruning ratio. 
While GeoLLAVA indiscriminately discards essential information and causes severe degradation in the region of interest (red circle), our region-aware strategy consistently preserves critical semantic features. 
This demonstrates the robustness of our approach even under extreme pruning conditions. 
Such high-fidelity preservation ensures that the model retains sufficient visual cues for accurate scene interpretation and spatial reasoning. 
% By successfully retaining these fine-grained details, UHR-BAT guarantees the information sufficiency required for the model to perform correct logical reasoning in subsequent stages.

% \textbf{Resolution Robustness Analysis.} To evaluate the impact of input resolution, we downsampled 8K images to 4K for model inference. Empirical results on XLRS-Bench demonstrate that 4K inputs yield performance on par with the original 8K images (Table \ref{tab:percep_reason_mean}). This indicates that our model effectively preserves critical semantic information even at a reduced spatial scale, maintaining high accuracy without the computational overhead of 8K inputs. 
\textbf{Resolution Robustness and Scale Trade-offs.} To evaluate the impact of input resolution, we conducted inference by downsampling 8K images to 4K. 
Empirical results on XLRS-Bench demonstrate that 4K inputs yield performance on par with, and in some cases superior to, the original 8K images (Table \ref{tab:percep_reason_mean}). 
Interestingly, we observe that under an identical token budget, the 4K configuration marginally outperforms the 8K baseline by approximately $0.1\%$. 
This phenomenon can be attributed to the increased contextual coverage per token. 
At lower resolutions, each token encapsulates a broader spatial area, thereby providing the model with richer global semantic cues. 
This suggests that for resource-constrained inference, the benefit of enhanced contextual density in 4K can outweigh the loss of extreme fine-grained pixel details in 8K. 
These findings underscore the robustness of our framework and its ability to maintain high accuracy without the prohibitive computational overhead associated with ultra-high-resolution inputs.
% Consequently, this validates that our framework effectively treats 4K resolution as an optimal operating point for 8K imagery, maximizing semantic information density while significantly mitigating resource demands.
% Moreover, we can observe a positive correlation between token density and model performance from the results.

\section{Conclusion}
In this paper, we present UHR-BAT, a novel budget-aware token compression framework specifically engineered for ultra-high-resolution (UHR) remote sensing understanding.
By integrating query-guided importance estimation with a region-wise preserve-and-merge strategy, our method effectively captures task-relevant details and reduces visual redundancy.
Extensive experiments on the XLRS-Bench and RSHR-Bench demonstrate that UHR-BAT achieves state-of-the-art performance with substantially reduced token budgets, offering a superior trade-off between accuracy and efficiency. 
Our work provides a practical and scalable solution for the deployment of advanced MLLMs in resource-constrained geospatial applications.

\section*{Impact Statement}
UHR-BAT provides a resource-efficient solution for deploying multimodal large language models (MLLMs) in ultra-high-resolution remote sensing scenarios, even under stringent computational and context budgets.
By decoupling input resolution from the quadratic explosion of visual tokens, our framework makes it feasible to process kilometer-scale imagery on commodity hardware or resource-constrained edge platforms.
This democratization of UHR processing capabilities has significant implications for critical real-world applications.
Furthermore, by enforcing a fixed-budget mechanism, our approach inherently reduces the energy consumption associated with large-scale vision-language model inference.

% =========================================================
% 7. 参考文献
% =========================================================
\bibliographystyle{icml2026} % 必须指定 ICML 格式
\bibliography{custom}

% =========================================================
% 8. 附录
% =========================================================
\newpage
\appendix
\onecolumn % 附录通常推荐单栏

\clearpage

\section{Background.}
\label{Motivation}

\paragraph{Resource-Efficient Training and Inference for High-Resolution Remote Sensing.}
The rapid proliferation of high-resolution remote sensing platforms, including satellites, Unmanned Aerial Vehicles (UAVs), and precision agriculture systems has significantly expanded the practical utility of aerial imagery. 
However, deploying state-of-the-art models in these domains is often hindered by stringent hardware constraints. 
Motivated by these real-world challenges, we investigate model training and inference under restricted computational budgets. 
Our framework enables the training of \textbf{8K ultra-high-resolution} images on a single \textbf{NVIDIA RTX A6000 (48GB)}, while maintaining a peak inference memory footprint of approximately \textbf{30GB}. 
To achieve this, we implement a multi-scale fixed-budget strategy, allocating $80, 320, 600,$ and $2000$ tokens for input resolutions of $672 \times 672, 1344 \times 1344, 2688 \times 2688,$ and $4032 \times 4032$, respectively. 
Our approach yields a practical inference latency of roughly \textbf{4 seconds} per image, striking a favorable balance between high-fidelity processing and computational feasibility. 
This performance underscores the potential of our model for broader integration into resource-constrained edge platforms, such as onboard UAV processing systems.

\section{Detailed Experiment Settings.}
\label{Detailed Experiment Settings}

\paragraph{Benchmarks and Metrics.}
We conduct experiments on three widely-used benchmarks to comprehensively evaluate our model's
multimodal capabilities, including general visual understanding, compositional reasoning, OCR-centric
reasoning, robustness in real-world scenarios, and object hallucination behavior: XLRS-Bench~\citep{wang2025xlrs}, MME-Realworld-RS~\citep{mme}, and RSHR-Bench~\citep{dang2025benchmark}.
%LRSVQA~\citep{luo2024lrsvqa}

\paragraph{XLRS-Bench.}
XLRS-Bench is a comprehensive benchmark designed to evaluate multimodal perception and reasoning on \emph{extremely large} ultra-high-resolution remote-sensing imagery.
It features the largest average image size reported to date (approximately $8500\times8500$ pixels) with human-verified annotations, and defines $16$ sub-tasks spanning diverse perceptual skills and higher-level reasoning abilities.
By emphasizing complex semantic relations and spatiotemporal understanding in real-world RS scenes, XLRS-Bench provides a diagnostic testbed for a model's fine-grained recognition and multi-step decision-making on large-scale aerial imagery.

\noindent\textbf{XLRS-Bench Task Abbreviations.}
We report results on the perception and reasoning dimensions using Level-3 VQA capabilities.
For Perception, OC denotes Overall Counting, RC denotes Regional Counting, OLUC denotes Overall Land Use Classification, RLUC denotes Regional Land Use Classification, OCC denotes Object Classification, OCL denotes Object Color, OMS denotes Object Motion State, and OSR denotes Object Spatial Relationship.
For Reasoning, RP denotes Route Planning, AD denotes Anomaly Detection and Interpretation, ECR denotes Environmental Condition Reasoning, CCR denotes Counting with Complex Reasoning, and RCCD denotes Regional Counting with Change Detection.
All tasks use multiple-choice answering, where most sub-tasks adopt A/B/C/D options and OMS adopts a binary A/B format.

\paragraph{MME-RealWorld.}
MME-RealWorld is a large-scale, fully human-annotated benchmark targeting high-resolution, real-world visual scenarios via multiple-choice question answering across multiple domains.
In this work, we adopt its Remote Sensing split (MMERealWorld-RS) to assess robustness under high-resolution overhead imagery, where models must accurately perceive small, dense objects and perform reasoning such as identification and counting in large maps.
The RS split thus complements conventional VQA-style evaluations by stressing real-world difficulty induced by resolution, clutter, and long-range spatial context.

\paragraph{RSHR-Bench.}
RSHR-Bench is a super-high-resolution remote-sensing benchmark designed for
faithful assessment of multimodal visual understanding and complex reasoning at \emph{native} spatial resolutions.
It comprises $5{,}329$ full-scene satellite/aerial/UAV images with long side $\geq 4{,}000$ pixels (up to $\sim300$ megapixels),
sourced from widely used RS corpora and UAV collections.
In this work, we focus on its multiple-choice VQA (MCQ) setting, which evaluates decision-making within a fixed answer space (e.g., A/B/C/D),
and spans Perception, Reasoning, and Multi-turn tasks.
To explicitly reduce reliance on language priors and mitigate guessing effects in the multiple-choice format, RSHR-Bench applies strong LLM adversarial filtering followed by rigorous human verification.

\noindent\textbf{RSHR-Bench Task Abbreviations.}
For Perception, COL denotes Color Detection, SHP denotes Shape Recognition, DET denotes Detection, OC denotes Object Classification, REL denotes Object Spatial Relationship, OGD denotes Object Grounding, RG denotes Regional Grounding, OCN denotes Object Counting, and RCN denotes Regional Counting.
For Reasoning, AD denotes Anomaly (single-turn), FP denotes Future Prediction (multi-image), MRJC denotes Multi-region Joint Contrast (multi-image), MRJCS denotes Multi-region Joint Contrast (single-image, multi-box), and OSJ denotes Object State Judgment (single-turn).
% For Multi-turn, MAD denotes Anomaly, MTFP denotes Future Prediction, and MOSJ denotes Object State Judgment.

% \paragraph{LRS-VQA.}
% LRS-VQA is a VQA benchmark tailored for \emph{large remote-sensing images}, constructed to probe whether vision--language models can effectively perceive and reason over wide-area scenes under limited computation.
% It contains $7{,}333$ QA pairs across $8$ categories, with images reaching up to $27{,}328$ pixels on the long side (average resolution around $7099\times6329$).
% With many questions requiring localized evidence from small regions (e.g., text-related or fine-detail cues) while maintaining global scene awareness, LRS-VQA offers a complementary evaluation of coarse-to-fine understanding and reasoning on large-scale RS imagery.

\paragraph{Baselines.}
As summarized in Table~\ref{tab:visual_reasoning_perception_table2}, we compare against three baseline groups.
Remote-sensing VLMs include EarthDial~\cite{soni2025earthdial}, GeoChat~\cite{geochat}, GeoLLaVA-8K~\cite{wang2025geollava8k}, and VHM~\cite{pang2025vhm}, which are tailored to overhead imagery and geospatially grounded vision--language understanding.
Open-source general-purpose VLMs cover InternVL2.5-8B~\cite{chen2024expanding}, InternVL3.5-8B~\cite{wang2025internvl3}, MiniCPM2\_6~\cite{yao2024minicpm}, Phi-3.5-Vision~\cite{abdin2024phi3}, Qwen2.5-VL-7B~\cite{Qwen2.5-VL}, DeepSeek-VL~\citep{lu2024deepseek}, and VILA-HD~\cite{shi2025scaling}, representing strong off-the-shelf multimodal perception and reasoning.
We also report closed-source VLMs (GPT5~\cite{openai2025gpt5thinking}, GPT-4o / GPT-4o-mini~\cite{hurst2024gpt}, and Gemini-2.5-pro~\cite{comanici2025gemini}) as reference baselines, and for models that support high-resolution inference (e.g., GeoLLaVA-8K and VILA-HD), we follow their recommended settings to handle 4K+ remote-sensing inputs.

\paragraph{More Details of Our Method.}
% We adopt the pre-training approach of GeoLLaVA-8K as a foundation, and then fine-tune our model using only 10K image-text pairs from SuperRS-VQA and HighRs-VQA. 
We initialize our model based on the LLaVA-Next architecture, incorporating CLIP-ViT-L/14-336 as the vision encoder and LongVA-7B as the language backbone. 
The model is subsequently fine-tuned using only 10K image-text pairs sampled from SuperRS-VQA and HighRs-VQA.
During fine-tuning, we utilize four different resolutions—672$\times$672, 1344$\times$1344, 2688$\times$2688, and 4032$\times$4032.
Images are resized and padded to reach these target scales.
We use a batch size of 16, learning rates of $1 \times 10^{-6}$ for the visual components and $5 \times 10^{-6}$ for the projection layers interacting with the LLM, and the use of ZeRO-2 parallelism with a training epoch of 2.
For evaluation, baseline results are sourced from the original dataset papers. 
Regarding our model, in the case of XLRS-Bench, the budget $B_s$ is assigned as $180, 1320, 1600$, and $8000$ corresponding to the four resolutions. 
Similarly, in the case of MMERealWorld-RS and RSHR-Bench, we adopt budgets of $80, 320, 1600$, and $4000$, respectively.

% =========================
% APPENDIX - Implementation Details and Derivations
% =========================

\section{Implementation Details and Derivations}
\label{sec:appendix}

\subsection{Detailed Pseudocode}
\label{sec:algorithm_process}
To efficiently reduce the sequence length while preserving local semantic structures, we introduce the Region-aware Token Pruning mechanism. As detailed in Algorithm \ref{alg:method}, our method first partitions the feature map into disjoint semantic regions to capture spatial priors. We then assess token importance within each region and employ a preserve-and-merge strategy to retain salient information. Finally, we prioritize regions based on their average importance and truncate the sequence to satisfy a predefined computational budget $B_s$.

\begin{algorithm}[h]
\caption{Region-aware Token Pruning}
\label{alg:method}
\begin{algorithmic}[1]
\REQUIRE Input feature map $\mathbf{X}$, Number of regions $k$, Pruning ratio $\rho$.
\ENSURE Pruned tokens $\mathbf{X}_{out}$.

\STATE \textbf{// Step 1: Region Partition}
\STATE Construct clustering embeddings $\mathbf{u}_i$ for all tokens via Eq.~\eqref{eq:kmeans-embed}.
\STATE Obtain regions $\{\mathcal{S}_1, \dots, \mathcal{S}_k\}$ by minimizing Eq.~\eqref{eq:kmeans-assign} (K-means) or via Eq.~\eqref{eq:sam-vote} (SAM).

\STATE \textbf{// Step 2: Importance Assessment}
\STATE Compute the attention score map $A^{(s)}_{u,v}$ for each scale according to Eq.~\eqref{eq:attn-align}.
\STATE Derive the individual token importance scores $\{a_i^{(s)}\}_{i=1}^{N_s}$ following Eq.~\eqref{eq:token-attn}.
\STATE Calculate the average importance score $\{\mu_m^{(s)}\}_{m=1}^{R_s}$ for each region as defined in Eq.~\eqref{eq:region-mean-attn}.
% \STATE Compute importance scores $\mathbf{a} \in \mathbb{R}^N$ (e.g., using attention weights w.r.t. the [CLS] token).

% \STATE \textbf{// Step 3: Region-wise Selection}
% \STATE Initialize output set $\mathbf{X}_{out} \leftarrow \emptyset$.
% \FOR{each region $\mathcal{S}_r \in \{\mathcal{S}_1, \dots, \mathcal{S}_k\}$}
%     \STATE Calculate the number of tokens to keep $N_r = \lceil \rho \cdot |\mathcal{S}_r| \rceil$.
%     \STATE Identify the top-$N_r$ tokens with highest scores in $\mathcal{S}_r$ as set $\mathcal{K}_r$.
%     \STATE $\mathbf{X}_{out} \leftarrow \mathbf{X}_{out} \cup \{ \mathbf{x}_i \mid i \in \mathcal{K}_r \}$. \quad 
% \ENDFOR
\STATE \textbf{// Step 3: Region-wise Preserve, Merge, and Priority Budgeting}
\STATE Initialize candidate sequence $E_{\mathrm{cand}}^{(s)} \leftarrow [\,]$
\FOR{each region $\mathcal{S}_m^{(s)} \in \mathcal{P}^{(s)}$}
    \STATE $\mu_m^{(s)} \leftarrow \frac{1}{|\mathcal{S}_m^{(s)}|} \sum_{i \in \mathcal{S}_m^{(s)}} a_i^{(s)}$
    \STATE $\mathcal{K}_m^{(s)} \leftarrow \{i \in \mathcal{S}_m^{(s)} \mid a_i^{(s)} \ge \mu_m^{(s)}\}$
    \STATE $\mathcal{R}_m^{(s)} \leftarrow \mathcal{S}_m^{(s)} \setminus \mathcal{K}_m^{(s)}$
    \IF{$|\mathcal{R}_m^{(s)}| > 0$}
        \STATE $\tilde{\mathbf{e}}_m^{(s)} \leftarrow \text{MeanPool}(\{\mathbf{h}_i^{(s)} \mid i \in \mathcal{R}_m^{(s)}\})$
    \ENDIF
\ENDFOR
\STATE Sort region indices $\pi$ by descending average importance $\mu_m^{(s)}$
\FOR{$r = 1$ \textbf{to} $R_s$}
    \STATE $m \leftarrow \pi(r)$
    \STATE Sort tokens in $\mathcal{K}_m^{(s)}$ by $a_i^{(s)}$ in descending order
    \STATE Append sorted tokens in $\mathcal{K}_m^{(s)}$ to $E_{\mathrm{cand}}^{(s)}$
    \IF{$\tilde{\mathbf{e}}_m^{(s)}$ exists}
        \STATE Append $\tilde{\mathbf{e}}_m^{(s)}$ to $E_{\mathrm{cand}}^{(s)}$
    \ENDIF
\ENDFOR
\STATE $\bar{E}^{(s)} \leftarrow E_{\mathrm{cand}}^{(s)}[1 : B_s]$ \COMMENT{Truncate to budget $B_s$}
\STATE \textbf{return} $\bar{E}^{(s)}$

% \STATE \textbf{return} $\mathbf{X}_{out}$ 
\end{algorithmic}
\end{algorithm}
\subsection{Region Partition Instantiations}
\label{sec:appendix_partition}

% \textbf{Feature+coordinate K-means partition.}
% Let $\mathbf{f}_i\in\mathbb{R}^{D}$ be the visual feature associated with token $i$ and $(x_i,y_i)\in[0,1]^2$ be its normalized center coordinate.
% We $\ell_2$-normalize features and concatenate spatial coordinates to form the clustering embedding $\mathbf{u}_i \in \mathbb{R}^{D+2}$:
% \begin{equation}
% \hat{\mathbf{f}}_i=\frac{\mathbf{f}_i}{\|\mathbf{f}_i\|_2},
% \qquad
% \mathbf{u}_i=\big[ \lambda_f \hat{\mathbf{f}}_i^\top, \ \lambda_{xy}x_i, \ \lambda_{xy}y_i \big]^\top.
% \label{eq:kmeans-embed}
% \end{equation}
% K-means with $k$ centers minimizes the objective $J$ to find centroids $\{\bm{\mu}_r\}_{r=1}^{k}$:
% \begin{equation}
% J=\sum_{i=1}^{N}\|\mathbf{u}_i-\bm{\mu}_{c_i}\|_2^2,
% \qquad
% c_i=\arg\min_{1\le r\le k}\|\mathbf{u}_i-\bm{\mu}_r\|_2^2,
% \label{eq:kmeans-assign}
% \end{equation}
% yielding clusters $\{\mathcal{S}_r\}_{r=1}^{k}$ as the partition $\mathcal{P}^{(s)}$ (thus $R_s=k$ for this instantiation).
\textbf{Feature+coordinate K-means Partition.}
Let $\mathbf{f}_i\in\mathbb{R}^{D}$ be the visual feature associated with token $i$ and $(x_i,y_i)\in[0,1]^2$ be its normalized center coordinate. 
Distinct from natural images, remote sensing scenes often exhibit strong spatial correlation, where proximal tokens are highly likely to represent the same physical object or homogeneous region. 
To exploit this characteristic, we incorporate the spatial coordinates of tokens into the clustering process to identify semantic redundancy by imposing explicit spatial constraints on clustering.
Specifically, we $\ell_2$-normalize the features and concatenate these scaled coordinates to form the clustering embedding $\mathbf{u}_i \in \mathbb{R}^{D+2}$:
\begin{equation}
\hat{\mathbf{f}}_i=\frac{\mathbf{f}_i}{\|\mathbf{f}_i\|_2},
\qquad
\mathbf{u}_i=\big[ \lambda_f \hat{\mathbf{f}}_i^\top, \ \lambda_{xy}x_i, \ \lambda_{xy}y_i \big]^\top.
\label{eq:kmeans-embed}
\end{equation}
K-means~\citep{mcqueen1967some,hartigan1979algorithm,lloyd1982least} with $k$ centers iteratively partitions the input space and minimizes the objective $J$ to find centroids $\{\bm{\mu}_r\}_{r=1}^{k}$:
\begin{equation}
J=\sum_{i=1}^{N}\|\mathbf{u}_i-\bm{\mu}_{c_i}\|_2^2,
\qquad
c_i=\arg\min_{1\le r\le k}\|\mathbf{u}_i-\bm{\mu}_r\|_2^2,
\label{eq:kmeans-assign}
\end{equation}
yielding clusters $\{\mathcal{S}_r\}_{r=1}^{k}$ as the partition $\mathcal{P}^{(s)}$. 
By integrating spatial distance, the clustering process effectively groups tokens that are both semantically similar and geographically close. 
During token selection, we merge these redundant tokens within each cluster, thereby condensing the representation while preserving the essential land-cover structures.

\textbf{SAM-induced Partition and Token Mapping.} 
To identify tokens that represent the same semantic information, we employ a frozen SAM model to partition the image into distinct regions. To this end, we perform a token-level mapping of the pixel-level segmentation results. Specifically, given pixel-level labels from a segmentation model (e.g., SAM~\citep{kirillov2023segment}), we map them to token regions by majority vote over the pixel set $\Omega_i$ covered by token $i$:
\begin{equation}
\mathrm{label}_{\mathrm{token}}(i)
=
\arg\max_{c}\sum_{(x,y)\in \Omega_i} \mathbb{I}\big[\mathrm{label}(x,y)=c\big].
\label{eq:sam-vote}
\end{equation}
Tokens with the same token-level label form a region set $\mathcal{S}_m$, thereby effectively producing the final spatial partition $\mathcal{P}^{(s)}$. 
Ties encountered in this process are broken deterministically (e.g., by choosing the smallest label id).

\subsection{Spatial Continuity and Bilinear Interpolation}
\label{sec:appendix_interp}

To bridge the resolution discrepancy between the coarse importance scores and the target high-resolution feature maps, we employ bilinear interpolation to derive spatially aligned importance values. 

Formally, let $G$ be a discrete grid-valued map representing the low-resolution importance scores. For any continuous coordinate $(x,y) \in \mathbb{R}^2$ on the target high-resolution grid, we define its four surrounding integer neighbors on the source grid as $x_0=\lfloor x\rfloor, x_1=x_0+1, y_0=\lfloor y\rfloor,$ and $y_1=y_0+1$. 

Let $Q_{mn}=G(x_m,y_n)$ denote the sampled importance values at these grid corners for $m,n \in\{0,1\}$. The interpolation operator $\Psi$ computes the refined importance value at the high-resolution coordinate as:
\begin{equation}
\begin{aligned}
\Psi(G,(x,y)) = \;& Q_{00}(x_1-x)(y_1-y) + Q_{10}(x-x_0)(y_1-y) \\
& + Q_{01}(x_1-x)(y-y_0) + Q_{11}(x-x_0)(y-y_0).
\end{aligned}
\label{eq:bilinear}
\end{equation}

\subsection{Basic Properties of the Preserve-and-Merge Rule}
\label{sec:appendix_properties}
% In ultra high resolution (UHR) remote sensing, task critical evidence often manifests as sparse, pixel scale targets, such as small vessels or vehicles, that occupy a negligible fraction of the multi million pixel input. 
% While the statistical existence of elements exceeding the mean importance is a trivial certainty, its explicit algorithmic enforcement is vital for UHR scenarios. 
% Traditional global pruning mechanisms, which prioritize tokens based on a global attention ranking, frequently suffer from a regional token vacuum in task relevant sectors where global attention weights are overshadowed by dominant land use layouts.
% To address this, our region wise strategy mandates a minimum representation for every semantically coherent region. 
% We formalize this as a structural coverage guarantee and provide an upper bound on the feature representation error.

\textbf{Lemma 1 (Structural Coverage and Representation Bound).} 
% For any semantically coherent region $\mathcal{S}_m^{(s)}$ with feature variance $\sigma_m^2 = \frac{1}{|\mathcal{S}_m^{(s)}|} \sum_{i \in \mathcal{S}_m^{(s)}} \|h_i^{(s)} - \bar{h}_m^{(s)}\|^2$, the proposed preserve and merge rule ensures that each selected region contributes at least one high salience token such that the retained set defined in Eq.~\eqref{eq:keep-set} is guaranteed to be non-empty, i.e., $\mathcal{K}_m^{(s)}\neq \emptyset$, while the local reconstruction error $\epsilon_m$ incurred by the merged tokens is strictly bounded by the intra region feature dispersion.
For any given semantically coherent region $\mathcal{S}_m^{(s)}$ with intrinsic feature variance $\sigma_m^2 = \frac{1}{|\mathcal{S}_m^{(s)}|} \sum_{i \in \mathcal{S}_m^{(s)}} \|h_i^{(s)} - \bar{h}_m^{(s)}\|^2$, the proposed preserve and merge rule theoretically ensures that each selected region contributes at least one high salience token such that the retained set defined in Eq.~\eqref{eq:keep-set} is guaranteed to be non-empty, i.e., $\mathcal{K}_m^{(s)}\neq \emptyset$, while simultaneously, the local reconstruction error $\epsilon_m$ incurred by the merged tokens is strictly bounded by the intra region feature dispersion.

\textbf{Proof.} The non empty property follows from the definition of the regional mean $\mu_m^{(s)}$: if $\forall i \in \mathcal{S}_m^{(s)}, a_i^{(s)} < \mu_m^{(s)}$, then $\sum a_i^{(s)} < |\mathcal{S}_m^{(s)}| \mu_m^{(s)}$, which contradicts the formulation of average importance. 
This ensures that critical evidence in every partitioned sector is prioritized at the algorithmic level rather than being marginalized by global noise.

To prove the error bound, let the compressed representation of $\mathcal{S}_m^{(s)}$ be $\hat{E}_m = \{h_i^{(s)} : i \in \mathcal{K}_m^{(s)}\} \cup \{\tilde{e}_m^{(s)}\}$. Since $\tilde{e}_m^{(s)}$ is the centroid of the merged subset $\mathcal{R}_m^{(s)}$, the local reconstruction error $\epsilon_m$ (measured by the sum of squared distances) is:
\begin{equation}
\epsilon_m = \sum_{i \in \mathcal{R}_m^{(s)}} \|h_i^{(s)} - \tilde{e}_m^{(s)}\|^2 = |\mathcal{R}_m^{(s)}| \cdot \text{Var}(\mathcal{R}_m^{(s)})
\end{equation}
Because the region $\mathcal{S}_m^{(s)}$ is constructed via semantic clustering, tokens within $\mathcal{S}_m^{(s)}$ exhibit high homogeneity. By selectively preserving high salience tokens $\mathcal{K}_m^{(s)}$ and merging the remaining redundant background tokens into their local centroid, we minimize the semantic shift. The total error is thus strictly constrained by the compactness of the semantic partition, ensuring that the compressed sequence remains region faithful even under extreme budget constraints.

\textbf{Lemma 2 (First-moment preservation on the merged subset).}
For any region with a non-empty merging set $|\mathcal{R}_m^{(s)}|>0$, the merged token $\tilde{\mathbf{e}}_m^{(s)}$ defined in Eq.~\eqref{eq:merge-token} preserves the first moment (the centroid) of the features within $\mathcal{R}_m^{(s)}$:
\begin{equation}
\sum_{i\in \mathcal{R}_m^{(s)}} \mathbf{h}_i^{(s)} = |\mathcal{R}_m^{(s)}|\tilde{\mathbf{e}}_m^{(s)}.
\label{eq:first-moment}
\end{equation}

\textbf{Proof.}
By definition, the merged representation is the arithmetic mean of the features in the merging set: $\tilde{\mathbf{e}}_m^{(s)} = \frac{1}{|\mathcal{R}_m^{(s)}|} \sum_{i\in \mathcal{R}_m^{(s)}} \mathbf{h}_i^{(s)}$. Multiplying both sides by the cardinality $|\mathcal{R}_m^{(s)}|$ directly yields Eq.~\eqref{eq:first-moment}, confirming that the global feature sum is invariant under the merging operation.

\noindent\paragraph{Effect of Cluster Count ($k$).}
Table~\ref{tab:k_sweep} shows that the model is highly robust to the number of clusters, with performance fluctuations remaining within 1\%. 
We observe that at $k=600$, which matches the average granularity of our segmentation-based partition ($\approx 600$), the model achieves an average accuracy of 41.4\%. This result is nearly identical to the peak performance, demonstrating that our clustering mechanism effectively merges redundant visual tokens and captures structural information comparable to explicit image segmentation.
% Table~\ref{tab:k_sweep} investigates the efficacy of the proposed clustering-based token aggregation by varying the number of clusters $k$. The results demonstrate a clear convex trajectory: performance steadily improves as $k$ increases from 10 to 600, indicating that an overly sparse representation fails to preserve fine-grained semantic information. The model achieves global optimality at $k=600$ (41.4\% w.Avg), effectively balancing local feature preservation for Perception tasks with semantic abstraction for Reasoning tasks. Beyond this threshold ($k \ge 720$), we observe a performance degradation, likely attributable to feature over-fragmentation which disrupts the continuity of visual semantics required for complex reasoning.

\begin{table}[h]
\centering
\footnotesize
\caption{Ablation study: Accuracy (\%) across Perception and Reasoning tasks under different cluster counts ($k$).}
\label{tab:k_sweep}
\begin{tabular}{c|ccccccccc}
\toprule
\textbf{Metric} & \textbf{10} & \textbf{50} & \textbf{120} & \textbf{240} & \textbf{360} & \textbf{480} & \textbf{600} & \textbf{720} & \textbf{1000} \\
\midrule
Perception & 37.2 & 37.6 & 37.7 & 37.3 & 37.4 & 37.3 & 37.5 & 37.1 & 37.2 \\
Reasoning  & 62.4 & 64.1 & 63.5 & 62.4 & 63.3 & 63.7 & 63.9 & 63.3 & 63.0 \\
Average    & 40.9 & 41.6 & 41.6 & 41.1 & 41.3 & 41.2 & \textbf{41.4} & 41.0 & 41.1 \\
\bottomrule
\end{tabular}
\end{table}

\noindent\paragraph{Robustness to Region Partitioning Methods.} 
To demonstrate that our framework is agnostic to the specific choice of region partitioning strategy, we evaluate its performance across diverse algorithms, including both clustering-based methods and the Segment Anything Model (SAM). 
Table~\ref{tab:clustering-compare} compares the results of k-means, BIRCH, and SAM on XLRS-Bench across various computational budgets. 
The empirical results indicate that the choice of partitioning technique has no significant impact on the final performance, with the highly efficient clustering methods achieving results comparable to the much more computationally intensive SAM. 
This consistency highlights the robustness of our approach and underscores its flexibility in utilizing different spatial priors without compromising accuracy.

While our framework can utilize either the Segment Anything Model (SAM) or clustering-based methods for region partitioning, we prioritize the latter in practical applications due to the prohibitive preprocessing latency associated with SAM.
This efficiency gain enables our approach to be seamlessly extended to a broader range of latency-critical scenarios, particularly those requiring real-time inference and high-throughput processing.

\begin{table*}[htbp]
\centering
\footnotesize

\begin{minipage}[t]{0.48\textwidth}
\centering
\captionof{table}{Evaluation of various region partitioning methods.}
\label{tab:clustering-compare}
\resizebox{0.9\linewidth}{!}{%
\begin{tabular}{l|ccccc}
\toprule
\textbf{Method} & 4k & 5k & 6k & 8k & 1w \\
\midrule
BIRCH   & 43.05 & 43.28 & 43.28 & 43.31 & 43.51 \\
K-means & 43.34 & 43.57 & 41.43 & 43.73    & 43.31    \\
SAM     & 43.28 & 43.54 & 43.54 & 43.57 & 43.67\\
\bottomrule
\end{tabular}%
}
\end{minipage}\hfill
\begin{minipage}[t]{0.48\textwidth}
\centering
\captionof{table}{Performance on XLRS-Bench across different resolutions.}
\label{tab:context_sweep}
\resizebox{0.9\linewidth}{!}{%
\begin{tabular}{c c c c}
\toprule
\textbf{Context} & Qwen2.5VL-7B & InternVL-3.5 & Phi-3.5 \\
\midrule
2K & \textbf{38.08} & 38.41 & 37.08 \\
4K & \textbf{40.58} & 38.60 & 37.11 \\
8K & \textbf{40.36} & 39.80 & 37.90 \\
\bottomrule
\end{tabular}%
}
\end{minipage}

\end{table*}

\noindent\textbf{Impact of Context Length and Resolution Efficiency.}
Table~\ref{tab:context_sweep} ablates the performance of various MLLMs across different context lengths on XLRS-Bench.
While scaling from 2K to 4K tokens yields substantial gains (e.g., Qwen2.5VL-7B improves from 38.08\% to 40.58\%),
further extending the context to 8K offers negligible improvement or even slight regression (40.36\%).
This plateau indicates that the semantic information density required for XLRS-Bench is effectively saturated at the 4K resolution level.
Consequently, processing 8K native inputs within a compressed 4K context window preserves critical visual details while significantly reducing computational overhead.
Based on this observation, we adopt the 4K setting as the optimal operating point, prioritizing token efficiency without compromising downstream accuracy.

\begin{figure}[h]
    \centering
    \includegraphics[width=0.9\linewidth]{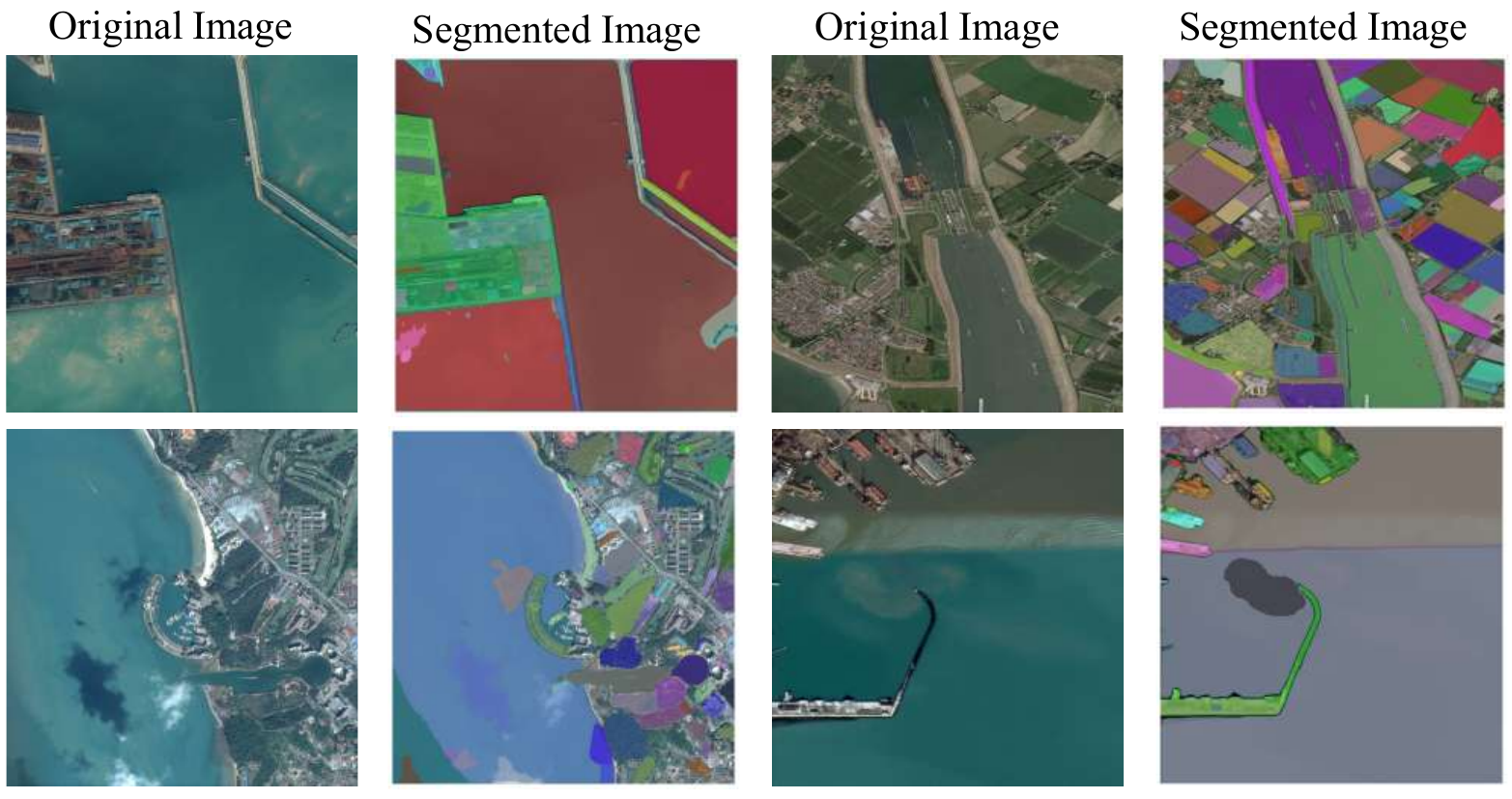}
    \caption{Qualitative examples of coarse segmentation on remote-sensing images produced by our segmentation module, used to form semantically coherent regions for region-aware processing.}
    \label{fig:sam_visualization}
\end{figure}

\begin{figure}[h]
\centering
\includegraphics[width=0.9\linewidth]{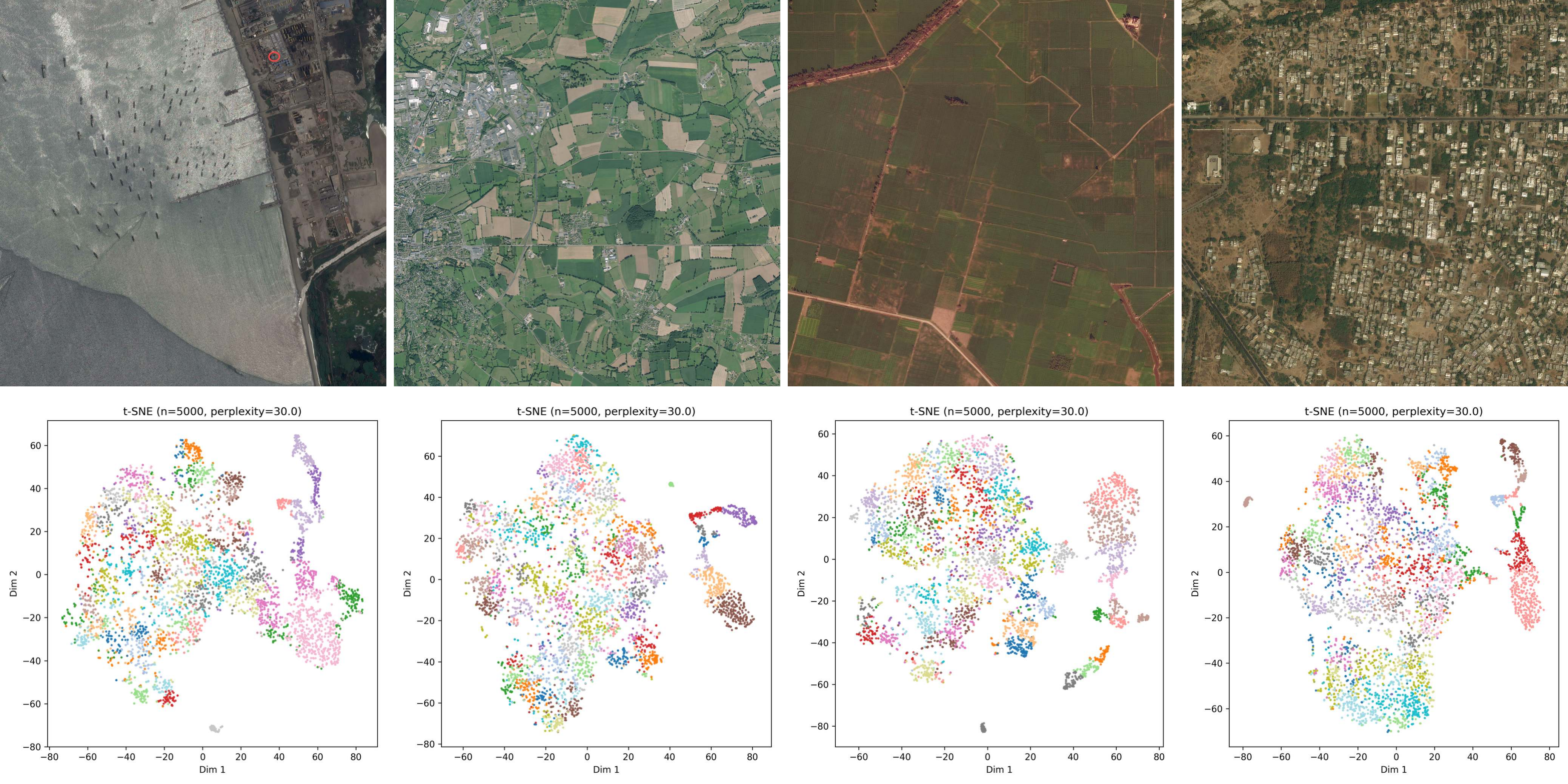}
\caption{t-SNE visualization of visual tokens. Each color represents a cluster assigned by our region-based grouping module. Although the high-dimensional nature of tokens leads to some spatial overlap in the 2D projection, tokens with the same semantic labels exhibit clear clustering behavior, validating the redundancy of visual information and the feasibility of our token merging strategy.}
\label{fig:tsne_analysis}
\end{figure}

\paragraph{Visualization of Region Partition.}
% We provide qualitative examples of the region partition in Figure~\ref{fig:sam_visualization} to illustrate the intermediate results of our framework.
% The primary goal of this module is not to generate pixel-level accurate semantic masks, but rather to decompose the high-resolution image into distinct spatial groups.
% As shown in the figure~\ref{fig:sam_visualization}, the SAM-based method divides the image into multiple local regions based on feature similarity.
% This partition provides a structural basis for the subsequent token compression step.
% It allows us to process the image in a region-wise manner instead of a global manner.
% Even if the segmentation boundaries are coarse or contain noise, the partition successfully separates the large background areas from the complex foreground details.
% This spatial grouping ensures that our method can evaluate token importance within each local scope.
% Consequently, we can preserve the diversity of the visual information across the image and avoid the domination of a single large object, regardless of the precise quality of the segmentation.
We provide qualitative examples of the region partition in Figure~\ref{fig:sam_visualization} to illustrate the intermediate results of our framework. 
The primary goal of this module is not to generate pixel-level accurate semantic masks, but rather to decompose the high-resolution image into distinct spatial groups. 
As shown in Figure~\ref{fig:sam_visualization}, the SAM-based method divides the image into multiple local regions based on feature similarity. 
Qualitative inspection reveals that the resulting partitions generally exhibit high visual homogeneity, effectively encapsulating consistent semantic information within each region. 
Conversely, to account for the background which often encompasses diverse and heterogeneous visual constituents, we treat each token within the background as an independent mask category.
This partition provides a structural basis for the subsequent token compression step, allowing us to process the image in a region-wise manner instead of a global manner. 
Even if the segmentation boundaries are coarse or contain noise, the partition successfully separates the large background areas from the complex foreground details. 
This spatial grouping ensures that our method can evaluate token importance within each local scope; consequently, we can preserve the diversity of the visual information across the image and avoid the domination of a single large object, regardless of the precise quality of the segmentation.

\paragraph{t-SNE Visualization of K-means Clustering Results.} To further investigate the rationale behind our fixed-budget compression, we employ t-SNE to project the high-dimensional visual tokens into a 2D plane (see Figure~\ref{fig:tsne_analysis}). 
While the projection exhibits a degree of proximity between clusters—inherent to the complexity of high-resolution remote sensing manifolds—it is evident that tokens belonging to the same cluster (denoted by color) are significantly aggregated in the latent space. 
This \textit{semantic proximity} confirms that a substantial number of tokens convey nearly identical visual information. 
By identifying and merging these redundant tokens within each semantic cluster, we can drastically reduce the computational overhead without sacrificing critical structural details, thereby enabling efficient inference under a fixed-budget constraint.

\begin{figure*}[t]
    \centering
    \includegraphics[width=0.9\linewidth,keepaspectratio]{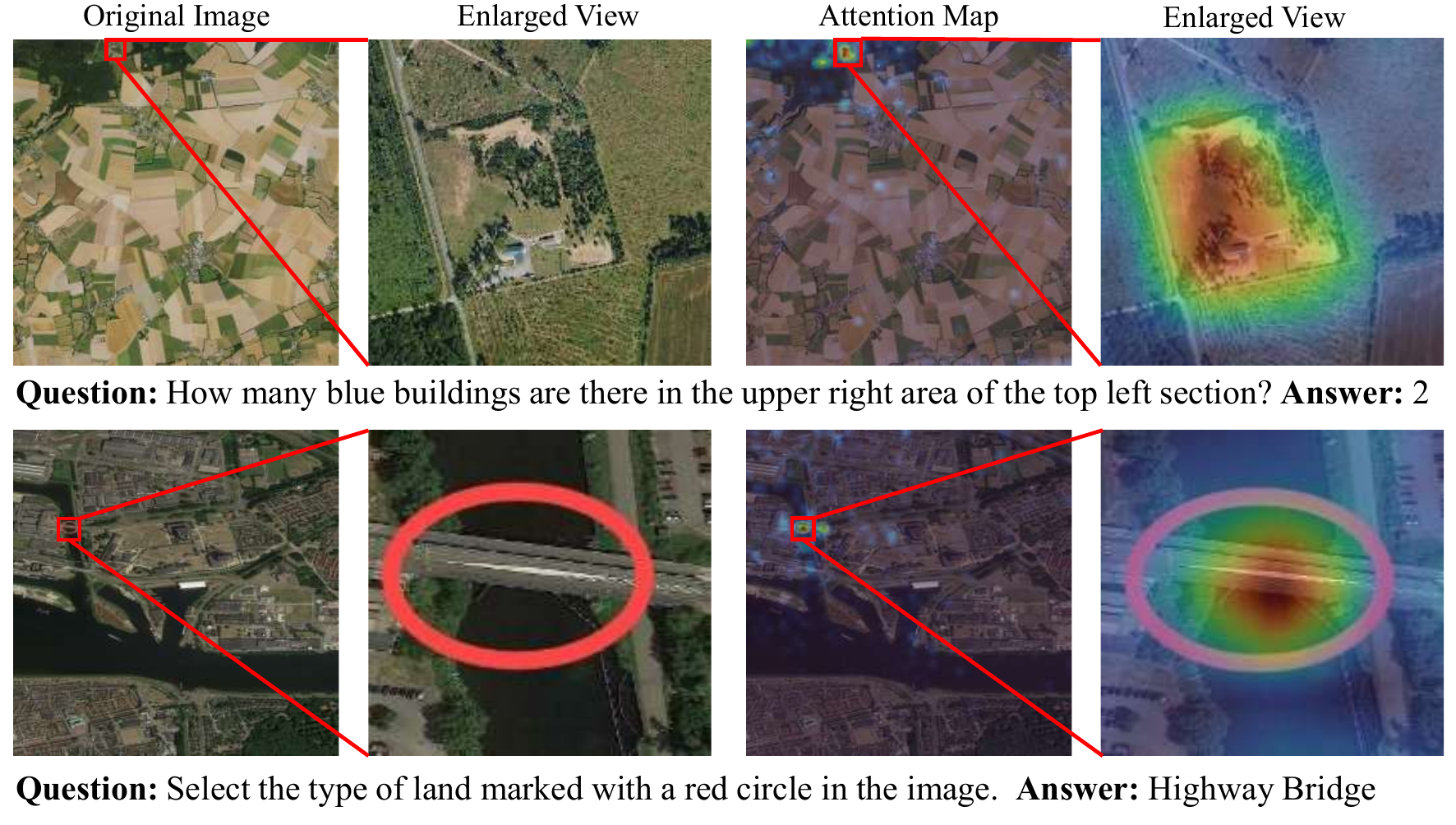}
    \vspace{0.8em}
    \includegraphics[width=0.9\linewidth,keepaspectratio]{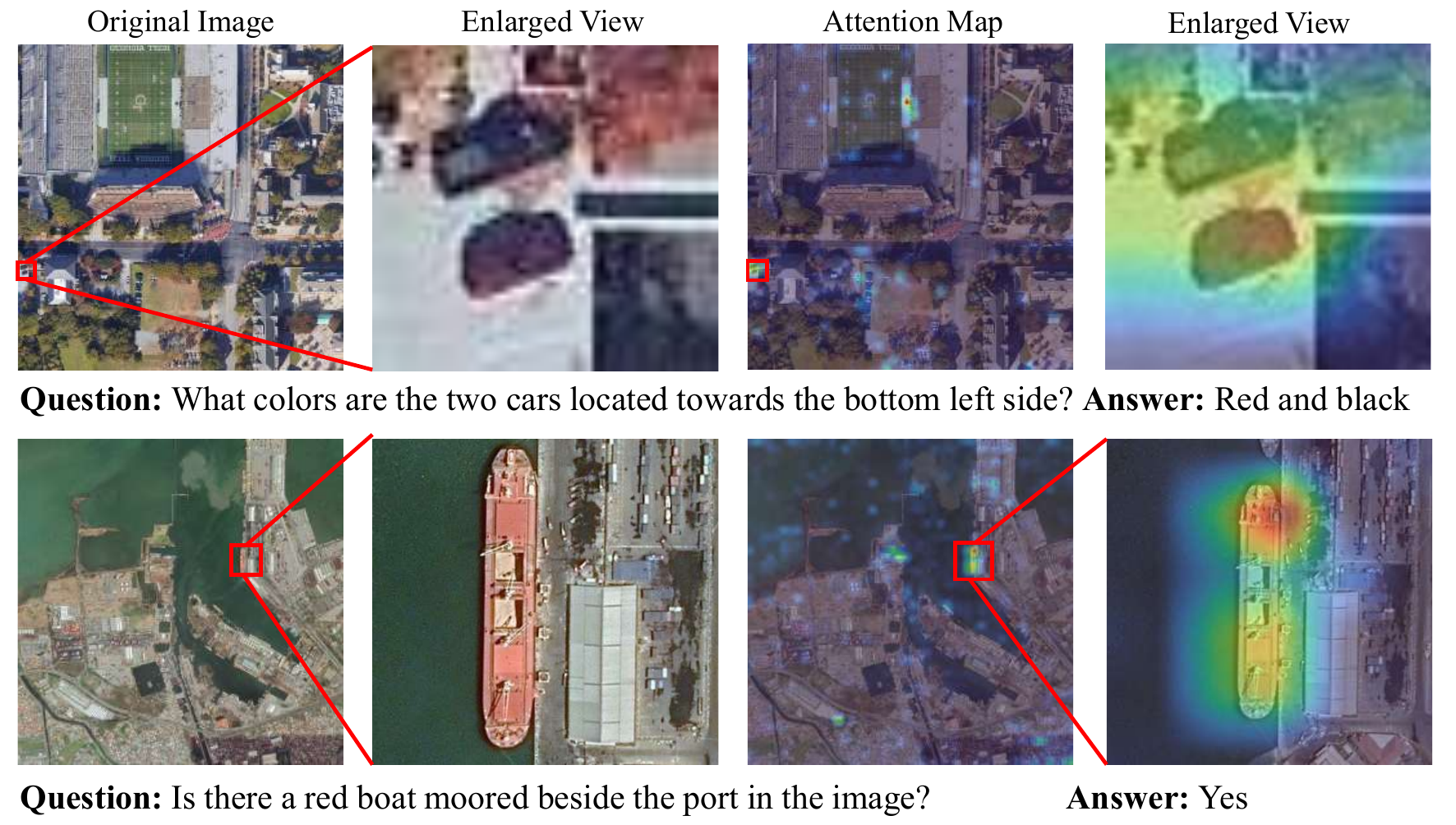}
    
    \caption{Qualitative visualization of region-aware attention on remote-sensing images. We display original images and enlarged views alongside their corresponding attention maps. The heatmaps demonstrate that the model accurately focuses on semantically relevant regions—such as specific vehicles or infrastructure—aligned with the textual query, while assigning lower importance to the background.}
    \label{fig:attention_visualization}
\end{figure*}

\paragraph{Visualization of Query-Guided Attention.}
We visualize the text-to-image attention maps in Figure~\ref{fig:attention_visualization} to validate the reliability of our importance assessment.
These visualizations demonstrate a precise spatial alignment between the textual queries and the visual activations.
As evidenced in the enlarged views, the model accurately localizes the specific objects described in the questions.
For instance, when querying regarding minute details like car colors or a moored boat, the attention mechanism sharply focuses on the target pixels.
Simultaneously, it effectively suppresses the activations of irrelevant background areas or non-target objects.
This selective activation is crucial for our method.
It indicates that the cross-modal attention successfully filters out noise.
Consequently, using these scores for token pruning ensures that the system retains only the critical semantic evidence required for accurate reasoning.

\begin{figure*}[t]
    \centering
    \includegraphics[width=1\linewidth]{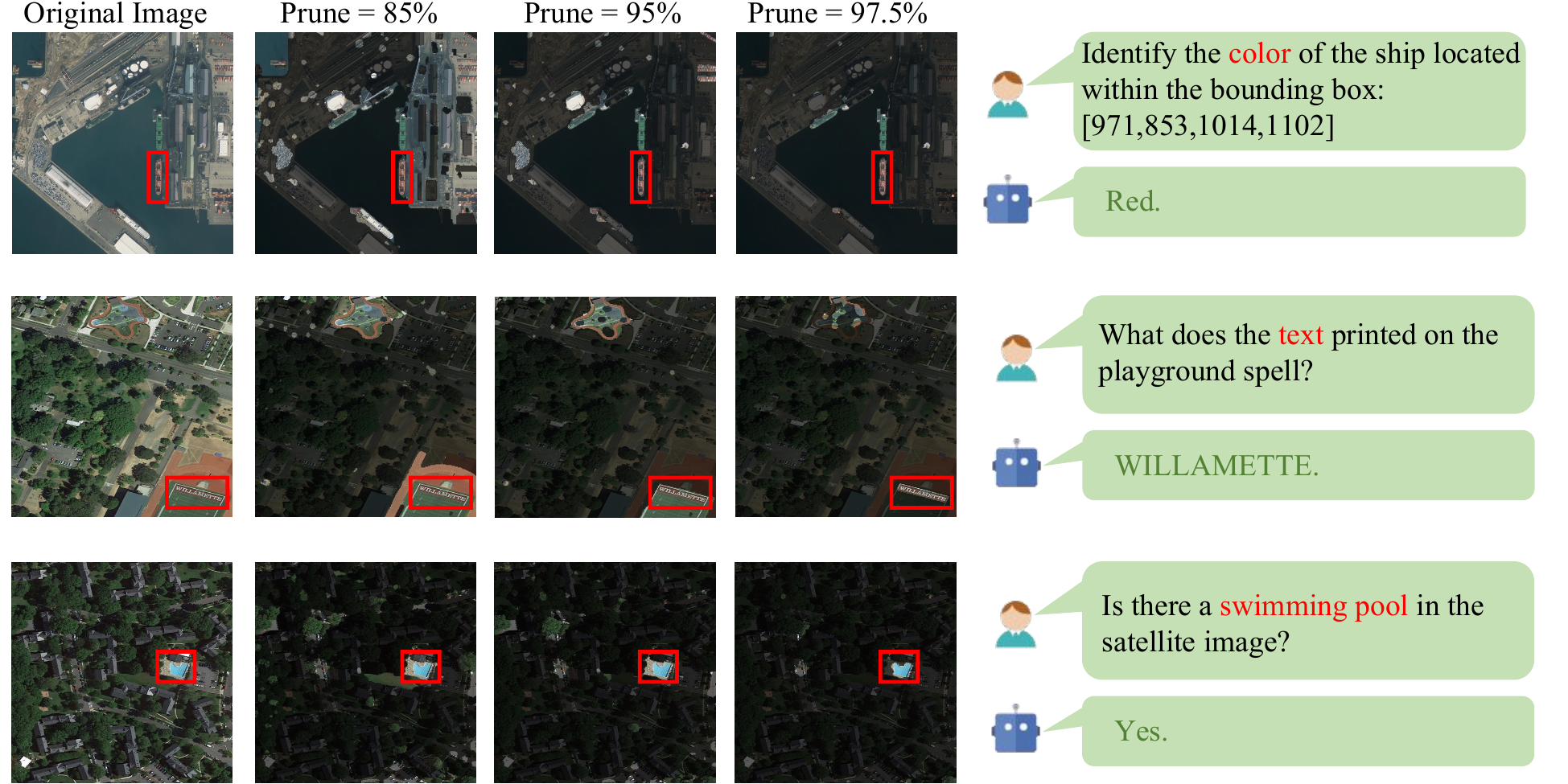}
    \caption{Visualization of attention-guided token pruning. We display the original image and the retained visual tokens at increasing pruning ratios. The method effectively focuses on high-attention areas, preserving query-relevant details (red boxes) like specific colors or text, while discarding the non-informative background.}
    \label{fig:attn_prune}
\end{figure*}

\begin{figure*}[t]
    \centering
    \includegraphics[width=1\linewidth]{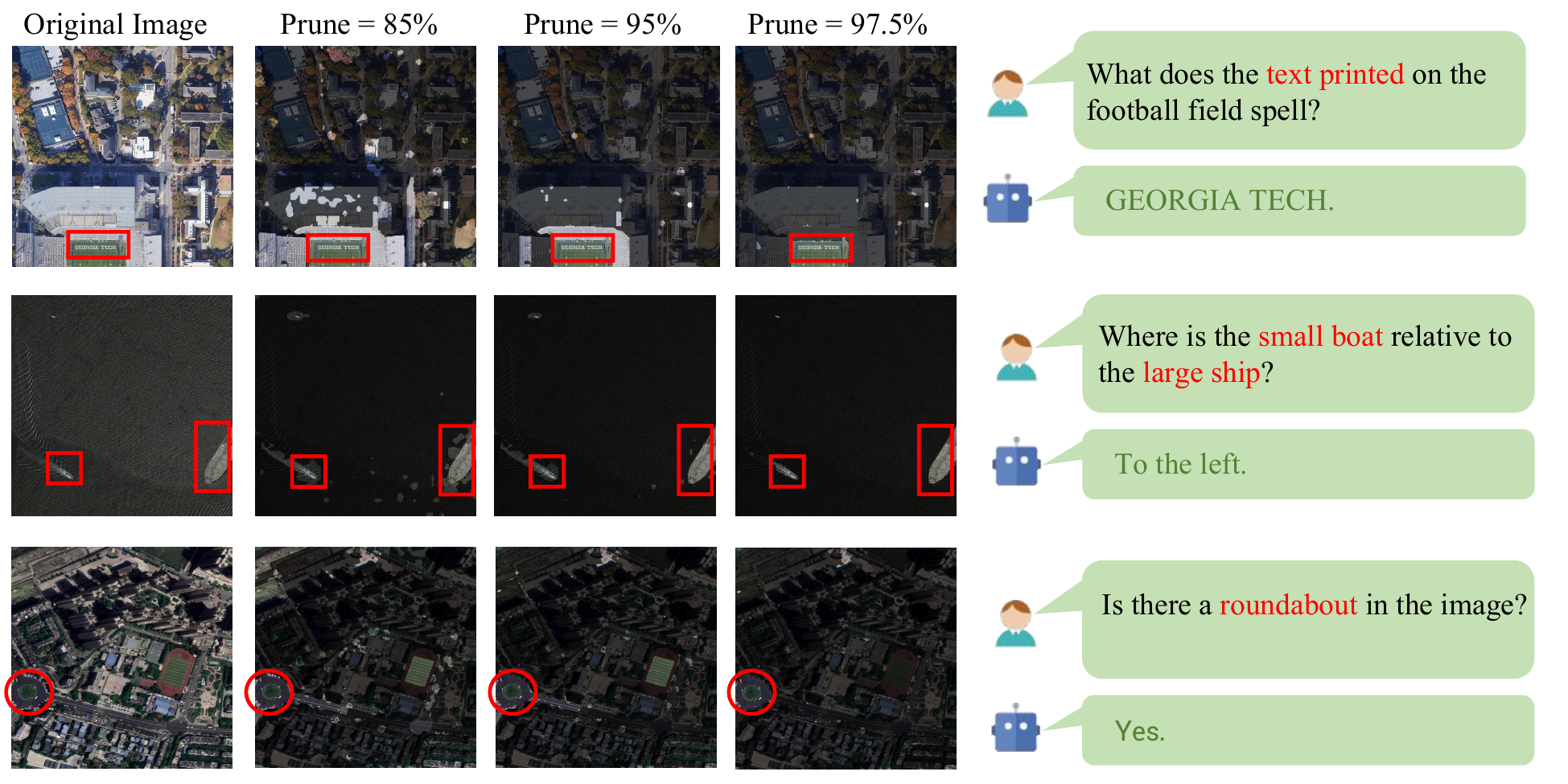}
    \caption{Visualization of clustering-guided token pruning. This example demonstrates how region-based partitioning helps retain structural objects. Even at a 97.5\% pruning ratio, small targets like boats or roundabouts are preserved within their clusters, ensuring the model retains sufficient context for spatial reasoning.}
    \label{fig:cluster_prune}
\end{figure*}

\paragraph{Visualization of Token Pruning.}
We visualize the spatial distribution of retained tokens in Figure~\ref{fig:attn_prune} and Figure~\ref{fig:cluster_prune} to demonstrate the effectiveness of our compression strategy.
The figures illustrate the surviving visual tokens at varying pruning ratios, ranging from 85\% to 97.5\%.
Figure~\ref{fig:attn_prune} displays the results using the attention-based selection, while Figure~\ref{fig:cluster_prune} shows the results based on the clustering partition.
In both scenarios, we observe that the method consistently preserves the semantic regions relevant to the user query.
As the pruning ratio increases, the irrelevant background areas are progressively removed (shown as black).
However, the critical visual evidence—such as specific ships, text on the ground, or swimming pools—remains visible even under extreme compression rates.
This visualization confirms that our approach successfully reduces computational redundancy while maintaining the necessary information for correct reasoning.

\paragraph{Qualitative Comparison and Analysis.}
We present a qualitative evaluation of our method in comparison with state-of-the-art Multimodal Large Language Models, including InternVL3.5-8B, GeoLLaVA-8K, and Qwen2.5-VL-72B.
The comparison results are illustrated in Figure~\ref{fig:cluster_visualation}.
These examples cover a range of challenging tasks such as fine-grained object counting (e.g., oil tanks, swimming pools), small object attribute recognition, and spatial reasoning.
As observed in the examples, baseline models often struggle with the high-resolution nature of remote sensing imagery; they frequently suffer from hallucinations or fail to recognize small objects, leading to incorrect counts.
In contrast, our method successfully identifies and reasons about these minute details.
This verifies that our region-aware token pruning strategy effectively retains the informative visual cues necessary for accurate answering.
Furthermore, we provide additional qualitative examples in Figure~\ref{fig:our_model_responses} to demonstrate the robustness and versatility of our model across various scenarios.

\paragraph{Additional Qualitative Results.}
We present additional qualitative examples in Figure~\ref{fig:our_model_responses} to further demonstrate the versatility of our method. 
These examples cover a wide range of tasks, including spatial reasoning, fine grained attribute recognition, and scene classification. 
As shown in the figures, our model effectively handles high resolution inputs and accurately answers questions regarding small targets, such as identifying the status of a bridge or the color of a chimney. 
This confirms that our approach maintains strong performance across diverse semantic categories and complex visual scenes.

\begin{figure*}[t]
    \centering
    \includegraphics[width=0.9\linewidth,height=0.28\textheight,keepaspectratio]{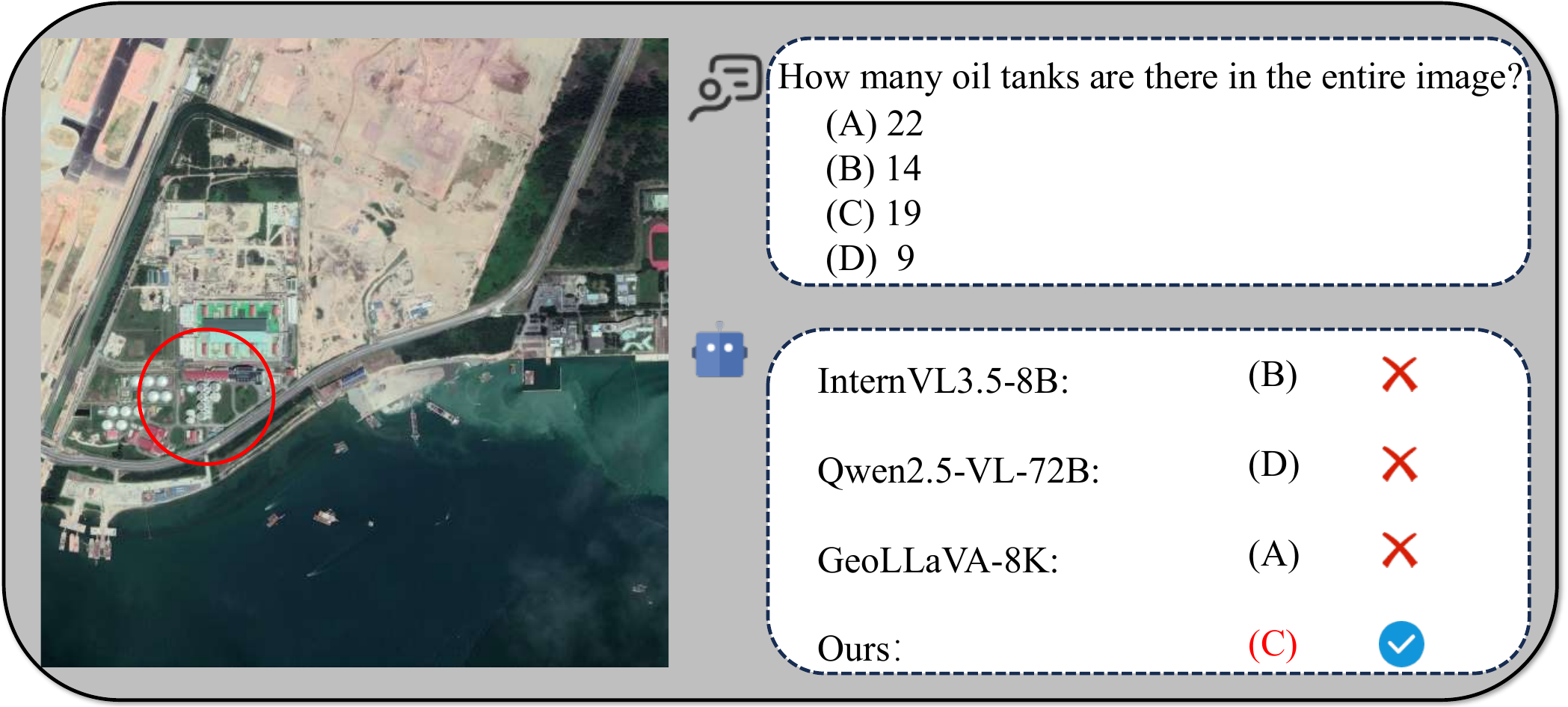}\vspace{0.6em}
    \includegraphics[width=0.9\linewidth,height=0.28\textheight,keepaspectratio]{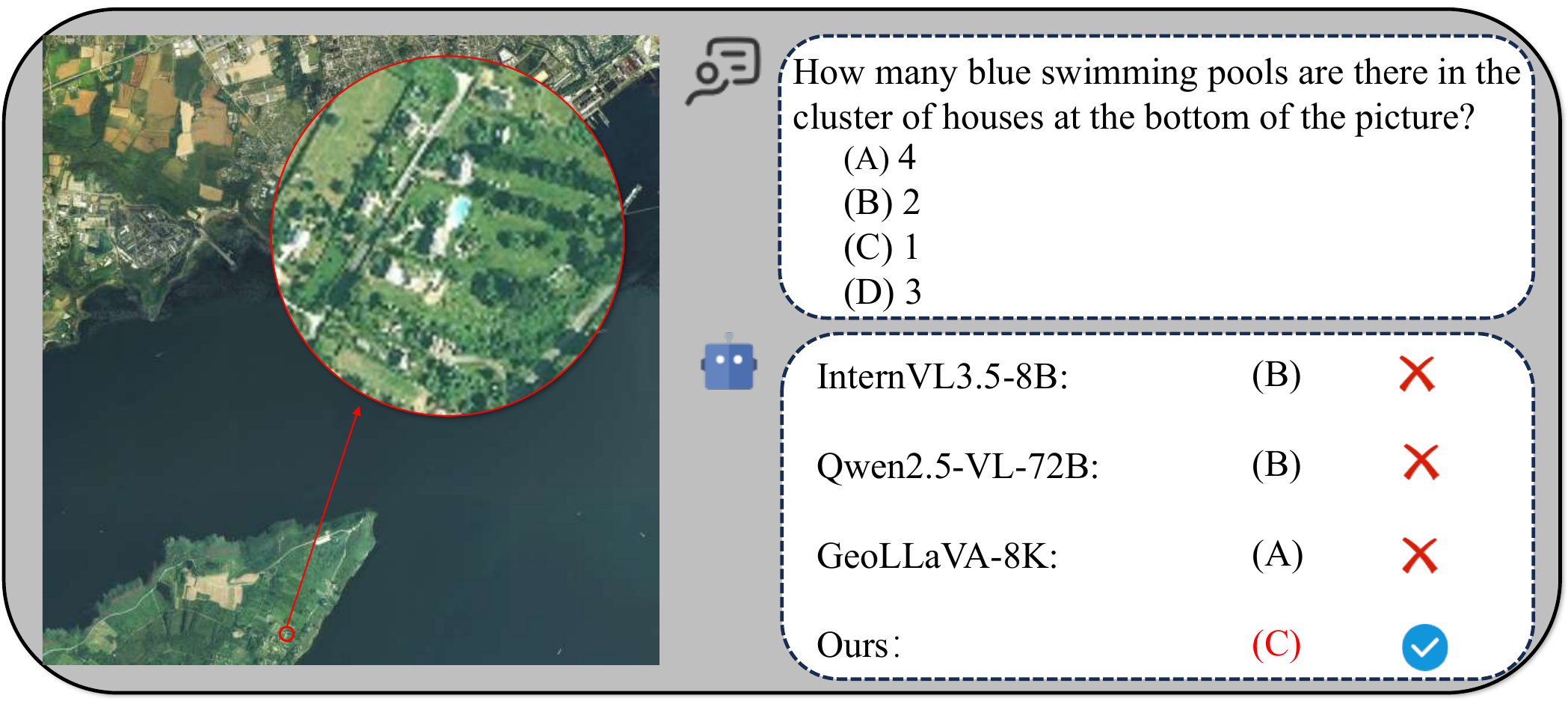}\vspace{0.6em}
    \includegraphics[width=0.9\linewidth,height=0.28\textheight,keepaspectratio]{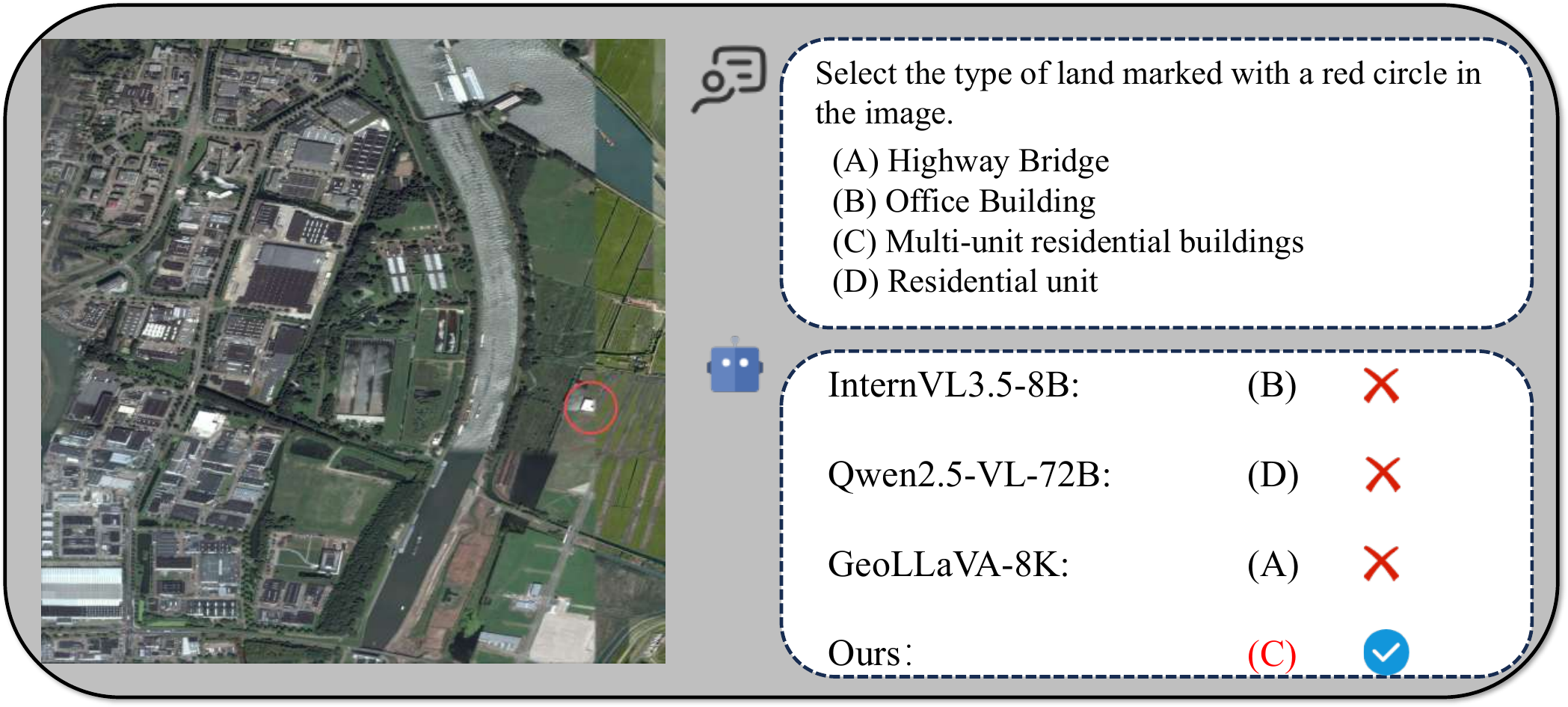}
    \caption{Qualitative comparison of model responses across InternVL3.5-8B, Qwen2.5-VL-72B, and our method. The examples demonstrate that our model correctly handles fine grained tasks such as counting dense oil tanks or identifying specific land types.}
    \label{fig:cluster_visualation}
\end{figure*}
\clearpage

\begin{figure*}[t]\ContinuedFloat
    \centering
    \includegraphics[width=0.9\linewidth,height=0.28\textheight,keepaspectratio]{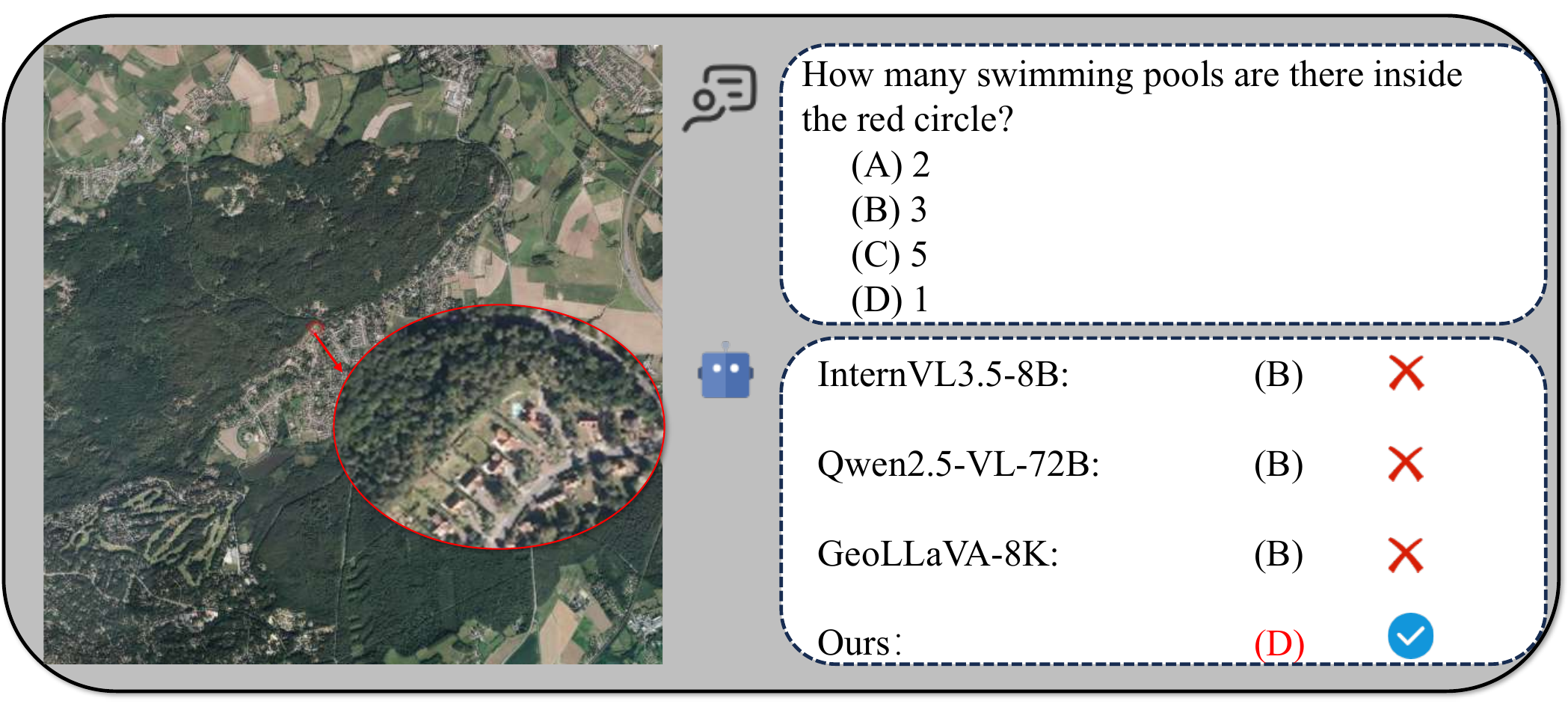}\vspace{0.6em}
    \includegraphics[width=0.9\linewidth,height=0.28\textheight,keepaspectratio]{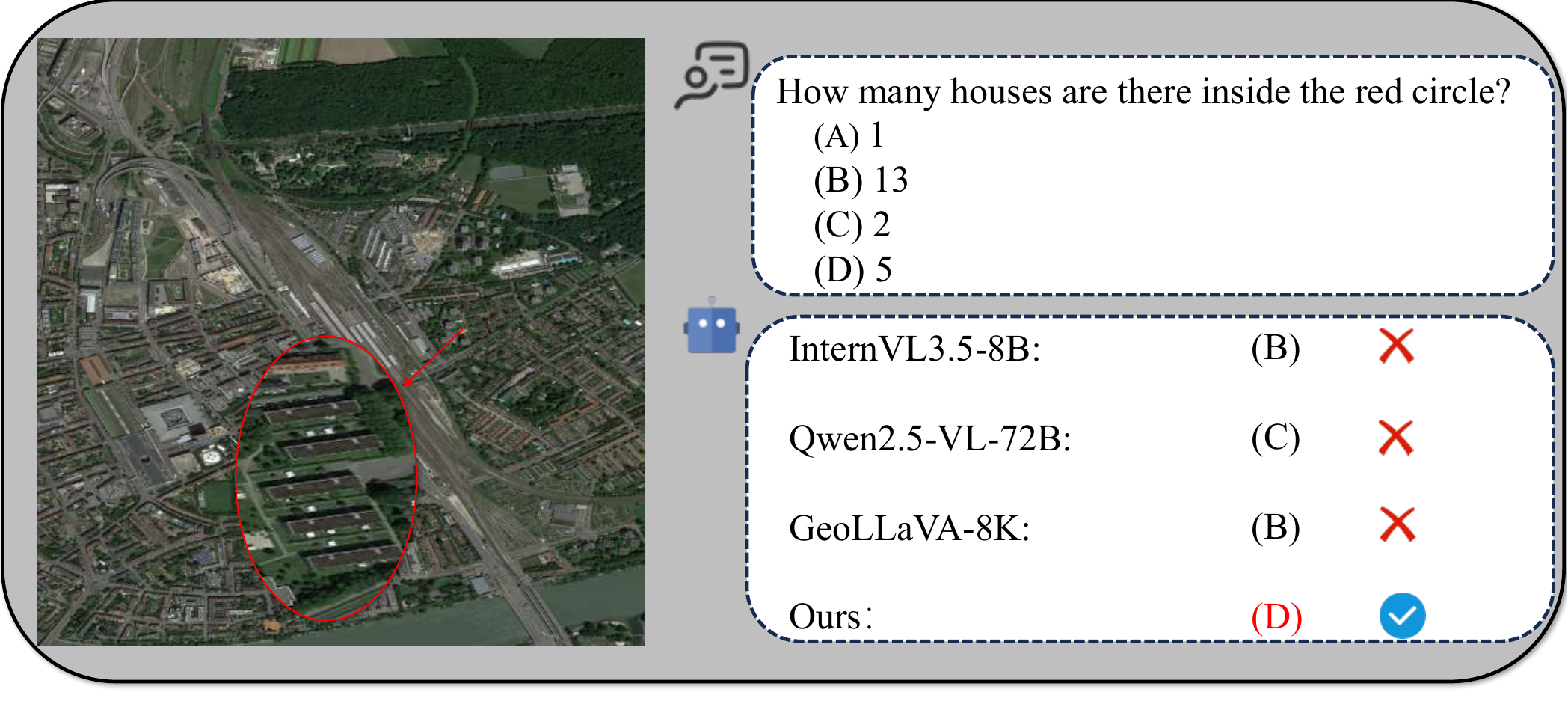}\vspace{0.6em}
    \includegraphics[width=0.9\linewidth,height=0.28\textheight,keepaspectratio]{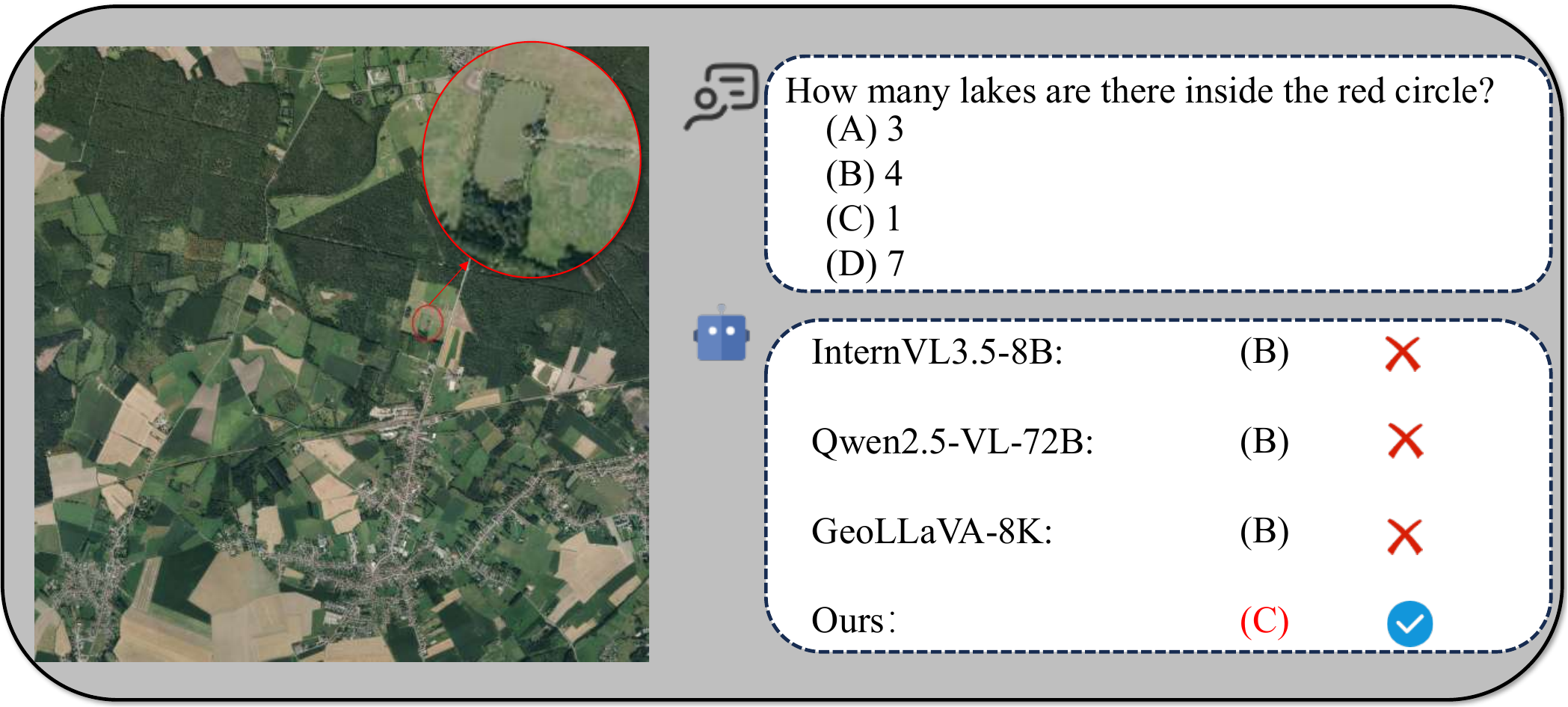}
    \caption{Qualitative comparison of model responses. Our method consistently provides accurate answers for challenging counting tasks within localized regions, effectively distinguishing targets like swimming pools or houses from complex backgrounds.}
\end{figure*}
\clearpage

\begin{figure*}[t]\ContinuedFloat
    \centering
    \includegraphics[width=0.9\linewidth,height=0.28\textheight,keepaspectratio]{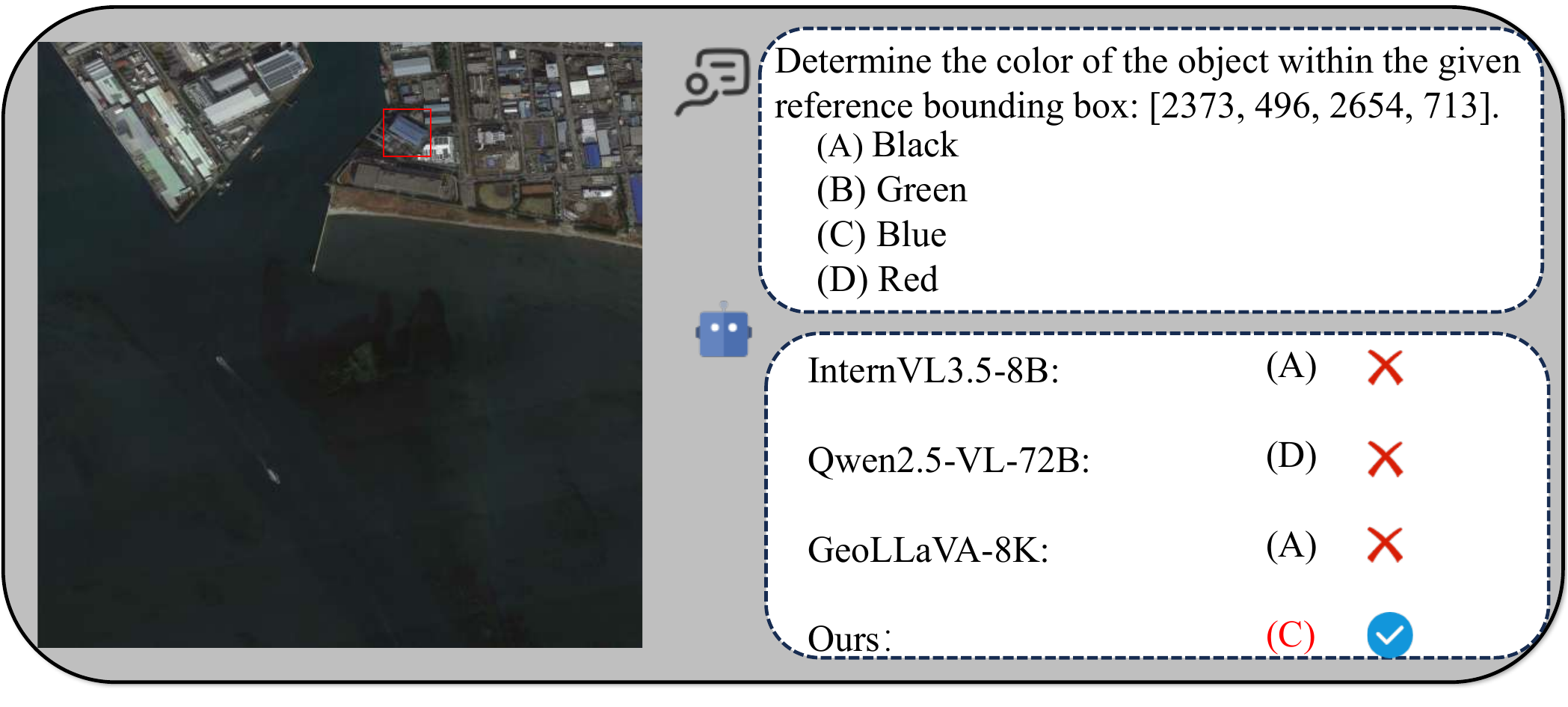}\vspace{0.6em}
    \includegraphics[width=0.9\linewidth,height=0.28\textheight,keepaspectratio]{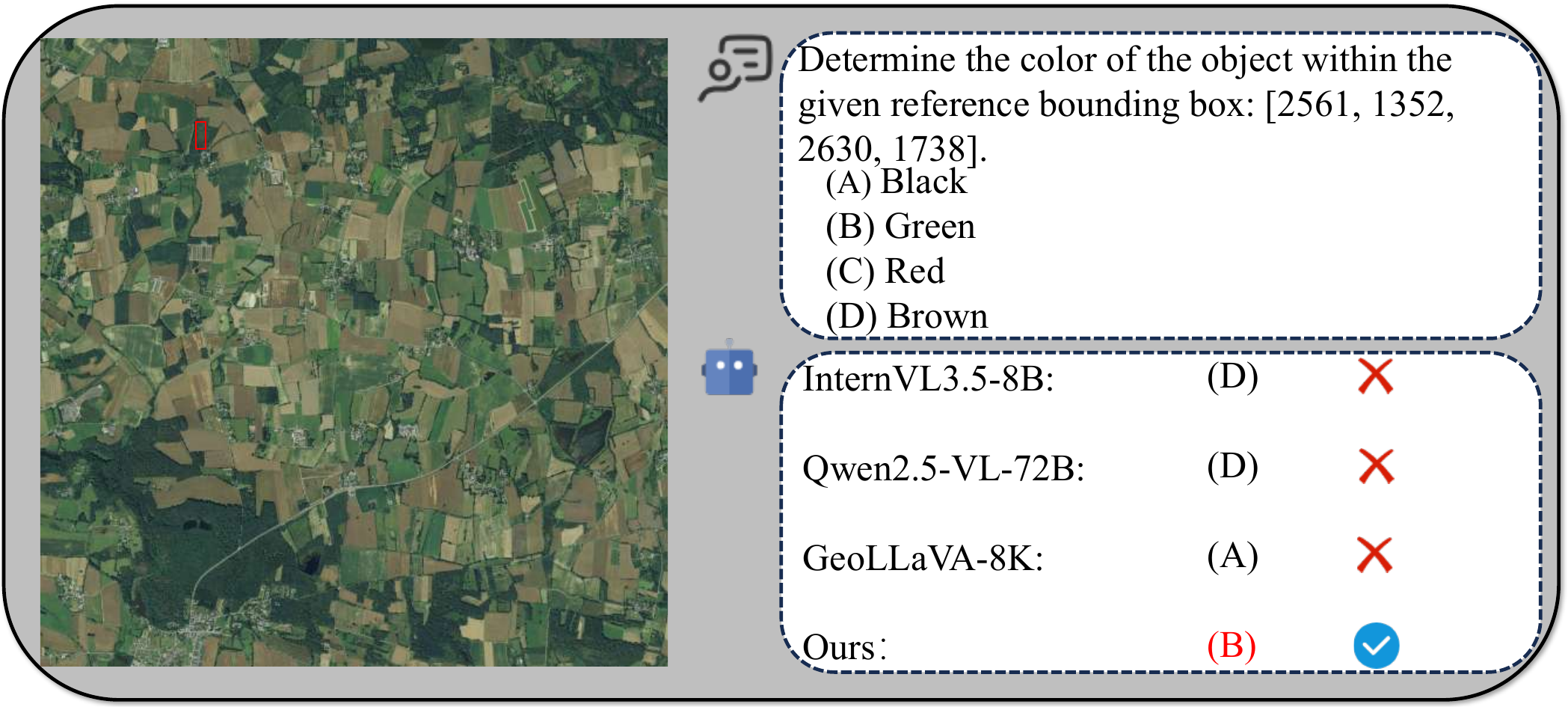}\vspace{0.6em}
    \includegraphics[width=0.9\linewidth,height=0.28\textheight,keepaspectratio]{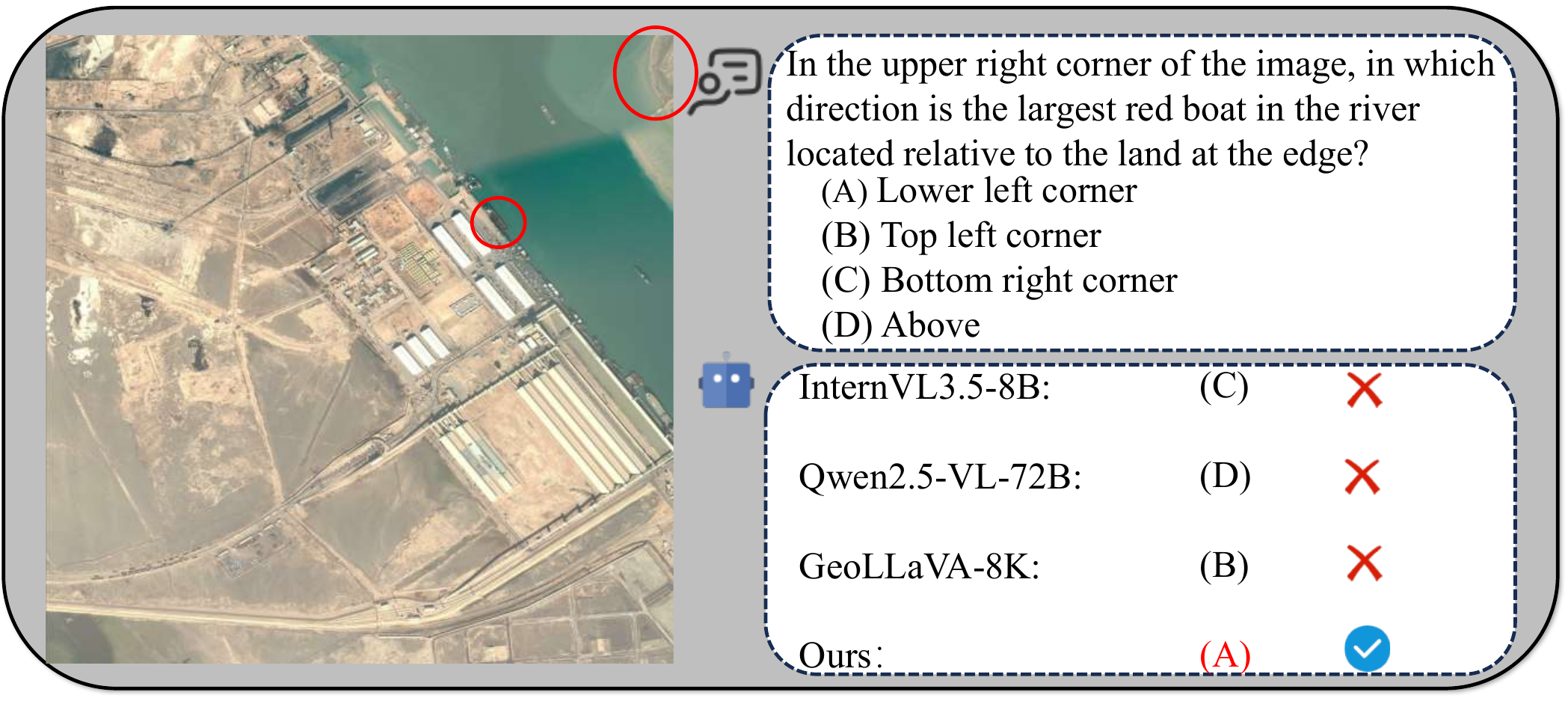}
    \caption{Qualitative comparison of model responses. The visualization highlights the ability of our model to ground the correct visual evidence for fine grained queries, specifically including color identification of small objects and reasoning about spatial directions.}
\end{figure*}

\begin{figure*}[t]
    \centering
    \includegraphics[width=0.95\linewidth,height=0.21\textheight,keepaspectratio]{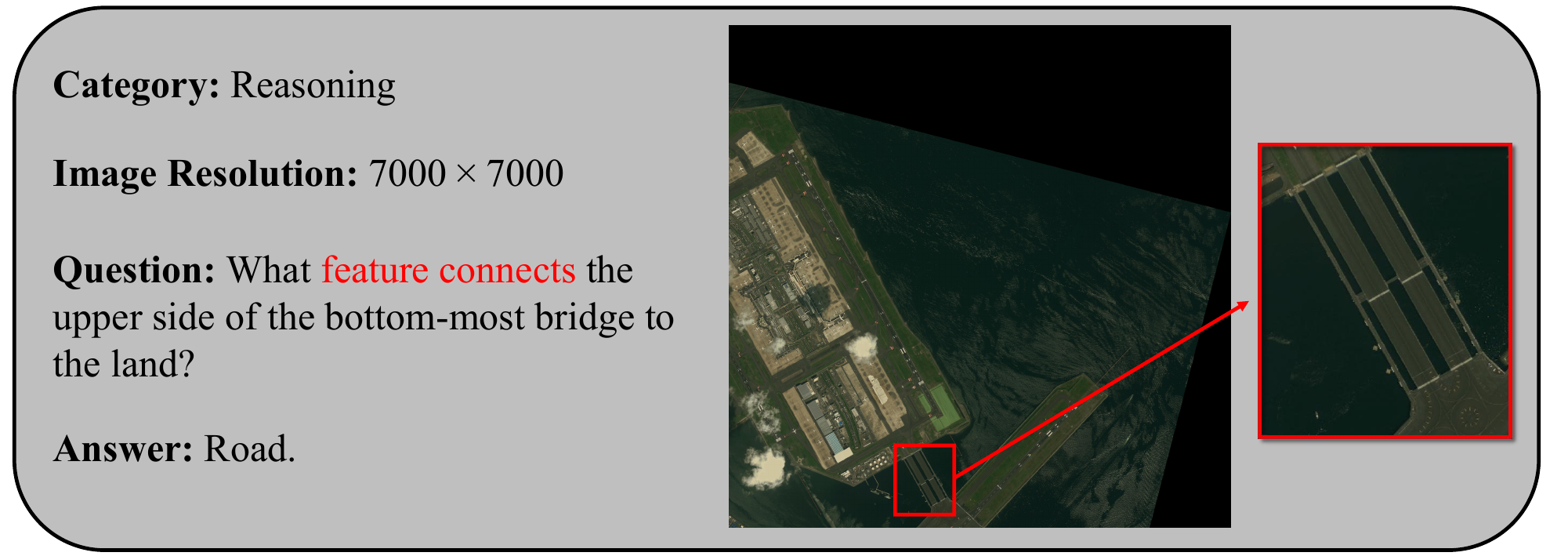}\vspace{0.9em}
    \includegraphics[width=0.95\linewidth,height=0.21\textheight,keepaspectratio]{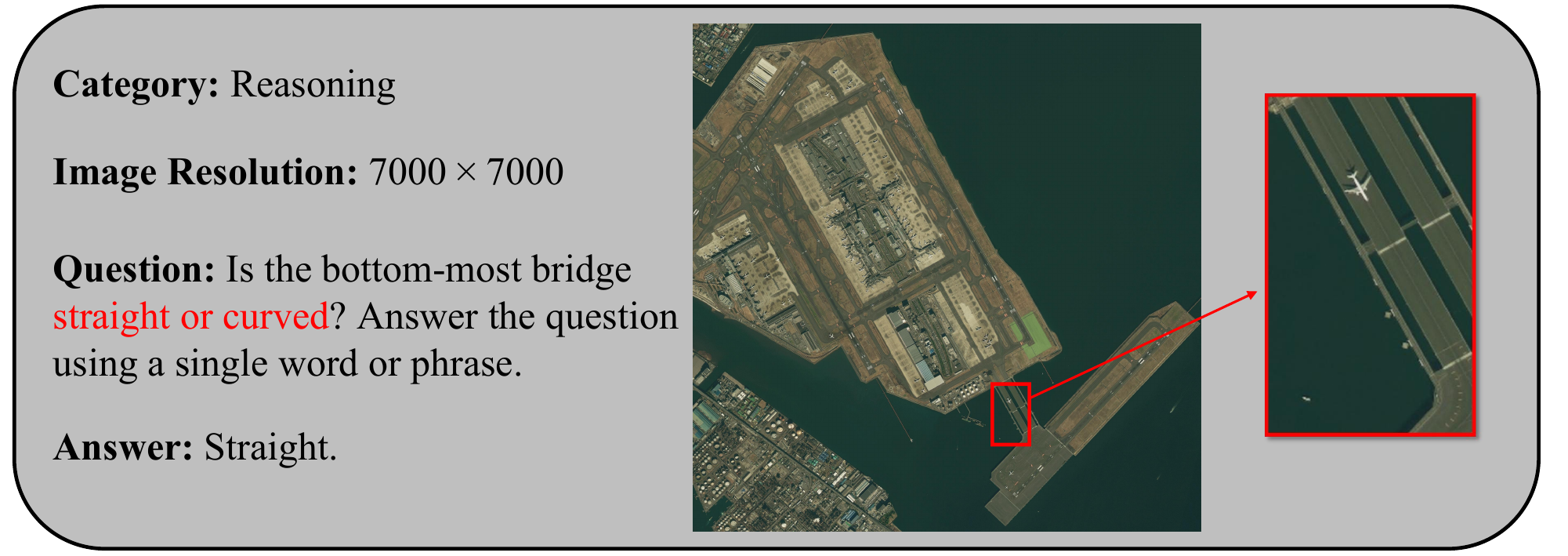}\vspace{0.9em}
    \includegraphics[width=0.95\linewidth,height=0.21\textheight,keepaspectratio]{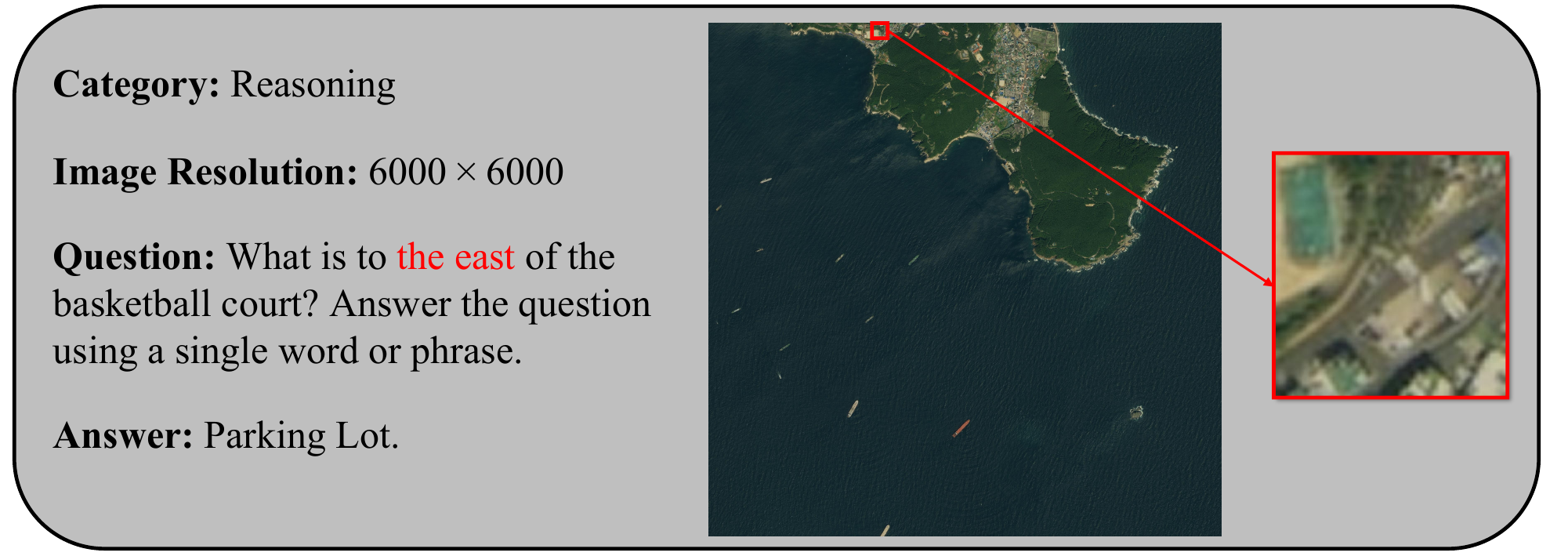}\vspace{0.9em}
    \includegraphics[width=0.95\linewidth,height=0.21\textheight,keepaspectratio]{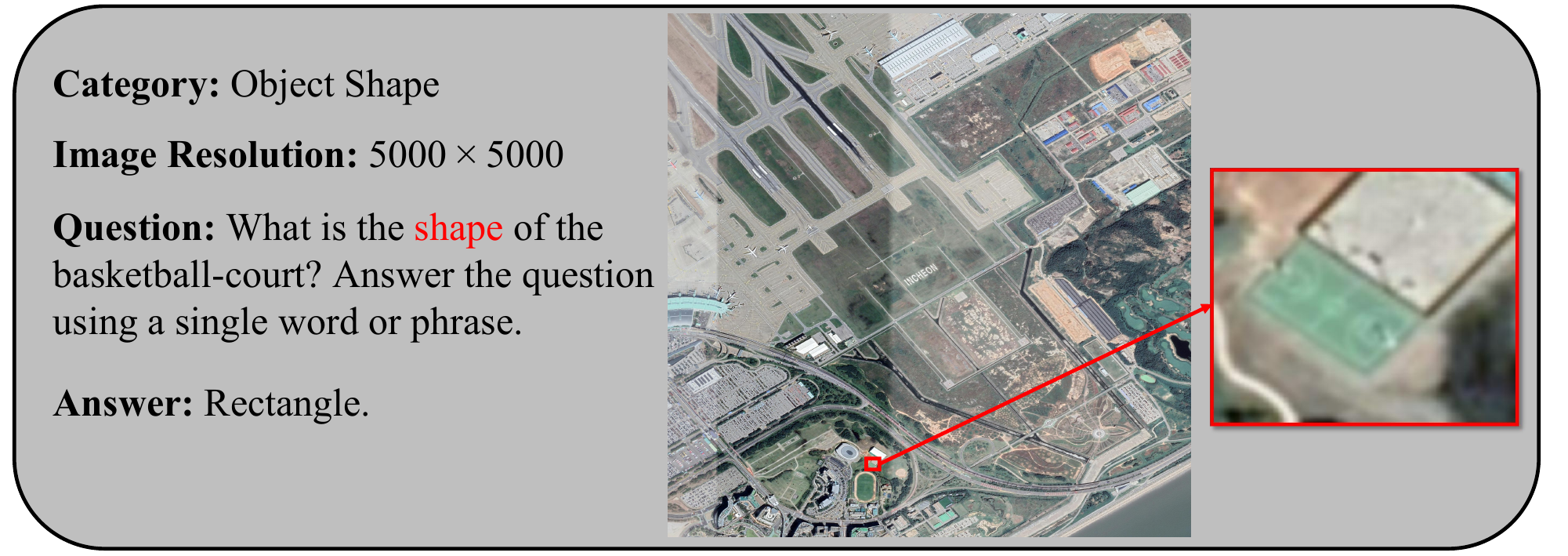}
    \caption{Qualitative examples of responses generated by our model. These samples illustrate the versatility of our method in handling diverse tasks, including spatial reasoning about road orientation and fine-grained attribute recognition of small objects.}
    \label{fig:our_model_responses}
\end{figure*}
\clearpage

\begin{figure*}[t]\ContinuedFloat
    \centering
    \includegraphics[width=0.95\linewidth,height=0.21\textheight,keepaspectratio]{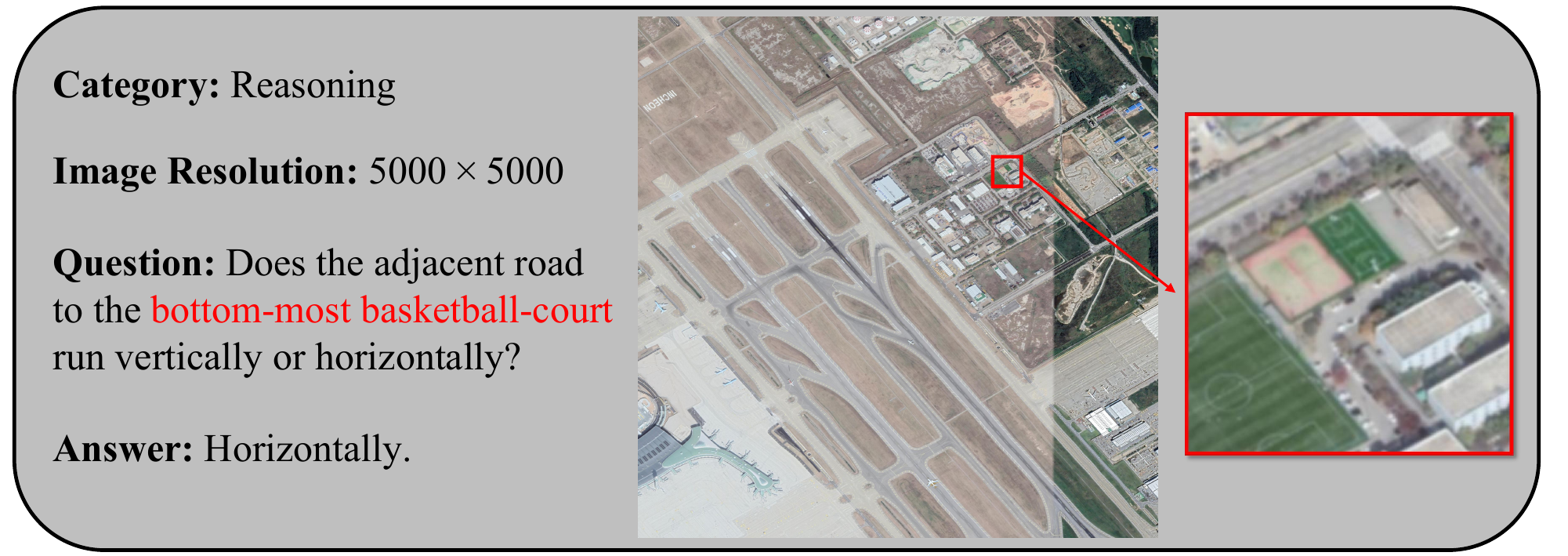}\vspace{0.9em}
    \includegraphics[width=0.95\linewidth,height=0.21\textheight,keepaspectratio]{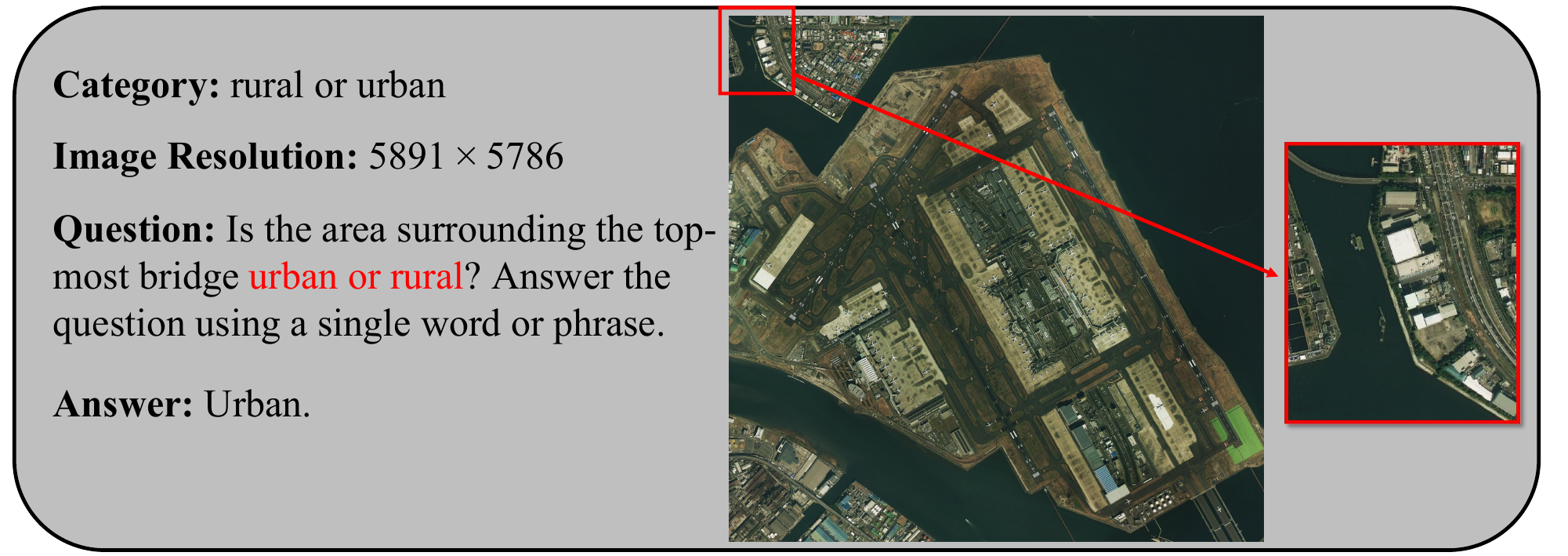}\vspace{0.9em}
    \includegraphics[width=0.95\linewidth,height=0.21\textheight,keepaspectratio]{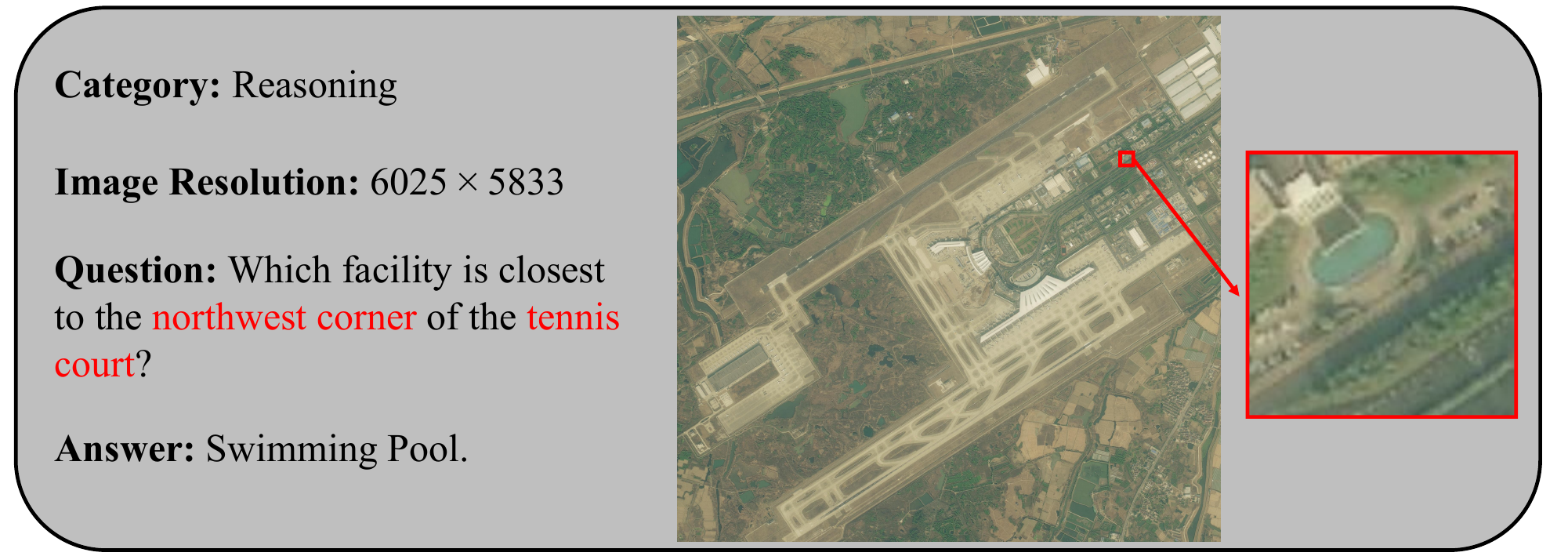}\vspace{0.9em}
    \includegraphics[width=0.95\linewidth,height=0.21\textheight,keepaspectratio]{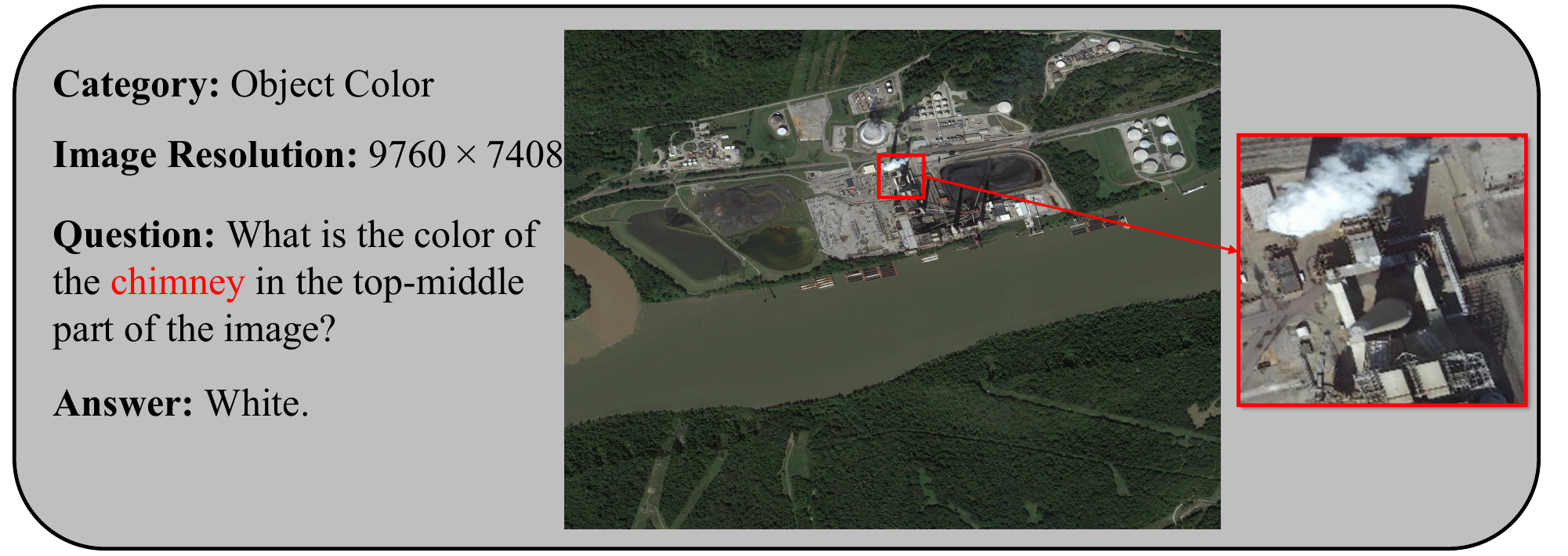}
    \caption{Qualitative examples of responses generated by our model. The visualization demonstrates the ability of the model to interpret geometric shapes and complex spatial relationships, such as identifying the curvature of a bridge or determining the relative position.}
\end{figure*}
\clearpage

\begin{figure*}[t]\ContinuedFloat
    \centering
    \includegraphics[width=0.95\linewidth,height=0.21\textheight,keepaspectratio]{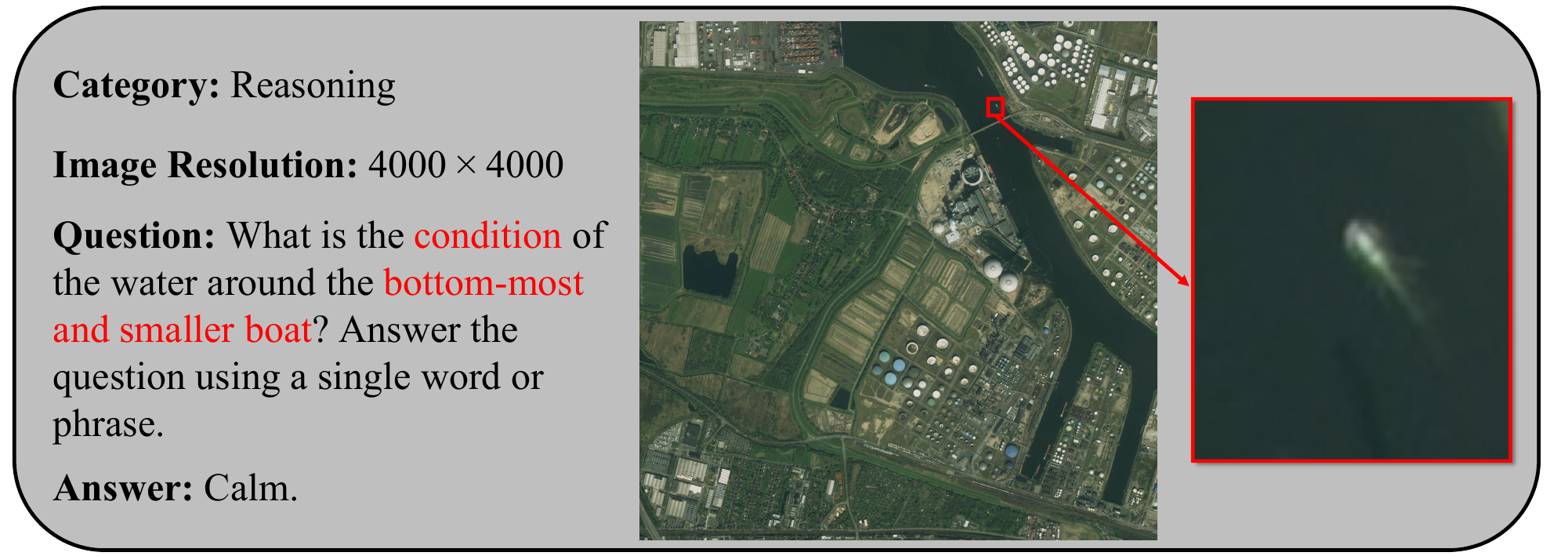}\vspace{0.9em}
    \includegraphics[width=0.95\linewidth,height=0.21\textheight,keepaspectratio]{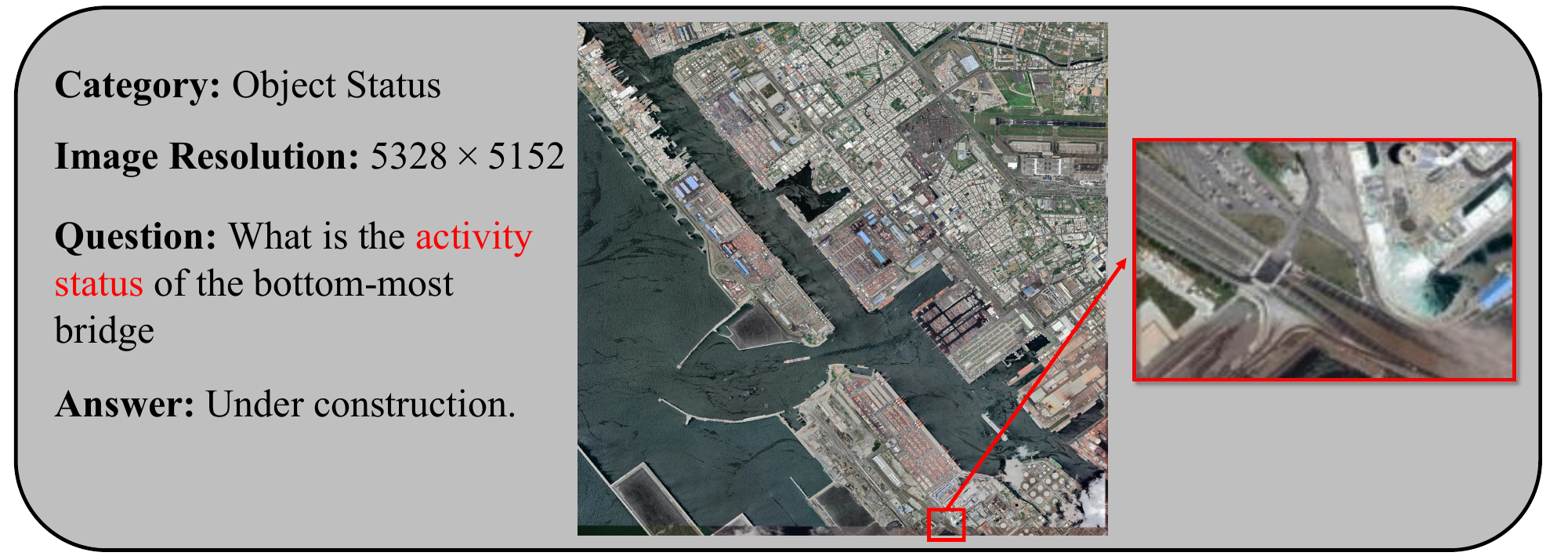}\vspace{0.9em}
    \includegraphics[width=0.95\linewidth,height=0.21\textheight,keepaspectratio]{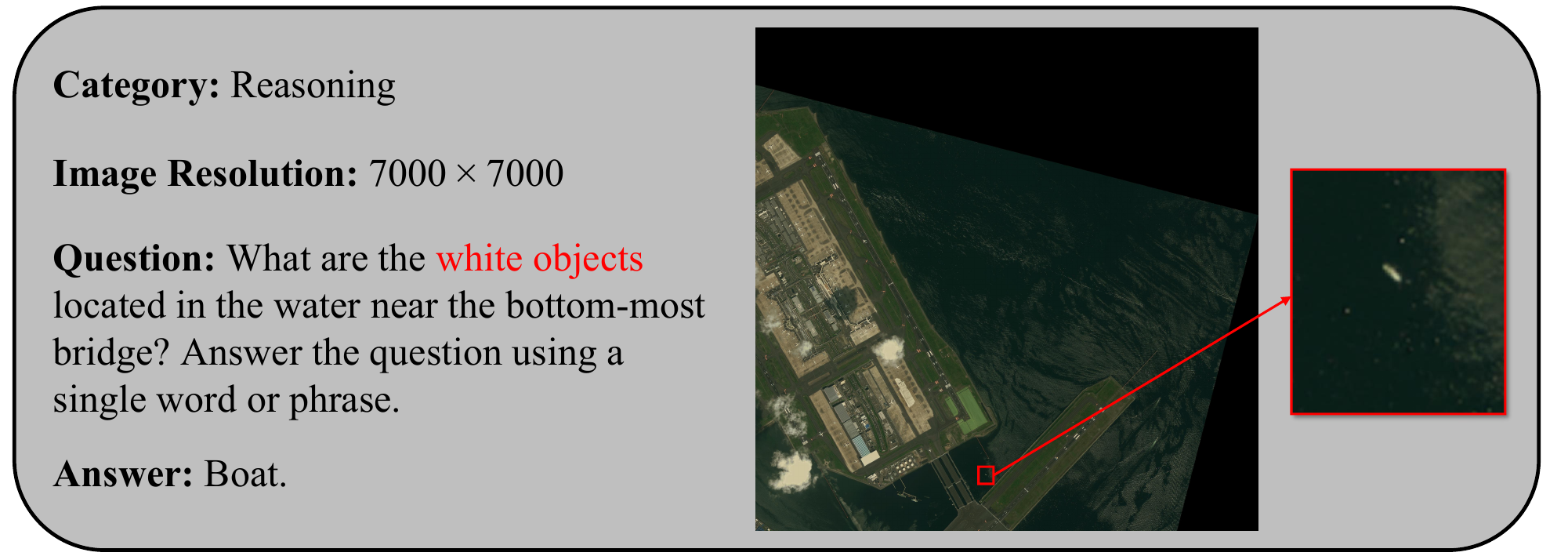}
    \includegraphics[width=0.95\linewidth,height=0.21\textheight,keepaspectratio]{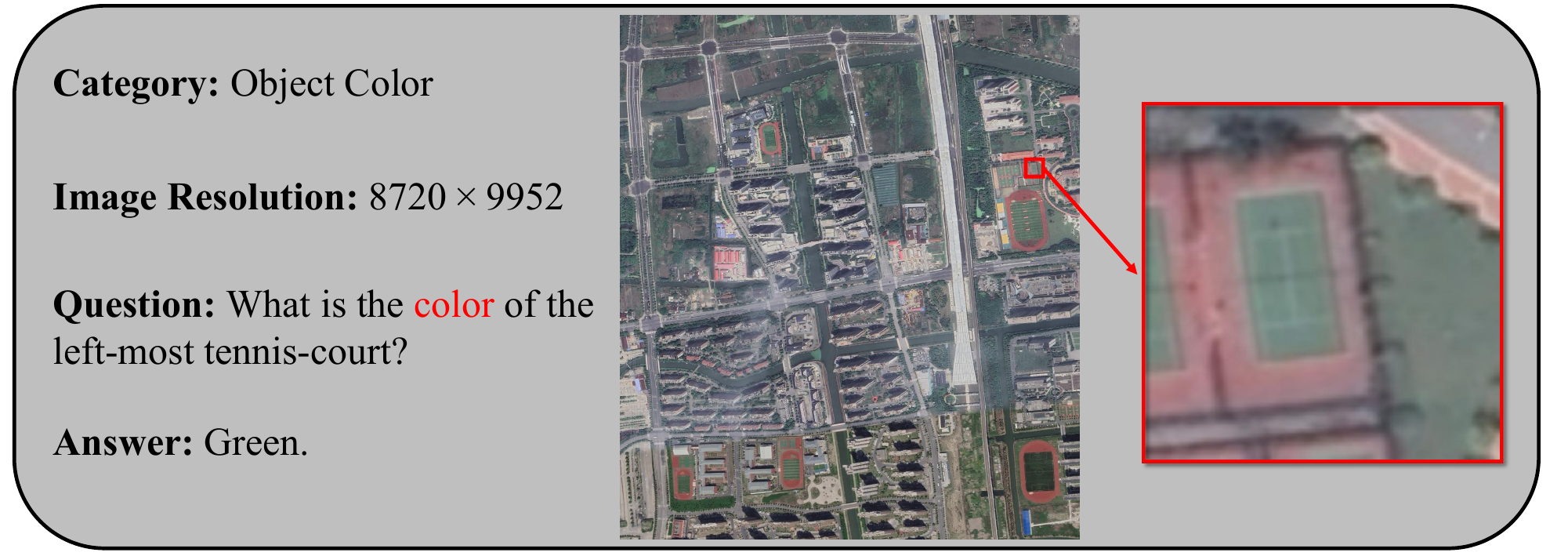}
    \caption{Qualitative examples of responses generated by our model. These results show that our model effectively captures dynamic scene details, such as assessing water conditions, identifying construction activities, detecting minute objects, and recognizing colors.}
\end{figure*}
\clearpage

\end{document}